%% file: neurips_2026_final.tex
\documentclass{article}

\usepackage[nonatbib, main, final]{neurips_2026}

\usepackage[utf8]{inputenc} %
\usepackage[T1]{fontenc}    %
\usepackage{hyperref}       %
\usepackage{url}            %
\usepackage{booktabs}       %
\usepackage{amsfonts}       %
\usepackage{nicefrac}       %
\usepackage{xcolor}         %

\usepackage{soul}

\usepackage{subcaption}
\usepackage{graphicx}
\usepackage{amsmath}
\usepackage{colortbl}
\usepackage{multirow}
\usepackage{rotating}
\usepackage{amsthm}
\newtheorem{theorem}{Theorem}
\newtheorem{lemma}[theorem]{Lemma}
\newtheorem{proposition}[theorem]{Proposition}
\usepackage{enumitem}
\soulregister\ref7
\soulregister\eqref7
\soulregister\cite7
\usepackage{wrapfig}     %
\usepackage{booktabs}    %
\usepackage{algorithm}
\usepackage{algpseudocode}
\usepackage{amssymb}
  \usepackage{longtable}
  \usepackage[table]{xcolor}
  \usepackage{array}

  \makeatletter
\providecommand{\@trackname}{}
\makeatother

\title{Spectral Unlearning: Transformer Structure-Preserving Updates for Language Models}

\author{%
    Sung~Il~Choi \\
  {\small \qquad\quad\:\, School of Electrical Engineering \qquad\quad\:\,} \\
  Korea University \\
  Seoul 02841, South Korea \\
  \texttt{sungchoi@korea.ac.kr} \\
 \And
    Junhao~Cai \\
  {\small Department of Computer Science and Engineering} \\
  Korea University \\
  Seoul 02841, South Korea \\
  {\texttt{junhochae@korea.ac.kr}}
  \And
  Dohun~Kim \\
  {\small Department of Computer Science and Engineering}  \\
  Korea University \\
  Seoul 02841, South Korea \\
  {\texttt{dohunkim@korea.ac.kr}}
  \And
  Changhee~Joo\thanks{Corresponding author.} \\
  {\small Department of Computer Science and Engineering} \\
  Korea University \\
  Seoul 02841, South Korea \\
  \texttt{changhee@korea.ac.kr}
}

\begin{document}

\maketitle

\begin{abstract}
Machine unlearning in large language models (LLMs) aims to remove specific learned knowledge, capabilities, or behaviors while preserving model utility. Existing methods have mostly focused on the design of unlearning objectives, while largely overlooking the intrinsic matrix structure of transformer weights. In particular, standard optimizers operate on independent scalar coordinates, thereby ignoring this matrix structure. We propose \emph{spectral unlearning}, an update framework that combines standard unlearning objectives with matrix-structured optimization. We show that the spectral update corresponds to the steepest-descent direction for transformers composed of linear layers with normalized inputs and outputs. Our approach is applied on top of existing unlearning methods and evaluated across complementary axes of memorization, privacy, and utility, summarized via their harmonic mean. Spectral unlearning consistently outperforms a broad range of baselines and closes most of the gap to a retain-trained reference. Code is available at \url{https://github.com/unlearn-ml/spectral-unlearning}.
\end{abstract}

\section{Introduction}

Large language models can memorize private information~\cite{carlini2023quantifying} and produce harmful outputs~\cite{wei2023jailbroken}. Misuse of these models even extends to cyberattacks~\cite{fang2024llmagentsautonomouslyhack} and biological threats~\cite{phuong2024evaluating}. Machine unlearning seeks to remove such knowledge from a trained model while preserving its overall utility~\cite{maini2024tofu, pmlr-v235-li24bc}. The challenge in LLM machine unlearning is that residual knowledge surfaces through diverse pathways, including verbatim generation~\cite{tirumala2022memorization}, extraction attacks~\cite{wang2025towards,274574}, and privacy leakage~\cite{zhang2025mink}. Prior methods have addressed this challenge primarily by designing forget and retain loss functions \cite{liu2025rethinking}. However, in essence, an unlearning update is determined by its direction and magnitude in parameter space~\cite{wang2025rethinking}. Designing a new loss therefore reshapes the update direction.

The update direction depends not only on the loss but also on the geometry assumed on the parameter space. Most LLM unlearning methods rely on entry-wise update rules that scale each parameter independently, ignoring correlations between parameters~\cite{yao2024large}. However, transformer weights are matrix-valued and act on root-mean-square (RMS) normalized hidden states \cite{NEURIPS2019_1e8a1942}. Therefore, the natural geometry on these parameters is matrix-valued rather than entry-wise. Aligning the update direction with this geometry requires update rules defined directly on matrix-valued parameters.

We propose \emph{spectral unlearning}, a transformer structure-preserving unlearning framework that updates the entries of each transformer weight matrix jointly. Spectral unlearning is general across unlearning objectives, applying uniformly to gradient-difference, preference-based, and representation-level losses.
In particular, the weight matrices in transformers map RMS-normalized inputs to outputs. We show that the proposed updates are the steepest-descent directions \cite{boyd2004convex} in this RMS-induced space. 
Computing this direction exactly is infeasible at LLM scale, so we introduce three practical methods with different matrix-structured strategies. %

\begin{figure}[t]
    \centering
    \includegraphics[width=\textwidth]{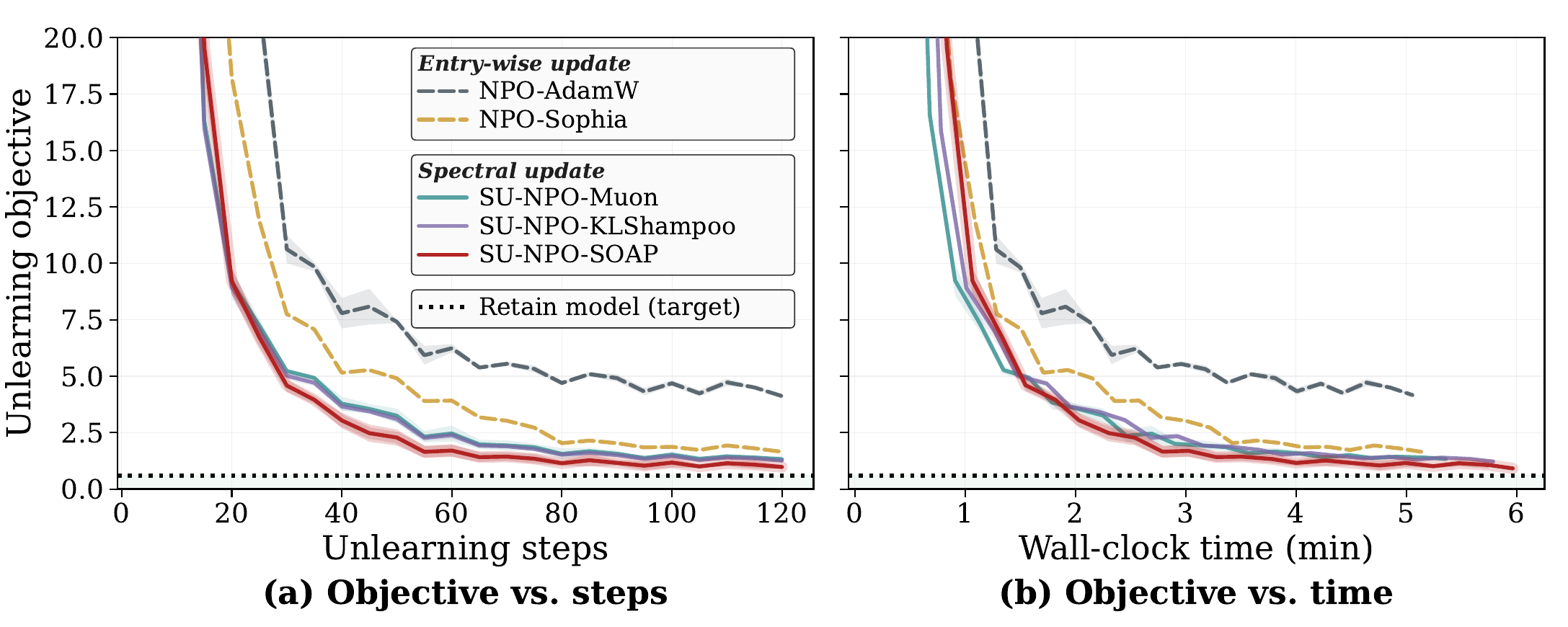}
    \caption{The Llama-3.2-1B model forgets data from the TOFU benchmark through the negative preference optimization (NPO) unlearning method. Loss curves are plotted against (a) unlearning steps and (b) wall-clock time. Dashed curves use conventional entry-wise updates. Solid curves use our proposed spectral update (SU), realized through Muon, KLShampoo, and SOAP.}
    \label{fig:fig1}
\end{figure}

Fig.~\ref{fig:fig1} previews unlearning objective curves, comparing entry-wise updates (NPO-AdamW \cite{zhang2024negative} and NPO-Sophia \cite{jia-etal-2024-soul}) with our spectral variants. Spectral updates reach the retain-trained target loss in fewer steps and lower wall-clock time. This advantage holds across unlearning objectives, model scales, spectral variants, and learning rates. Our contributions are summarized as follows.

\begin{itemize}[leftmargin=*]
     \item 
     We propose \emph{spectral unlearning}, a loss-agnostic framework that updates the entries of each weight matrix jointly. We identify the geometry of the update as a separate and complementary design choice. We realize the spectral update with Muon, KLShampoo, and SOAP.
     \item 
     We show that the natural norm of pre-norm transformers, the RMS-to-RMS norm, is a rescaled spectral norm. Steepest descent of the unlearning objective under this norm gives the spectral update direction. Spectral updates capture the Kronecker structure of unlearning gradients.
     \item
     We evaluate unlearning methods across memorization, privacy, and utility, aggregated via the harmonic mean. Spectral updates improve the aggregate score and the memorization–utility trade-off. The improvement holds over a wide range of learning rates and under matched wall-clock time. Spectral updates also reduce the recovery of forgotten knowledge under relearning attacks.
\end{itemize}

\section{Preliminaries}

\subsection{LLM unlearning objectives}
\label{sec:unlearning_objectives}

\paragraph{Notation.}
Let $\pi_{w}$ denote a language model parameterized by $w$, mapping an input context $x$ to a distribution $\pi_{w}(y \mid x)$ over output sequences $y$. Let $\mathcal{D}_u = \{(x_u, y_u)\}$ denote the \emph{forget set}, whose influence  we seek to remove,
and $\mathcal{D}_r = \{(x_r, y_r)\}$ the \emph{retain set}, whose behavior we aim to preserve.

\paragraph{Unlearning as optimization \cite{liu2025rethinking}.}
Machine unlearning aims to recover parameters that behave as if the model had not been trained on $\mathcal{D}_u$ while preserving performance on $\mathcal{D}_r$. A common abstraction formulates machine unlearning as the regularized objective
\begin{align}
    \min_{w}
    \bigl\{ 
        \mathcal{L}_{\mathrm{u}}(w; \mathcal{D}_u)
        + \alpha\, \mathcal{L}_{\mathrm{r}}(w; \mathcal{D}_r)
    \bigr\},
\label{eq:mu_general}
\end{align}
where $\mathcal{L}_{\mathrm{u}}$ and $\mathcal{L}_{\mathrm{r}}$ denote the forget and retain losses, respectively, and $\alpha \geq 0$ controls the trade-off. Different choices of $\mathcal{L}_{\mathrm{u}}$ and $\mathcal{L}_{\mathrm{r}}$ give rise to different unlearning methods.

\paragraph{Gradient difference (GD) \cite{maini2024tofu}.}
GD minimizes the log-likelihood on the forget set while maximizing it on the retain set. In expectation, this corresponds to the objective
\begin{align}
    \min_{w}
    \bigl\{ \,
        \mathbb{E}_{\mathcal{D}_u}
            \!\left[ \log \pi_{w}(y_u \mid x_u) \right]
    + \alpha\,
        \mathbb{E}_{\mathcal{D}_r}
            \!\left[ -\log \pi_{w}(y_r \mid x_r) \right]
    \, \bigr\},
\label{eq:GD}
\end{align}
where the first term pushes the model away from $\mathcal{D}_u$ and the second term pulls it toward $\mathcal{D}_r$.

\paragraph{Negative preference optimization (NPO) \cite{zhang2024negative}.}
The forget term in \eqref{eq:GD} is unbounded below, which can lead to divergence during unlearning.
NPO addresses this issue by introducing a bounded forget term inspired by preference optimization \cite{rafailov2023direct}. Let $w_0$ denote a reference model, typically the parameters before unlearning. The learning objective is
\begin{align}
    \min_{w} \, \frac{2}{\beta} \, \textstyle
    \mathbb{E}_{\mathcal{D}_u} 
    \!\left[
        \log \!\left( 
            1 + \left( \frac{\pi_{w}(y_u \mid x_u)}{\pi_{w_0}(y_u \mid x_u)} \right)^{\!\beta}
        \right)
    \right],
\label{eq:npo}
\end{align}
where $\beta > 0$ controls the smoothness. 
The ratio $\pi_{w}/\pi_{w_0}$ acts as an implicit anchor, preventing the model from drifting too far from its initial behavior.

\paragraph{Representation misdirection for unlearning (RMU) \cite{pmlr-v235-li24bc}.}
RMU operates at the representation level, steering hidden states rather than output probabilities. Inspired by representation engineering \cite{zou2023transparency}, it removes forget-set information by modifying intermediate activations.
Let $h_{w}^{(\ell)}(x) \in \mathbb{R}^{d_h}$ denote the hidden state at the output of transformer block $\ell$ for input $x$. The objective is
\begin{align}
    \min_{w} \, \Bigl\{ \,
    \mathbb{E}_{\mathcal{D}_u}
        \!\bigl[ \|h_{w}^{(\ell)}(x_u) - c\,u\|_{2}^{2} \bigr]
    + \alpha\,
    \mathbb{E}_{\mathcal{D}_r}
        \!\bigl[ \|h_{w}^{(\ell)}(x_r) - h_{w_0}^{(\ell)}(x_r)\|_{2}^{2} \bigr]
    \, \Bigr\},
\label{eq:rmu}
\end{align}
where $u \in \mathbb{R}^{d_h}$ is a fixed random direction and $c > 0$ controls its magnitude. The forget term steers representations of forget samples toward $c\,u$, while the retain term anchors representations of retain samples to the reference model. %

\subsection{Steepest descent in a normed space}

Neural networks are typically trained with gradient descent, which updates the weights along the negative gradient. The negative gradient is the steepest direction under the Euclidean norm. Steepest descent extends the idea to an arbitrary norm~\cite{boyd2004convex}. Steepest descent chooses a weight update $\Delta w$ by minimizing a local quadratic model of the loss. 
If the unlearning objective $\mathcal{L}$ is locally $\lambda$-smooth with respect to a norm $\|\cdot\|$, then the loss function satisfies
\begin{align}
    \mathcal{L}(w + \Delta w) \leq
        \mathcal{L}(w) + g^{\top}\Delta w + \frac{\lambda}{2}\|\Delta w\|^{2},
\label{eq:local_quad_model}
\end{align}
where $g := \nabla \mathcal{L}(w)$ denotes the gradient. Minimizing the right-hand side yields the steepest descent update under the geometry induced by $\|\cdot\|$. Different choices of the norm lead to different optimization algorithms.

\paragraph{Direction and magnitude.}
The minimizer of \eqref{eq:local_quad_model} admits a decomposition into a direction and a magnitude.
Let $\Delta w = c \, t$ with $c \geq 0$ and $\|t\| = 1$, then the solution is given by
\begin{align}
\Delta w^{*}
=
\arg\min_{\Delta w}
\Bigl[g^{\top}\Delta w + \tfrac{\lambda}{2}\|\Delta w\|^{2}\Bigr]
=
-\,c^{*}(g, \lambda)\cdot t^{*}(g),
\qquad
t^{*}(g) = \arg\max_{\|t\|=1} g^{\top} t,
\label{eq:steepest_split}
\end{align}
where $t^{*}(g)$ is the unit-norm vector most aligned with the gradient, %
and $c^{*}(g, \lambda)$ determines the step size. %
In the context of unlearning, this decomposition recovers the two design choices of an unlearning update. The update direction is determined by the unlearning objective with respect to the chosen norm, while the magnitude is controlled by the learning rate. %

\subsection{Steepest descent for entry-wise updates}
\label{sec:adam_as_steepest}

Unlearning is most commonly performed using entry-wise updates, i.e., AdamW. Steepest descent offers a way to interpret the choice of AdamW~\cite{bernstein2024old}. Removing the exponential moving averages and decoupled weight decay from AdamW leaves an update that depends only on the current gradient
\begin{align}
    w_{t+1}
        \textstyle = w_{t} - \eta \frac{g_{t}}{\!\sqrt{g_{t}^{2}}}
        = w_{t} - \eta \cdot \operatorname{sign}(g_{t}),
\end{align}
where $g_{t} \in \mathbb{R}^{n}$ is the gradient at step $t$, $\eta > 0$ is the step size, and all operations are applied entry-wise. The simplified update is \emph{sign descent} on the flattened parameter vector.

Sign descent is a special case of steepest descent. Choosing the vector $\ell_{\infty}$ norm $\|v\|_{\infty} := \max_{i}|v_{i}|$ in \eqref{eq:steepest_split} gives
\begin{align}
\arg\min_{\Delta w \in \mathbb{R}^{n}}
\Bigl[g^{\top}\Delta w + \tfrac{\lambda}{2}\|\Delta w\|_{\infty}^{2}\Bigr]
=
-\,\tfrac{\|g\|_{1}}{\lambda}\,\operatorname{sign}(g).
\label{eq:sign_direction}
\end{align}
The sign of the gradient determines the update direction, and the scalar $\|g\|_{1}/\lambda$ determines the update magnitude. The simplified AdamW update therefore matches steepest descent under the $\ell_{\infty}$ norm in direction, so applying AdamW to unlearning amounts to measuring the update size by the largest change in any single entry. The $\ell_{\infty}$ norm treats every entry independently and ignores the row and column structure of each weight matrix. The remaining task is to identify a norm appropriate for LLM unlearning.

\section{Spectral Unlearning: Transformer Structure-Preserving Updates}
\label{sec:method}

We develop structure-preserving unlearning updates for the matrix-valued parameters of a transformer. Specifically, we equip each weight matrix with a spectral-norm geometry and update it along the corresponding steepest-descent direction.

\subsection{Steepest descent under the matrix spectral norm}
\label{sec:vec_to_mat}

We now derive an update rule for matrix-valued parameters by applying steepest descent under the matrix spectral norm. The matrix spectral norm, also known as the operator $2$-norm, is defined as
\begin{align}
    \|W\|_2 := \max_{x \neq 0} \frac{\|Wx\|_2}{\|x\|_2} = \sup_{\|x\|_2 = 1} \|Wx\|_2,
\label{eq:spectral_norm_def}
\end{align}
and equals the largest singular value of $W$. In contrast to
the vector $\ell_\infty$ norm, which yields the entry-wise sign direction in \eqref{eq:sign_direction}, the spectral norm produces
an update aligned with the singular subspaces. Let $G_l$ denote the gradient of the unlearning objective with respect to the weight matrix $W_l$ at layer $l$.  We define the spectral steepest-descent step by
\begin{align}
    \Delta W_l^* \in \arg\min_{\Delta W_l} \left[ \langle G_l, \Delta W_l \rangle + \frac{\lambda}{2} \|\Delta W_l\|_2^2 \right],
\label{eq:layerwise_spectral_sd_problem}
\end{align}
where $\lambda > 0$ controls local curvature
and $\langle A, B \rangle := \sum_{i,j} A_{ij} B_{ij}$ denotes the Frobenius inner product. If $G_l = U_l \Sigma_l V_l^\top$ is the singular value decomposition (SVD) of the gradient, the solution is
\begin{align}
    \Delta W_l^* = -\frac{\operatorname{tr} \Sigma_l}{\lambda} U_l V_l^\top.
\label{eq:layerwise_spectral_sd_solution}
\end{align}
where $\operatorname{tr}\Sigma_l$ is the sum of singular values. This update admits a decomposition into magnitude and direction.
The magnitude $\operatorname{tr} \Sigma_l / \lambda$ scales with total singular value mass, while the direction $U_l V_l^\top$ is the SVD-aligned direction of $G_l$, retaining the orientation of the gradient on its singular subspaces.

For the iterative algorithm, let $G_{l,t} = U_{l,t} \Sigma_{l,t} V_{l,t}^\top$ be the gradient at step $t$. Translating the magnitude into the learning rate $\eta_l$ and retaining the spectral direction yields the update
\begin{align}
    W_{l,t+1} = W_{l,t} - \eta_l\, U_{l,t} V_{l,t}^\top, \qquad l = 1, \dots, L.
\label{eq:spectral_sd_update}
\end{align}
We refer to \eqref{eq:spectral_sd_update} as the spectral update and use it throughout the paper.
In LLM unlearning, this spectral update yields a direction that respects the joint structure of the entries of each weight matrix.

\subsection{Practical realizations of the spectral update}
\label{sec:kronecker_alignment}

The update in \eqref{eq:spectral_sd_update} defines the ideal spectral direction $U_{l,t} V_{l,t}^\top$. However, computing a full SVD at every layer and every step is prohibitively expensive at the LLM scale. In practice, spectral optimizers approximate this direction using efficient preconditioning schemes. We instantiate spectral unlearning inspired by optimizers that approximate the spectral direction through preconditioning.

\textbf{Muon} \cite{jordan2024muon} approximates the SVD-aligned direction $U_l V_l^\top$ using a Newton--Schulz iteration applied to a momentum-averaged gradient. Starting from $X_0 = M / \|M\|_F$ for momentum $M$, it iterates $X_{t+1} = \tfrac{3}{2} X_t - \tfrac{1}{2} X_t X_t^\top X_t$ for a small number of steps. %

\textbf{KLShampoo} \cite{lin2026understanding} maintains two running estimates $L_t$ and $R_t$ of the gradient second moments $\mathbb{E}[G G^\top]$ and $\mathbb{E}[G^\top G]$, derived via Kullback--Leibler (KL) minimization, and updates the weight as $\Delta W_t \propto -L_t^{-1/4}\, G_t\, R_t^{-1/4}$. These estimates capture the singular structure of $G_t$ from both sides.

\textbf{SOAP} \cite{vyas2025soap} maintains Shampoo-style preconditioners, but performs Adam-style updates in their eigenbasis. This more closely tracks the evolution of the gradient spectrum than standard Shampoo.

\begin{figure}[htbp]
    \centering
    \includegraphics[width=\textwidth]{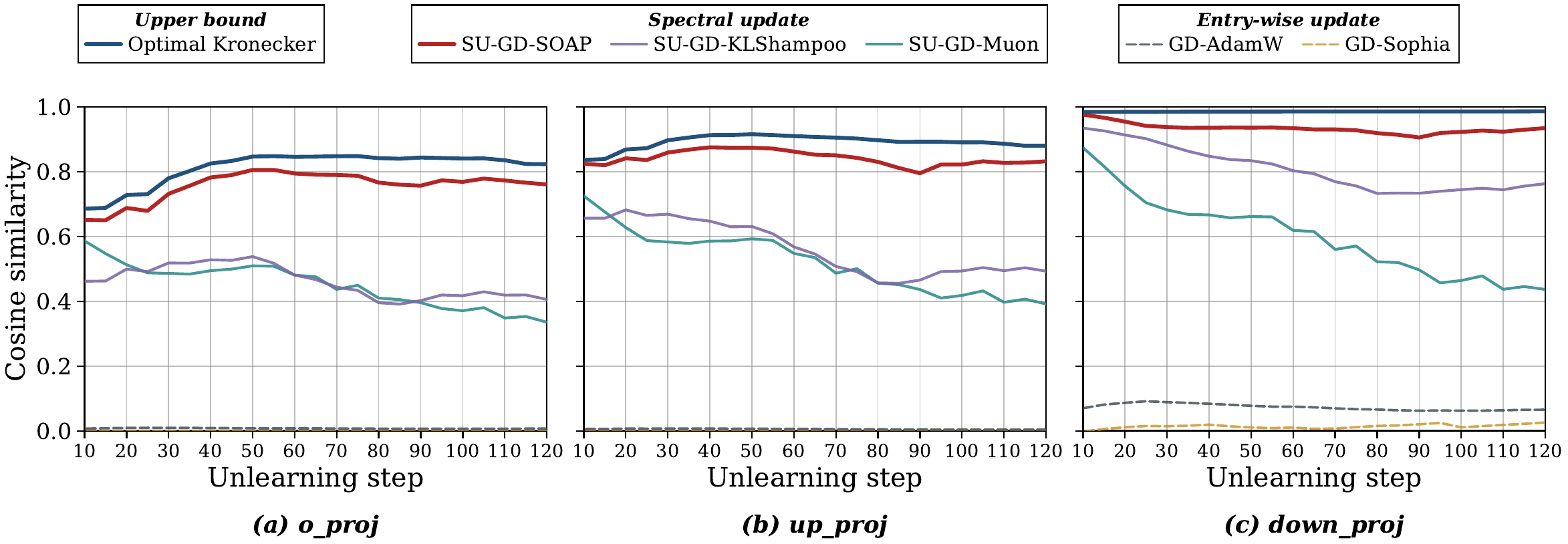}
    \caption{Cosine similarity between the gradient covariance estimated by update rule and the true covariance during GD unlearning on Llama-3.2-1B. Similarity is measured on the weight matrices of (a) the attention output projection, (b) the MLP up-projection, and (c) the MLP down-projection.}
    \label{fig:fig2}
\end{figure}

Fig.~\ref{fig:fig2} shows the cosine similarity between the gradient covariance estimated by each update rule and the true covariance on the three projection matrices of Llama-3.2-1B during GD unlearning. The reference approximates the full covariance by the optimal Kronecker product of two smaller matrices, one over the input dimension and one over the output dimension. A high reference score indicates that such a compact representation captures much of the measured covariance structure. SOAP attains scores closest to the reference. Muon and KLShampoo reach intermediate scores, while the entry-wise baselines remain near zero. The gap suggests that matrix-structured approximations capture gradient correlations across parameters that entry-wise statistics ignore.  Additional results are provided in Figs.~\ref{fig:gd_blocks7_15} and \ref{fig:npo_rmu_block1} of App.~\ref{app:extend_results}.

\begin{algorithm}[htbp]
\caption{Spectral unlearning with a matrix-structured update}
\label{alg:spectral_unlearning}
\begin{algorithmic}[1]
\Require Initial parameters $w_0$, forget set $\mathcal{D}_u$, retain set $\mathcal{D}_r$, unlearning objective $\mathcal{L}_{\mathrm{unlearn}}$, learning rate $\eta$, spectral update direction $\mathsf{OPT} \in \{$Muon, KLShampoo, SOAP$\}$, number of steps $S$
\State Initialize $w\leftarrow w_0$, optimizer states $\mathcal{S}_{l}$ for each weight matrix $W_l$
\For{$s=1,\dots,S$}
    \For{each weight matrix $W_l$}
        \State Compute $G_{l}=\nabla_{W_l}\mathcal{L}_{\mathrm{unlearn}}(w; \mathcal{D}_u, \mathcal{D}_r)$
        \State Update optimizer state $\mathcal{S}_{l} \leftarrow \mathsf{UpdateState}(\mathcal{S}_{l},G_{l})$
        \State Compute spectral update direction $P_{l} = \mathsf{OPT}(G_{l};\mathcal{S}_{l})$
        \State Update $W_l\leftarrow W_l-\eta P_{l}$
    \EndFor
\EndFor
\State \Return unlearned parameters $w$
\end{algorithmic}
\end{algorithm}

\subsection{Spectral unlearning algorithm}
\label{sec:spectral_algorithm}
Algorithm~\ref{alg:spectral_unlearning} summarizes spectral unlearning, where $\mathsf{UpdateState}$ holds optimizer states such as momentum and preconditioner statistics. The unlearning objective sets the gradient, and the update rule independently sets the approximation of the spectral direction. We apply the spectral update to attention and multi-layer perceptron (MLP) matrices and update embedding layers with AdamW. For matrices with only one RMS-normalized side, we apply one-sided projection. Further details on these adaptations are provided in App.~\ref{app:practical_realization}.

\section{Spectral Updates and Transformer Geometry}
\label{sec:theory}

\subsection{Pre-norm transformer architecture}
\label{sec:transformer_structure}
An LLM processes a sequence of $T$ tokens by embedding them as $d$-dimensional vectors and passing them through a stack of transformer layers.
At layer $l$, let $X_l = [x_{l,1}, \dots, x_{l,T}] \in \mathbb{R}^{d \times T}$ denote the $T$ token representations, where $x_{l,t} \in \mathbb{R}^{d}$ is the representation of token $t$. Let $\mathrm{RMSNorm}(\cdot)$ denote RMS normalization applied column-wise. A pre-norm transformer block can be written as
\begin{align}
H_l &= X_l + \mathrm{Attn}_l\bigl(\mathrm{RMSNorm}(X_l)\bigr), \label{eq:attn_block} \\
X_{l+1} &= H_l + \mathrm{MLP}_l\bigl(\mathrm{RMSNorm}(H_l)\bigr), \label{eq:ff_block}
\end{align}
where $H_l \in \mathbb{R}^{d \times T}$ is the residual stream after the attention sublayer. The attention sublayer $\mathrm{Attn}_l$ is parameterized by projection matrices $W_{Q,l}, W_{K,l}, W_{V,l}, W_{O,l} \in \mathbb{R}^{d \times d}$, and the feed-forward module $\mathrm{MLP}_l$ uses $W_{1,l} \in \mathbb{R}^{d_{\mathrm{ff}} \times d}$ and $W_{2,l} \in \mathbb{R}^{d \times d_{\mathrm{ff}}}$ with a pointwise activation.

Every weight matrix in \eqref{eq:attn_block}--\eqref{eq:ff_block} is applied to RMS-normalized inputs, and therefore operates on vectors with approximately unit RMS norm. This induces a geometry in which inputs are effectively normalized, while outputs remain controlled in scale. 
These observations motivate treating each weight matrix as a linear map equipped with its own matrix norm. %

\subsection{RMS-to-RMS operator norm}
\label{sec:rms_norm}

The pre-norm structure motivates a matrix-valued geometry. Each weight matrix acts on normalized inputs and produces outputs of bounded scale. We show that this RMS-to-RMS structure induces the spectral norm as the natural geometry for each weight matrix, and that the corresponding steepest-descent update recovers \eqref{eq:spectral_sd_update}.
For a vector $v \in \mathbb{R}^{d}$, the RMS norm is defined as
\begin{align}
\|v\|_{\mathrm{RMS}} := \frac{\|v\|_{2}}{\sqrt{d}}.
\label{eq:rms_norm}
\end{align}
As observed in Sec.~\ref{sec:transformer_structure}, each weight matrix in a pre-norm transformer maps RMS-normalized inputs to outputs of comparable scale. The corresponding operator norm that captures the largest output-to-input ratio under this geometry is the RMS-to-RMS norm.

To relate this geometry to standard matrix norms, we next characterize the RMS-to-RMS norm in terms of the spectral norm.
\begin{lemma}[RMS-to-RMS is a rescaled spectral norm]
\label{lem:rms_equals_spec}
For $W \in \mathbb{R}^{d_{\mathrm{out}} \times d_{\mathrm{in}}}$,
\begin{align}
\|W\|_{\mathrm{RMS} \to \mathrm{RMS}}:=
\max_{x \,:\, \|x\|_{\mathrm{RMS}} = 1}
\|Wx\|_{\mathrm{RMS}}
=
\sqrt{\tfrac{d_{\mathrm{in}}}{d_{\mathrm{out}}}}\, \|W\|_{2}.
\label{eq:rms_equals_spec}
\end{align}
\end{lemma}
The proof is deferred to App.~\ref{app:rms_equals_spec}. Lemma~\ref{lem:rms_equals_spec} shows that  spectral norm is the natural induced norm.

\subsection{Steepest descent under transformer geometry}
\label{sec:rms_steepest}
We now extend the steepest descent method under the RMS-to-RMS geometry.
\begin{theorem}[Spectral descent under the transformer-native norm]
\label{thm:spectral_matches}
Let $G_l \in \mathbb{R}^{d_{\mathrm{out},l}\times d_{\mathrm{in},l}}$ denote the gradient of the unlearning objective at layer $l$, with reduced SVD $G_l = U_l\Sigma_l V_l^\top$. Under the layerwise-maximum RMS-to-RMS network norm, a steepest-descent update at layer $l$ is
\begin{align}
    \Delta W_l^* = -\eta_l U_lV_l^\top,
    \qquad
    \eta_l = \tfrac{1}{\lambda}\Bigl(\sum_{k=1}^{L}
\sqrt{\tfrac{d_{\mathrm{out},k}}{d_{\mathrm{in},k}}}
\operatorname{tr}\Sigma_k\Bigr) \sqrt{\tfrac{d_{\mathrm{out},l}}{d_{\mathrm{in},l}}}.
\end{align}
\end{theorem}
The proof is given in App.~\ref{app:multilayer}. %
Theorem~\ref{thm:spectral_matches} shows that the spectral update rule is the steepest-descent update under the norm induced by the transformer architecture. The matrix norm in \eqref{eq:layerwise_spectral_sd_problem} is not chosen for algorithmic convenience but is the geometry intrinsic to the transformer architecture. Since RMS-normalized weight matrices in transformers satisfy $d_{\mathrm{out},l} = d_{\mathrm{in},l}$, the layer-wise factor in $\eta_l$ vanishes, and a single scalar learning rate suffices. In the same spirit, the same scalar learning rate is applied to one-sided projections in Sec.~\ref{sec:spectral_algorithm}.

\section{Experiments}
\label{sec:exp}
\subsection{Experimental Setup}

\paragraph{Benchmarks and models.} 
We evaluate on TOFU using Llama-3.2-1B and 3B-Instruct~\cite{llama3herd2024technical}. Following TOFU~\cite{maini2024tofu}, we construct two references: an \textit{original model} fine-tuned on the full dataset, and a \textit{retrain model} fine-tuned only on the retain set. The retrain model serves as the target for successful unlearning, as it reflects the behavior of a model that has never seen the forget set.
We additionally evaluate on RWKU~\cite{NEURIPS2024_b1f78dfc} and WMDP~\cite{pmlr-v235-li24bc} benchmark using Llama-3-8B-Instruct~\cite{llama3herd2024technical} and Zephyr-7B-$\beta$~\cite{tunstall2024zephyr}, respectively.  Full hyperparameter details are provided in App.~\ref{app:hyper} and \ref{app:metric_details}.

\paragraph{Metrics.} 
Following the OpenUnlearning evaluation protocol \cite{dorna2026openunlearning}, we report an aggregate score (Agg), computed as the harmonic mean of memorization (Mem), privacy (Priv), and utility scores (Util). Each component score is itself computed as the harmonic mean of representative sub-metrics. This aggregation penalizes methods that improve one axis at the expense of another, making Agg a conservative measure of balanced unlearning performance. All results are averaged over three random seeds, with $10$ training epochs. Wall-clock-matched results are detailed in Fig. \ref{fig:traj_layout_A_4x3} of App.~\ref{app:wall-clock}.

\paragraph{Method naming.}
We denote each method by (PREFIX-)LOSS-OPT, where PREFIX, if it exists, identifies the spectral unlearning (SU) variant, LOSS specifies the unlearning objective, and OPT denotes the update rules. As baselines, we consider two entry-wise updates: \textbf{AdamW}~\cite{maini2024tofu, zhang2024negative, pmlr-v235-li24bc}, which normalizes each coordinate using an exponential moving average, %
 and \textbf{Sophia}~\cite{jia-etal-2024-soul}, which uses a diagonal Hessian estimate based on Hutchinson's stochastic trace approximation.

\begin{table*}[t]
\centering
\footnotesize
\setlength{\tabcolsep}{3.2pt}
\renewcommand{\arraystretch}{1.10}
\setlength{\heavyrulewidth}{0.9pt}
\setlength{\lightrulewidth}{0.5pt}
\caption{Unlearning results on TOFU with Llama-3.2-1B and 3B at forget splits of 5\% and 10\%.}
\label{tab:paper_main}
\resizebox{\textwidth}{!}{
\begin{tabular}{l >{\columncolor{blue!3}}c ccc >{\columncolor{blue!3}}c ccc >{\columncolor{blue!3}}c ccc >{\columncolor{blue!3}}c ccc}
\toprule
\textbf{Model} & \multicolumn{8}{c}{\textbf{Llama-3.2-1B-Instruct}} & \multicolumn{8}{c}{\textbf{Llama-3.2-3B-Instruct}} \\
\cmidrule(lr){2-9} \cmidrule(lr){10-17}
\textbf{Forget split} & \multicolumn{4}{c}{Forget 10\%} & \multicolumn{4}{c}{Forget 5\%} & \multicolumn{4}{c}{Forget 10\%} & \multicolumn{4}{c}{Forget 5\%} \\
\cmidrule(lr){2-5} \cmidrule(lr){6-9} \cmidrule(lr){10-13} \cmidrule(lr){14-17}
\textbf{Unlearning method} & \textbf{Agg} $\uparrow$ & Mem $\uparrow$ & Priv $\uparrow$ & Util $\uparrow$ & \textbf{Agg} $\uparrow$ & Mem $\uparrow$ & Priv $\uparrow$ & Util $\uparrow$ & \textbf{Agg} $\uparrow$ & Mem $\uparrow$ & Priv $\uparrow$ & Util $\uparrow$ & \textbf{Agg} $\uparrow$ & Mem $\uparrow$ & Priv $\uparrow$ & Util $\uparrow$ \\
\midrule
\rowcolor{gray!8} \textit{Original (init)} & 0.18 & 0.08 & 0.38 & 0.61 & 0.17 & 0.08 & 0.36 & 0.61 & 0.08 & 0.03 & 0.39 & 0.67 & 0.07 & 0.03 & 0.36 & 0.67 \\
\rowcolor{gray!8} \textit{Retrain (target)} & 0.72 & 0.65 & 1.00 & 0.61 & 0.72 & 0.64 & 1.00 & 0.62 & 0.74 & 0.64 & 1.00 & 0.67 & 0.74 & 0.64 & 1.00 & 0.68 \\
\midrule
GD\text{-}AdamW & 0.57 & 0.56 & 0.78 & 0.45 & 0.55 & 0.53 & 0.67 & 0.48 & 0.38 & 0.25 & 0.43 & 0.63 & 0.54 & 0.62 & 0.89 & 0.36 \\
GD\text{-}Sophia & 0.56 & 0.55 & 0.72 & 0.46 & 0.57 & 0.64 & 0.77 & 0.41 & 0.58 & 0.57 & 0.81 & 0.47 & 0.57 & 0.59 & 0.90 & 0.42 \\
SU\text{-}GD\text{-}Muon & 0.58 & 0.64 & 0.87 & 0.41 & 0.60 & 0.62 & 0.89 & 0.44 & 0.50 & 0.42 & 0.53 & 0.58 & 0.56 & 0.61 & 0.87 & 0.39 \\
SU\text{-}GD\text{-}KLShampoo & 0.59 & 0.63 & 0.88 & 0.42 & 0.60 & 0.63 & 0.87 & 0.44 & 0.51 & 0.43 & 0.55 & 0.58 & 0.56 & 0.61 & 0.88 & 0.39 \\
\addlinespace[2pt]
\rowcolor{red!6} SU\text{-}GD\text{-}SOAP & 0.60 & 0.61 & 0.89 & 0.45 & 0.61 & 0.61 & 0.87 & 0.46 & 0.61 & 0.61 & 0.87 & 0.48 & 0.65 & 0.57 & 0.94 & 0.55 \\
\midrule
NPO\text{-}AdamW & 0.69 & 0.66 & 0.87 & 0.59 & 0.54 & 0.56 & 0.61 & 0.48 & 0.72 & 0.66 & 0.89 & 0.66 & 0.61 & 0.58 & 0.65 & 0.60 \\
NPO\text{-}Sophia & 0.68 & 0.66 & 0.83 & 0.58 & 0.66 & 0.68 & 0.80 & 0.55 & 0.72 & 0.64 & 0.95 & 0.65 & 0.71 & 0.67 & 0.83 & 0.65 \\
SU\text{-}NPO\text{-}Muon & 0.69 & 0.65 & 0.92 & 0.57 & 0.67 & 0.66 & 0.83 & 0.56 & 0.72 & 0.67 & 0.87 & 0.65 & 0.71 & 0.64 & 0.94 & 0.61 \\
SU\text{-}NPO\text{-}KLShampoo & 0.68 & 0.66 & 0.89 & 0.57 & 0.67 & 0.68 & 0.80 & 0.56 & 0.72 & 0.67 & 0.87 & 0.65 & 0.70 & 0.65 & 0.92 & 0.61 \\
\addlinespace[2pt]
\rowcolor{red!6} SU\text{-}NPO\text{-}SOAP & 0.69 & 0.66 & 0.91 & 0.58 & 0.67 & 0.69 & 0.81 & 0.56 & \textbf{0.73} & 0.65 & 0.92 & 0.66 & 0.72 & 0.68 & 0.86 & 0.65 \\
\midrule
RMU\text{-}AdamW & 0.69 & 0.68 & 0.90 & 0.56 & 0.65 & 0.81 & 0.75 & 0.48 & 0.70 & 0.79 & 0.71 & 0.63 & 0.72 & 0.84 & 0.69 & 0.64 \\
RMU\text{-}Sophia & 0.56 & 0.84 & 0.71 & 0.36 & 0.68 & 0.76 & 0.81 & 0.54 & 0.69 & 0.84 & 0.66 & 0.59 & 0.73 & 0.69 & 0.90 & 0.65 \\
SU\text{-}RMU\text{-}Muon & 0.67 & 0.66 & 0.90 & 0.55 & 0.63 & 0.63 & 0.77 & 0.54 & 0.71 & 0.77 & 0.73 & 0.64 & 0.72 & 0.75 & 0.77 & 0.65 \\
SU\text{-}RMU\text{-}KLShampoo & 0.69 & 0.67 & 0.95 & 0.55 & 0.66 & 0.65 & 0.85 & 0.55 & 0.71 & 0.78 & 0.72 & 0.64 & 0.72 & 0.86 & 0.69 & 0.65 \\
\addlinespace[2pt]
\rowcolor{red!6} SU\text{-}RMU\text{-}SOAP & \textbf{0.70} & 0.71 & 0.87 & 0.58 & \textbf{0.69} & 0.80 & 0.76 & 0.55 & 0.72 & 0.76 & 0.75 & 0.65 & \textbf{0.74} & 0.72 & 0.85 & 0.67 \\
\bottomrule
\end{tabular}
}
\end{table*}

\subsection{Main Results on TOFU Benchmark}
Tab.~\ref{tab:paper_main} reports TOFU results for entry-wise updates (AdamW and Sophia) and spectral updates (Muon, KLShampoo, and SOAP). Across unlearning objectives and model scales, the spectral variants consistently improve the aggregate score over their entry-wise counterparts, and SOAP achieves the strongest overall performance. This is consistent with SOAP's role as the closest approximation to the spectral direction as reported in Fig.~\ref{fig:fig2}. Sophia approximates curvature through its diagonal Hessian estimate and outperforms AdamW. Nevertheless, it cannot capture inter-parameter correlations within a weight matrix. On Llama-3.2-3B at the 5\% forget split, SU-RMU-SOAP matches the retain-trained reference, rendering it indistinguishable from the target model under our metrics.

The three unlearning objectives exhibit distinct characteristics that interact with the choice of update rule. RMU steers intermediate representations toward a random target direction, which improves memorization removal and leads to strong aggregate performance.
NPO anchors the model to its pre-unlearning reference, yielding strong privacy scores. As the objective is bounded, it is stable but slow to converge. Consequently, on the 5\% forget split, NPO-AdamW does not reach the retrain target within the same epoch budget, whereas spectral updates close this gap quickly in Fig.~\ref{fig:fig1}. GD is less stable than NPO and RMU, since its forget loss can diverge under aggressive updates. On Llama-3.2-3B at the 10\% forget split, GD-AdamW fails to forget, whereas the spectral variants remain effective in the same regime. Across every model size, forget split, and loss family, SU-SOAP exceeds
\begin{wrapfigure}[14]{r}{0.5\textwidth}
  \vspace{-1.2em}
  \includegraphics[width=\linewidth]{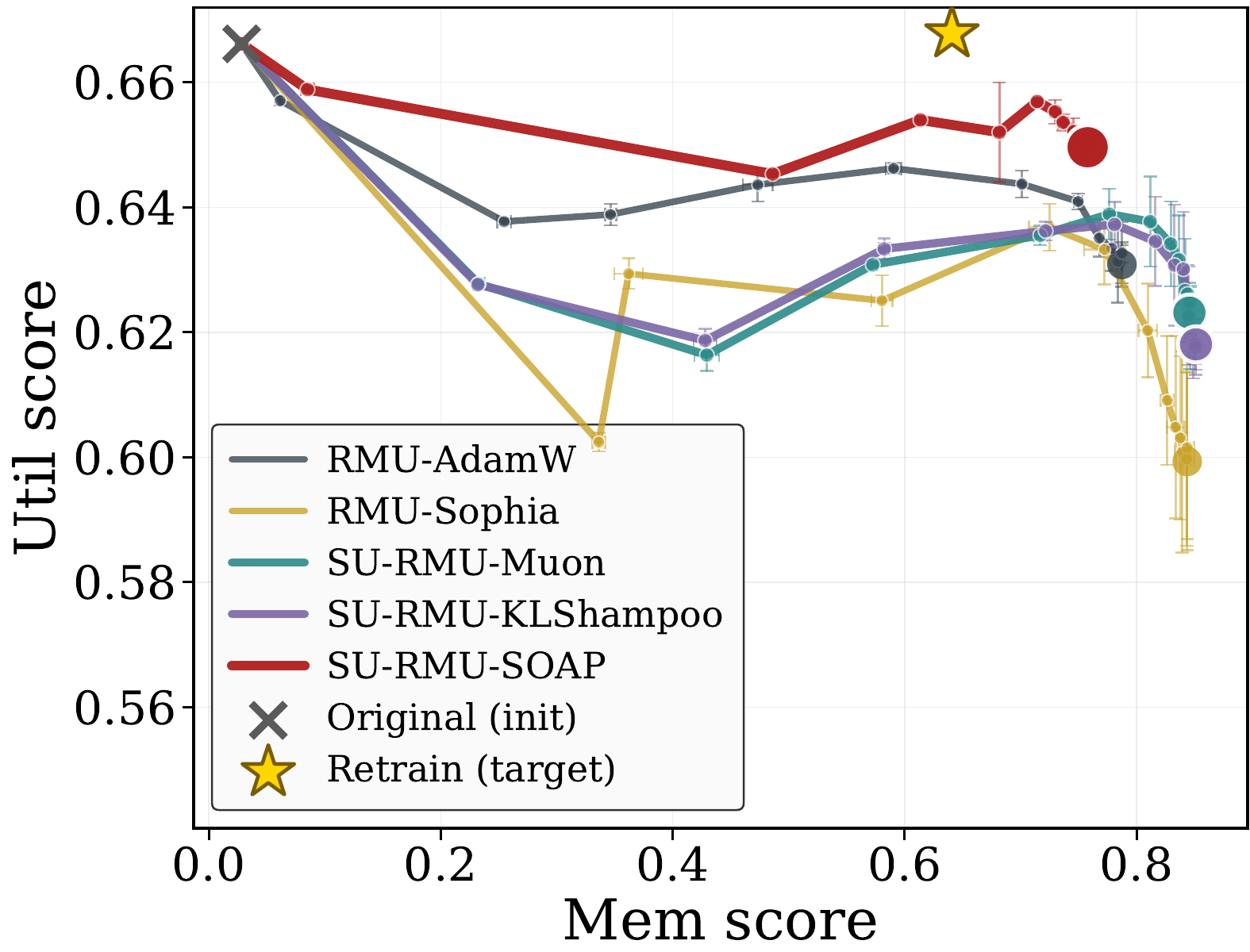}
    \vspace{-1em}
    \caption{Util-Mem score Trade-off.}
    \label{fig:fig3}
\end{wrapfigure}
the aggregate score of both AdamW and Sophia. Additional results, including the forget 1\% split, variance, and nine further baselines, are provided in App.~\ref{app:Detail_metric}.

Fig.~\ref{fig:fig3} visualizes the utility--memorization trade-off on Llama-3.2-3B at the 10\% forget split. As unlearning progresses,  trajectories move toward the lower-right corner: forgetting improves, but at the cost of utility. This trade-off makes it essential to evaluate both axes jointly rather than in isolation. Among the methods, SU-RMU-SOAP traces the most favorable  trajectory and ends closest to the retain-trained target (the yellow star), placing it nearest to the Pareto frontier.

\begin{wraptable}{r}{0.5\textwidth}
\vspace{-1em}
\centering
\small
 \setlength{\tabcolsep}{4pt}
 \renewcommand{\arraystretch}{1.12}
 \setlength{\heavyrulewidth}{0.9pt}
 \setlength{\lightrulewidth}{0.5pt}
\caption{Sub-metric comparison between spectral and entry-wise update families.}
\label{tab:family_winrate_compact}
\resizebox{\linewidth}{!}{%
\begin{tabular}{@{}l c c c c@{}}
\toprule
\textbf{Sub-metric} & \textbf{Entry-wise} & \textbf{Spectral} & \textbf{Difference} & \textbf{Wins} \\
\midrule
\rowcolor{blue!5}\textbf{Forget truth ratio $\uparrow$} & $0.5756$ & $0.6059$ & \textcolor{green!55!black}{$+0.0303$} & \textbf{18/18} \\
\rowcolor{blue!5}Forget probability $\downarrow$ & $0.1585$ & $0.0859$ & \textcolor{green!55!black}{$-0.0727$} & 14/18 \\
\rowcolor{blue!5}Exact memorization $\downarrow$ & $0.6202$ & $0.5677$ & \textcolor{green!55!black}{$-0.0525$} & 13/18 \\
\rowcolor{blue!5}Extraction strength $\downarrow$ & $0.0900$ & $0.0687$ & \textcolor{green!55!black}{$-0.0213$} & 12/18 \\
\addlinespace[1.5pt]
\rowcolor{orange!5}Retain ROUGE $\uparrow$ & $0.5632$ & $0.5912$ & \textcolor{green!55!black}{$+0.0280$} & 12/18 \\
\rowcolor{orange!5}Retain probability $\uparrow$ & $0.6214$ & $0.6598$ & \textcolor{green!55!black}{$+0.0384$} & 11/18 \\
\rowcolor{orange!5}Retain truth ratio $\uparrow$ & $0.5295$ & $0.5345$ & \textcolor{green!55!black}{$+0.0050$} & 11/18 \\
\bottomrule
\end{tabular}
}
\vspace{-0.7em}
\end{wraptable}

\paragraph{Sub-metric Breakdown} To better understand the gains in Tab.~\ref{tab:paper_main}, we examine the sub-metric win rates in Tab.~\ref{tab:family_winrate_compact}. The largest improvements from spectral updates occur on both the forget and retain sides. In particular, spectral variants show clear gains in forget truth ratio and forget probability, indicating more effective removal of the target data, as well as in retain ROUGE and retain probability, indicating better preservation of retained knowledge. In contrast, spectral and entry-wise updates perform similarly on the out-of-distribution score. Detailed sub-metrics are reported in Tab.~\ref{tab:family_winrate_raw} of App.~\ref{app:Sub-metric_comparison}.

\begin{figure}[htbp]
    \centering
    \includegraphics[width=0.92\textwidth]{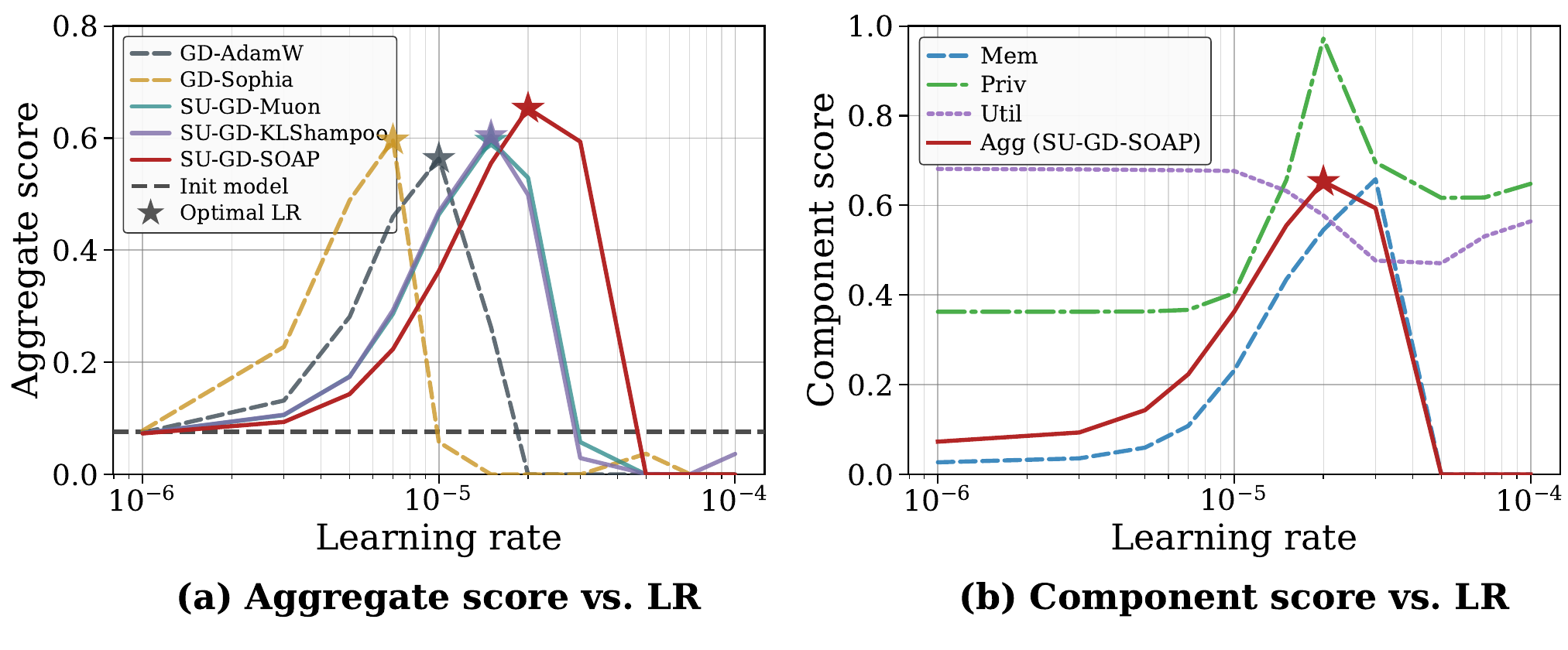}
    \vspace{-0.5em}
    \caption{Learning rate sensitivity on GD unlearning. (a) Aggregate score across update rules. Stars mark the optimal LR. (b) Agg score broken down into Mem, Priv, and Util for SU-GD-SOAP.}
    \label{fig:fig4}
\end{figure}

\subsection{Learning-rate Sensitivity}
Fig.~\ref{fig:fig4}(a) analyzes sensitivity of the aggregate score to the learning rate using GD on the 3B model with a 5\% forget ratio. Learning rate critically controls the trade-off: small values fail to propagate the forget signal, while large values collapse retention. Spectral updates achieve higher peak aggregate scores and remain effective over a wider range than entry-wise updates, with SOAP exhibiting the strongest robustness. This robustness is practically important, as the optimal learning rate depends on the forget-set size, model scale, and objective, and cannot always be tuned reliably. Additional results and normalized analyses are provided in Fig. \ref{fig:paper_lrsens} of App. \ref{app:extend_results}.

Fig.~\ref{fig:fig4}(b) decomposes the aggregate score of SU-GD-SOAP into memorization, privacy, and utility. Memorization increases with learning rate but drops sharply beyond a threshold due to the forget truth ratio: at high learning rates, the model assigns high probability to incorrect answers, and the harmonic-mean aggregation amplifies this failure. Privacy peaks at intermediate learning rates, reflecting a balance between insufficient forgetting and excessive drift. Utility decreases with increasing learning rate but may recover slightly at very high values; this recovery is misleading, as extreme learning rates induce degenerate outputs (e.g., repeated ``I don’t know''), which are weakly penalized by fluency metrics.

\begin{figure}[htbp]
    \centering
    \includegraphics[width=\textwidth]{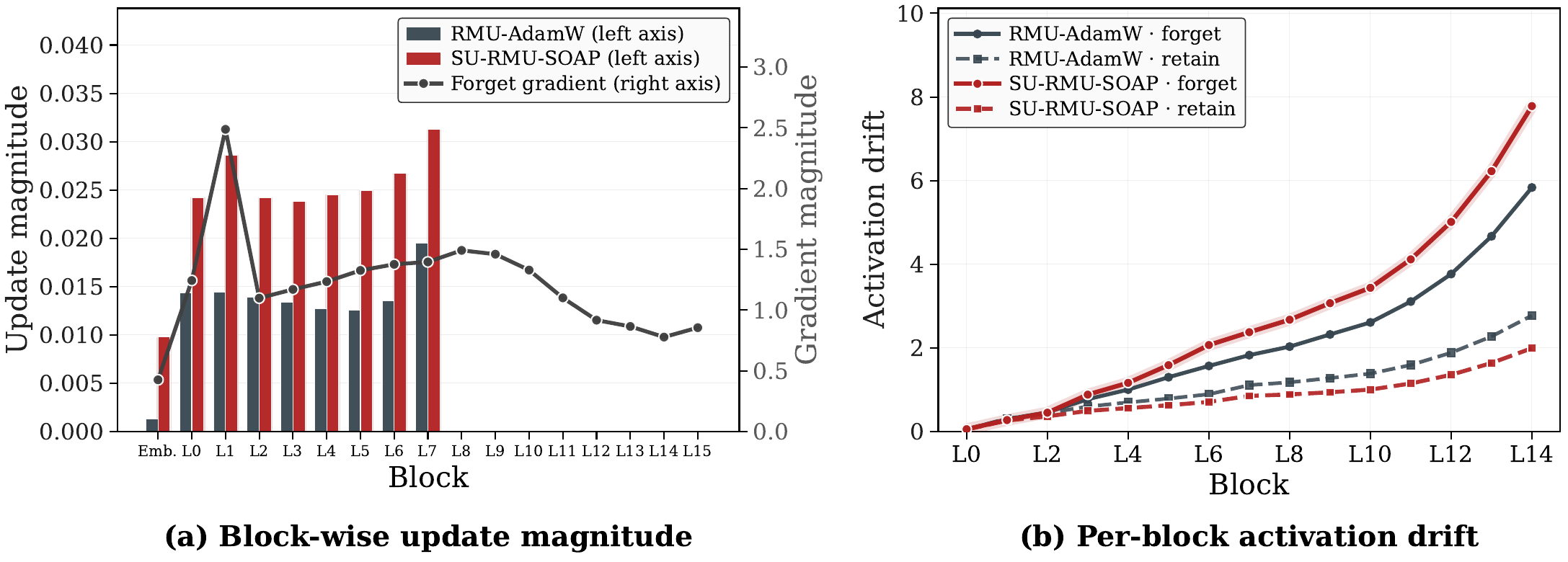}
    \caption{Block-wise comparison of RMU-AdamW and SU-RMU-SOAP. (a) Update magnitude (bars) and forget gradient magnitude (line). (b) Activation drift on forget and retain sets.}
    \label{fig:fig5}
\end{figure}

\subsection{Block-wise Analysis of RMU}
\label{sec:RMU_analysis}

Fig.~\ref{fig:fig5}(a) analyzes how spectral unlearning redistributes updates across transformer blocks in RMU. For each block, we measure the average per-step update magnitude, defined as the mean Frobenius norm of weight changes across unlearning steps. As a reference, we plot the gradient disagreement between the forget and retain objectives at initialization, which indicates where selective parameter changes are most needed. 
RMU with AdamW assigns relatively little update mass to early and mid blocks where disagreement is largest, whereas SU-RMU-SOAP allocates larger updates to these regions--especially the L1 block--yielding an update profile that more closely tracks the disagreement curve.

Fig.~\ref{fig:fig5}(b) shows the resulting activation drift. We measure drift as the average $\ell_2$ distance between hidden states of the unlearned and original models. SU-RMU-SOAP induces larger drift on forget examples while keeping retain drift close to RMU-AdamW, producing a wider separation in representation space. This suggests that spectral preconditioning improves selectivity by concentrating updates where the forget and retain objectives disagree most strongly.

\subsection{Results on RWKU and WMDP Benchmark}

\begin{table*}[t]
\centering
\small
\setlength{\aboverulesep}{0pt}
\setlength{\belowrulesep}{0pt}
\setlength{\tabcolsep}{1.5pt}
\renewcommand{\arraystretch}{1.15}
\setlength{\heavyrulewidth}{0.9pt}
\setlength{\lightrulewidth}{0.5pt}
\caption{Performance Comparison on RWKU and WMDP Benchmarks.}
\label{tab:wmdp_performance}
\begin{tabular}{ll cccc cccc}
\toprule
\multirow{2}{*}{{Method}} & \multirow{2}{*}{\shortstack[c]{Update\\Rule}} & \multicolumn{4}{c}{\textbf{RWKU (Llama-3-8B-Instruct)}} & \multicolumn{4}{c}{\textbf{WMDP (Zephyr-7B-$\beta$)}} \\
\cmidrule(lr){3-6} \cmidrule(lr){7-10}
 & & {Forget} $\downarrow$ & {Neighbor} $\uparrow$ & MMLU $\uparrow$ & Gain $\Delta$ $\uparrow$ & {Bio} $\downarrow$ & {Cyber} $\downarrow$ & {MMLU} $\uparrow$ & Gain $\Delta$ $\uparrow$ \\
\midrule
\rowcolor{gray!8} \multicolumn{2}{l}{\textit{Original (init)}} & 0.6905 & 0.8190 & 0.6287 & 0.0000 & 0.6387 & 0.4268 & 0.5879 & 0.0000 \\
\midrule
 & Entry-wise & 0.1127 & 0.5657 & \textbf{0.6392} & 0.1117 & 0.2916 & 0.2574 & 0.4527 & 0.1271 \\
\rowcolor{pink!15} \multirow{-2}{*}{GD} & Spectral & 0.1460 & 0.6539 & 0.6376 & 0.1295 & \textbf{0.2789} & \textbf{0.2473} & 0.5084 & 0.1533 \\
\cmidrule(lr){1-2} \cmidrule(lr){3-6} \cmidrule(lr){7-10}
 & Entry-wise & 0.3250 & \textbf{0.7520} & 0.6292 & 0.0997 & 0.5124 & 0.3376 & 0.5421 & 0.0566 \\
\rowcolor{pink!15} \multirow{-2}{*}{NPO} & Spectral & 0.2034 & 0.7334 & 0.6281 & \textbf{0.1336} & 0.4086 & 0.3182 & \textbf{0.5578} & 0.1029 \\
\cmidrule(lr){1-2} \cmidrule(lr){3-6} \cmidrule(lr){7-10}
 & Entry-wise & 0.1038 & 0.4557 & 0.6000 & 0.0649 & 0.3079 & 0.2723 & 0.5387 & 0.1454 \\
\rowcolor{pink!15} \multirow{-2}{*}{RMU} & Spectral & \textbf{0.0923} & 0.5880 & 0.6086 & 0.1157 & 0.3028 & 0.2581 & 0.5476 & \textbf{0.1548} \\
\bottomrule
\end{tabular}
\end{table*}

Tab.~\ref{tab:wmdp_performance} reports results on RWKU~\cite{NEURIPS2024_b1f78dfc} and WMDP~\cite{pmlr-v235-li24bc}. Entry-wise variants indicate AdamW, and spectral variants indicate SOAP. RWKU unlearns knowledge about real-world famous people, and WMDP unlearns hazardous knowledge in biosecurity and cybersecurity. Both benchmarks evaluate forgetting together with retention. Forget on RWKU and bio and cyber on WMDP measure the remaining target knowledge. Neighbor on RWKU measures knowledge closely related to the target, which unlearning should leave intact. MMLU measures general ability on both benchmarks. Since WMDP is a four-way multiple-choice benchmark, bio and cyber scores closer to $0.25$ indicate more thorough forgetting. Gain $\Delta$ is the average improvement over the original model across the three metrics, with each metric signed by its desired direction. For every unlearning method, spectral updates achieve a higher gain $\Delta$ than entry-wise updates on both benchmarks.

\subsection{Robustness to relearning}
\begin{wrapfigure}[14]{r}{0.5\textwidth}
  \vspace{-1.2em}
  \includegraphics[width=\linewidth]{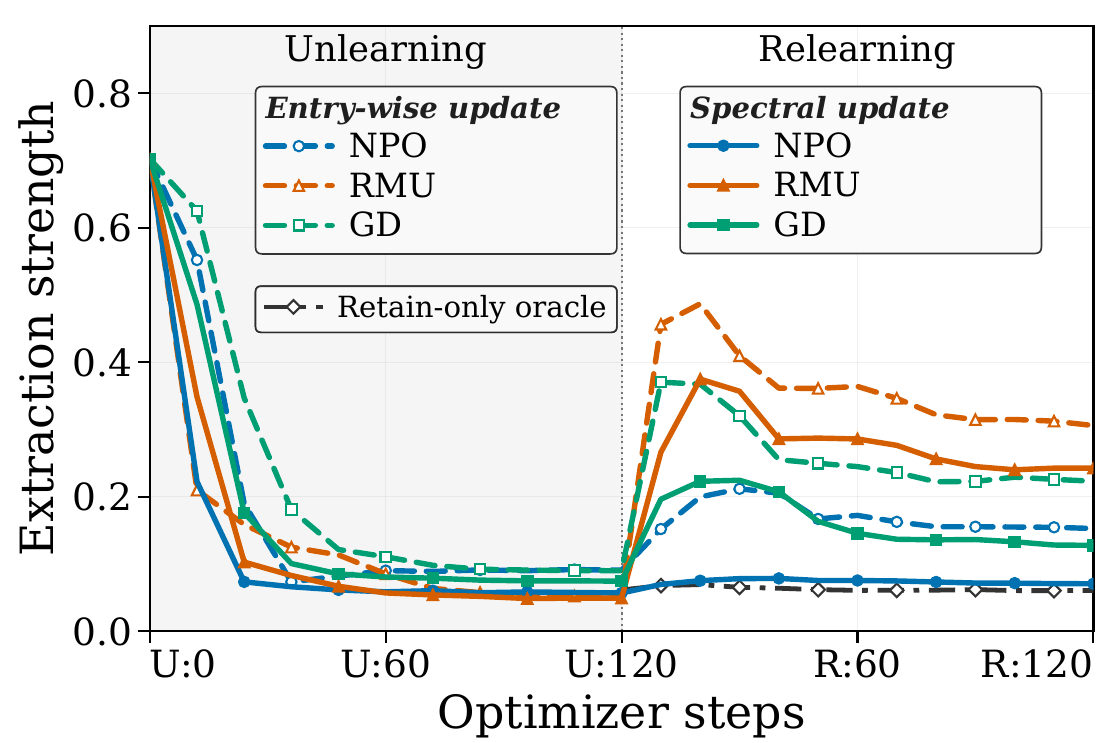}
    \vspace{-1.5em}
    \caption{Knowledge recovery under relearning.}
    \label{fig:extraction_strength}
\end{wrapfigure}

A relearning attack recovers the removed content by fine-tuning a small subset of the forget set~\cite{hu2025unlearning}. Each model first unlearns a forget set covering 10\% of the full dataset. The unlearned model is then fine-tuned on a random 10\% subset of the forget set, i.e., 1\% of the full dataset. Entry-wise variants use AdamW, and spectral variants use SOAP. Fig.~\ref{fig:extraction_strength} shows extraction strength throughout relearning. Extraction strength is high when a short prefix of an answer suffices to regenerate the rest~\cite{yoon2026rethinking}. The retain-only oracle keeps extraction strength low throughout relearning and serves as the target for unlearning. For every unlearning objective, the spectral variant ends relearning with lower extraction strength than the entry-wise variant. RMU unlearns by steering the hidden states of a single layer. Under RMU, extraction strength rebounds more sharply with entry-wise updates. Sec.~\ref{sec:RMU_analysis} shows that SU-RMU-SOAP moves forget-set representations farther from the original ones than RMU-AdamW does. A farther shift may require more fine-tuning to undo, so the original responses remain harder to recover with limited relearning. NPO yields the lowest final extraction strength under both update rules, and spectral NPO stays close to the oracle throughout relearning. The results suggest that matrix-structured updates make the forget data harder to extract again through relearning.

\section{Conclusion}
We presented spectral unlearning, a loss-agnostic framework that improves LLM unlearning by changing the geometry of the parameter update. Treating transformer weights as matrix-valued maps between RMS-normalized feature spaces, we show that the transformer-native RMS-to-RMS norm reduces to a rescaled spectral norm and induces a polar-factor steepest-descent direction. This provides a geometric justification for replacing entry-wise updates with spectral updates in existing unlearning methods. Across TOFU, RWKU, and WMDP, spectral updates consistently improve the memorization--privacy--utility trade-off and move the unlearned model closer to the retain-trained reference, highlighting update geometry as a simple yet underexplored axis of LLM unlearning.

\begin{ack}
This work is supported in part by the NRF grants funded by the Korea government(MSIT) (RS-2026-25488671, 33\% and RS-2022-NR070834, 33\%), and in part by ICT Creative Consilience Program through the IITP grant funded by the Korea government(MSIT)(IITP-2026-RS-2020-II201819, 33\%).
\end{ack}

\bibliographystyle{unsrt_etal3} %

\bibliography{references.bib}       %

\clearpage
\appendix

\section{Proofs}
\label{app:proofs}

This appendix proofs Lemma~\ref{lem:rms_equals_spec} and Theorem~\ref{thm:spectral_matches}. We proceed in three steps. First, we derive a general closed form for steepest descent under any norm via duality. Second, we identify the RMS-to-RMS norm as a rescaled spectral norm and compute its dual norm and duality map. Third, we combine these ingredients to prove the multi-layer result.

\subsection{Steepest descent via duality}
\label{app:duality}

Steepest descent under any norm reduces to two scalar quantities: the dual norm and the duality map. We make this precise here.

Let $\|\cdot\|$ be a norm on $\mathbb{R}^{n}$. The \emph{dual norm} of $g \in \mathbb{R}^{n}$ measures the largest possible inner product with $g$ over the unit ball of $\|\cdot\|$,
\begin{align}
\|g\|_{*} := \max_{t \in \mathbb{R}^{n}\,:\,\|t\| = 1} g^{\top} t,
\label{eq:app_dual_norm}
\end{align}
and the \emph{duality map} returns a unit vector achieving the maximum,
\begin{align}
\operatorname{dualize}_{\|\cdot\|}(g) := \arg\max_{\|t\|=1} g^{\top} t.
\label{eq:app_duality_map}
\end{align}
Intuitively, the duality map is the unit-norm direction most aligned with $g$; the dual norm is the inner product achieved along that direction.

\begin{lemma}[Steepest descent, closed form]
\label{lem:steepest_closed_form}
For any $g \in \mathbb{R}^{n}$ and $\lambda > 0$,
\begin{align}
\arg\min_{\Delta w \in \mathbb{R}^{n}}
\Bigl[g^{\top} \Delta w + \tfrac{\lambda}{2} \|\Delta w\|^{2}\Bigr]
=
-\,\frac{\|g\|_{*}}{\lambda}\, \operatorname{dualize}_{\|\cdot\|}(g).
\label{eq:app_steepest_dual}
\end{align}
\end{lemma}

\begin{proof}
Decompose $\Delta w = c \cdot t$ into magnitude $c \geq 0$ and direction $t$ with $\|t\| = 1$. The objective becomes
\begin{align}
c \cdot g^{\top} t + \tfrac{\lambda}{2} c^{2}.
\end{align}
With $c$ fixed, only the linear term $c\, g^{\top} t$ depends on $t$. Since $c \geq 0$, this is minimized by choosing $t$ to make $g^{\top} t$ as negative as possible, that is,
\begin{align}
t^{\star} = -\operatorname{dualize}_{\|\cdot\|}(g),
\end{align}
which gives $c \cdot g^{\top} t^{\star} = -c \cdot \|g\|_{*}$. Substituting back, the problem reduces to minimizing
\begin{align}
-c \cdot \|g\|_{*} + \tfrac{\lambda}{2} c^{2}
\end{align}
over $c \geq 0$. This is a convex quadratic, minimized at $c^{\star} = \|g\|_{*} / \lambda$. Combining,
\begin{align}
\Delta w^{\star} = c^{\star} \cdot t^{\star} = -\frac{\|g\|_{*}}{\lambda}\, \operatorname{dualize}_{\|\cdot\|}(g).
\end{align}
\end{proof}

Lemma~\ref{lem:steepest_closed_form} is the workhorse of the appendix. Specializing to a particular norm requires only computing $\|g\|_{*}$ and $\operatorname{dualize}_{\|\cdot\|}(g)$ for that norm. The next two sections do this for the RMS-to-RMS norm.

\subsection{The RMS-to-RMS norm as a rescaled spectral norm}
\label{app:rms_equals_spec}

The RMS-to-RMS norm differs from the spectral norm only in how it measures input and output magnitudes: $\|\cdot\|_{\mathrm{RMS}}$ divides by $\sqrt{d}$, whereas $\|\cdot\|_{2}$ does not. The two norms therefore differ by a ratio of the input and output dimensions.

\begin{proof}[Proof of Lemma~\ref{lem:rms_equals_spec}]
The constraint $\|x\|_{\mathrm{RMS}} = 1$ is equivalent to $\|x\|_{2} = \sqrt{d_{\mathrm{in}}}$, and $\|Wx\|_{\mathrm{RMS}} = \|Wx\|_{2} / \sqrt{d_{\mathrm{out}}}$. Substituting both into the RMS-to-RMS operator norm,
\begin{align}
\|W\|_{\mathrm{RMS} \to \mathrm{RMS}}
=
\max_{\|x\|_{\mathrm{RMS}} = 1} \|Wx\|_{\mathrm{RMS}}
=
\frac{1}{\sqrt{d_{\mathrm{out}}}} \max_{\|x\|_{2} = \sqrt{d_{\mathrm{in}}}} \|Wx\|_{2}.
\end{align}
The substitution $x = \sqrt{d_{\mathrm{in}}}\,\tilde{x}$ rescales the constraint to $\|\tilde{x}\|_{2} = 1$,
\begin{align}
\|W\|_{\mathrm{RMS} \to \mathrm{RMS}}
=
\frac{\sqrt{d_{\mathrm{in}}}}{\sqrt{d_{\mathrm{out}}}} \max_{\|\tilde{x}\|_{2} = 1} \|W\tilde{x}\|_{2}
=
\sqrt{\tfrac{d_{\mathrm{in}}}{d_{\mathrm{out}}}}\, \|W\|_{2},
\end{align}
where the last equality uses the standard characterization of the spectral norm. \qedhere
\end{proof}

A useful consequence: a matrix has unit RMS-to-RMS norm if and only if its largest singular value is exactly $\sqrt{d_{\mathrm{out}}/d_{\mathrm{in}}}$. We use this repeatedly below.

\subsection{The dual norm and duality map of the RMS-to-RMS norm}
\label{app:rms_dual}

The dual quantities of the RMS-to-RMS norm have a clean SVD-based description: the dual norm is a rescaled trace of singular values, and the duality map is the SVD-aligned direction $UV^{\top}$ (also rescaled).

\begin{lemma}
\label{lem:rms_dual}
Let $G \in \mathbb{R}^{d_{\mathrm{out}} \times d_{\mathrm{in}}}$ have reduced SVD $G = U \Sigma V^{\top}$, where $\Sigma = \operatorname{diag}(\sigma_{1}, \dots, \sigma_{r})$ with $r = \min(d_{\mathrm{out}}, d_{\mathrm{in}})$. Then
\begin{align}
\|G\|_{\mathrm{RMS} \to \mathrm{RMS}, *}
&=
\sqrt{\tfrac{d_{\mathrm{out}}}{d_{\mathrm{in}}}}\, \operatorname{tr}\Sigma,
\label{eq:app_rms_dual_norm}
\\
\operatorname{dualize}_{\|\cdot\|_{\mathrm{RMS} \to \mathrm{RMS}}}(G)
&=
\sqrt{\tfrac{d_{\mathrm{out}}}{d_{\mathrm{in}}}}\, U V^{\top}.
\label{eq:app_rms_dual_map}
\end{align}
The duality map is unique if and only if $G$ has full rank.
\end{lemma}

\begin{proof}
By Lemma~\ref{lem:rms_equals_spec}, the constraint $\|T\|_{\mathrm{RMS} \to \mathrm{RMS}} = 1$ is equivalent to $\|T\|_{2} = \sqrt{d_{\mathrm{out}}/d_{\mathrm{in}}}$.

\emph{Upper bound.} Expanding the Frobenius inner product using the SVD of $G$,
\begin{align}
\langle G, T \rangle
=
\operatorname{tr}(V \Sigma U^{\top} T)
=
\sum_{i=1}^{r} \sigma_{i} \, u_{i}^{\top} T v_{i}.
\label{eq:app_inner_expansion}
\end{align}
Since $u_{i}$ and $v_{i}$ are unit vectors and $\|T\|_{2} = \sqrt{d_{\mathrm{out}}/d_{\mathrm{in}}}$, each summand is bounded by
\begin{align}
|u_{i}^{\top} T v_{i}|
\leq
\|u_{i}\|_{2} \cdot \|T v_{i}\|_{2}
\leq
\|T\|_{2}
=
\sqrt{\tfrac{d_{\mathrm{out}}}{d_{\mathrm{in}}}}.
\label{eq:app_term_bound}
\end{align}
Since $\sigma_{i} \geq 0$, summing gives
\begin{align}
\langle G, T \rangle
\leq
\sqrt{\tfrac{d_{\mathrm{out}}}{d_{\mathrm{in}}}}\, \sum_{i=1}^{r} \sigma_{i}
=
\sqrt{\tfrac{d_{\mathrm{out}}}{d_{\mathrm{in}}}}\, \operatorname{tr}\Sigma.
\label{eq:app_dual_upper_bound}
\end{align}

Take
\begin{align}
T^{\star} := \sqrt{\tfrac{d_{\mathrm{out}}}{d_{\mathrm{in}}}}\, U V^{\top}.
\end{align}
The matrix $UV^{\top}$ has spectral norm $1$ (its nonzero singular values are all equal to $1$), so $\|T^{\star}\|_{2} = \sqrt{d_{\mathrm{out}}/d_{\mathrm{in}}}$ as required. Furthermore, $V^{\top} v_{i} = e_{i}$ and $U e_{i} = u_{i}$, so
\begin{align}
u_{i}^{\top} T^{\star} v_{i}
=
\sqrt{\tfrac{d_{\mathrm{out}}}{d_{\mathrm{in}}}}\, u_{i}^{\top} u_{i}
=
\sqrt{\tfrac{d_{\mathrm{out}}}{d_{\mathrm{in}}}}
\end{align}
for each $i$ with $\sigma_{i} > 0$. Substituting into \eqref{eq:app_inner_expansion},
\begin{align}
\langle G, T^{\star} \rangle
=
\sqrt{\tfrac{d_{\mathrm{out}}}{d_{\mathrm{in}}}}\, \operatorname{tr}\Sigma,
\end{align}
which attains the upper bound of \eqref{eq:app_dual_upper_bound}. This proves both \eqref{eq:app_rms_dual_norm} and that $T^{\star}$ is a valid duality map.

\emph{Uniqueness.} When $G$ has full rank, every $\sigma_{i} > 0$, so equality must hold in every term of \eqref{eq:app_term_bound}. By the equality conditions for Cauchy--Schwarz, this forces $T v_{i} = \sqrt{d_{\mathrm{out}}/d_{\mathrm{in}}}\, u_{i}$ for every $i$. The vectors $\{v_{i}\}$ form a basis of $\mathbb{R}^{d_{\mathrm{in}}}$, so $T$ is determined uniquely on its full domain, giving $T = T^{\star}$. When $G$ is rank-deficient, the basis $\{v_{i}\}$ spans only the row space of $G$, leaving $T$ free on the null space, so the duality map is not unique. \qedhere
\end{proof}

The intuition behind \eqref{eq:app_rms_dual_map} is that, among all matrices of bounded RMS-to-RMS norm, the one that aligns most with $G$ is the SVD-aligned direction $UV^{\top}$ (rescaled to satisfy the constraint). The SVD-aligned direction preserves the singular directions of $G$ but flattens its spectrum to all-ones, putting the entire ``budget'' uniformly along the directions $G$ already uses.

\subsection{Multi-layer steepest descent}
\label{app:multilayer}

We now combine Lemmas~\ref{lem:steepest_closed_form} and~\ref{lem:rms_dual} with the layerwise-max network norm to prove the main theorem. We adopt the main-text notation $d_{\mathrm{in},l}, d_{\mathrm{out},l}$ for layer-$l$ input/output dimensions.

\begin{proof}[Proof of Theorem~\ref{thm:spectral_matches}]
The steepest-descent problem under the layerwise-max RMS-to-RMS network norm reads
\begin{align}
\arg\min_{\Delta W_{1}, \dots, \Delta W_{L}}
\Biggl[
\sum_{l=1}^{L}
\langle G_{l}, \Delta W_{l} \rangle
+
\frac{\lambda}{2}
\max_{l}
\bigl\|\Delta W_{l}\bigr\|_{\mathrm{RMS} \to \mathrm{RMS}}^{2}
\Biggr].
\label{eq:app_multilayer_problem}
\end{align}
For each layer, decompose $\Delta W_{l} = c_{l} \cdot T_{l}$ into magnitude $c_{l} \geq 0$ and direction $T_{l}$ with $\|T_{l}\|_{\mathrm{RMS} \to \mathrm{RMS}} = 1$. The objective becomes
\begin{align}
\sum_{l=1}^{L} c_{l}\, \langle G_{l}, T_{l} \rangle
+
\frac{\lambda}{2}\, \max_{l} c_{l}^{2}.
\label{eq:app_decomposed_objective}
\end{align}
We minimize in three steps: directions $T_{l}$ first (each independent of the others), then the layerwise magnitudes $c_{l}$.

\emph{Step 1: optimal direction at each layer.}
Each $T_{l}$ appears in only one term of \eqref{eq:app_decomposed_objective}, namely $c_{l}\, \langle G_{l}, T_{l} \rangle$, and its constraint is independent of the other layers. By Lemma~\ref{lem:rms_dual}, the unit-norm direction minimizing $c_{l}\, \langle G_{l}, T_{l} \rangle$ is the negative of the duality map,
\begin{align}
T_{l}^{\star}
=
-\operatorname{dualize}_{\|\cdot\|_{\mathrm{RMS} \to \mathrm{RMS}}}(G_{l})
=
-\sqrt{\tfrac{d_{\mathrm{out},l}}{d_{\mathrm{in},l}}}\, U_{l} V_{l}^{\top},
\label{eq:app_optimal_direction}
\end{align}
and the resulting linear contribution is
\begin{align}
c_{l}\, \langle G_{l}, T_{l}^{\star} \rangle
=
-c_{l}\, \|G_{l}\|_{\mathrm{RMS} \to \mathrm{RMS}, *}
=
-c_{l}\, s_{l},
\quad\text{where}\quad
s_{l} := \sqrt{\tfrac{d_{\mathrm{out},l}}{d_{\mathrm{in},l}}}\, \operatorname{tr}\Sigma_{l}.
\label{eq:app_optimal_linear}
\end{align}
The scalar $s_{l} \geq 0$ is the per-layer ``reward rate'' for spending magnitude.

\emph{Step 2: all magnitudes meet at a common value.}
Substituting \eqref{eq:app_optimal_linear}, the problem reduces to a scalar optimization,
\begin{align}
\min_{c_{1}, \dots, c_{L} \geq 0}
\Biggl[
-\sum_{l=1}^{L} c_{l}\, s_{l}
+
\frac{\lambda}{2}\, \max_{l} c_{l}^{2}
\Biggr].
\label{eq:app_scalar_problem}
\end{align}
The structure of \eqref{eq:app_scalar_problem} is informative: the penalty depends only on the \emph{largest} $c_{l}$, while every $c_{l}$ contributes a linear reward $s_{l}$. Increasing any $c_{l}$ that is below $\max_{l} c_{l}$ therefore costs nothing (the max is unchanged) but yields a positive linear gain $s_{l} > 0$. Consequently, all $c_{l}$ with $s_{l} > 0$ must coincide at the optimum; layers with $s_{l} = 0$ have no incentive to use any magnitude and can be set to zero. Setting $c_{1} = \cdots = c_{L} = c^{\star}$, the problem collapses to
\begin{align}
\min_{c^{\star} \geq 0}
\Biggl[
-c^{\star} \sum_{l=1}^{L} s_{l}
+
\frac{\lambda}{2}\, (c^{\star})^{2}
\Biggr],
\end{align}
which is a convex quadratic in $c^{\star}$, minimized at
\begin{align}
c^{\star}
=
\frac{1}{\lambda} \sum_{l=1}^{L} s_{l}
=
\frac{1}{\lambda} \sum_{l=1}^{L} \sqrt{\tfrac{d_{\mathrm{out},l}}{d_{\mathrm{in},l}}}\, \operatorname{tr}\Sigma_{l}.
\label{eq:app_cstar}
\end{align}

\emph{Step 3: assemble the update.}
Combining $\Delta W_{l}^{\star} = c^{\star} \cdot T_{l}^{\star}$ with \eqref{eq:app_optimal_direction} and \eqref{eq:app_cstar},
\begin{align}
\Delta W_{l}^{\star}
=
-c^{\star}\, \sqrt{\tfrac{d_{\mathrm{out},l}}{d_{\mathrm{in},l}}}\, U_{l} V_{l}^{\top}
=
-\eta_{l}\, U_{l} V_{l}^{\top},
\end{align}
where the layer-$l$ learning rate is the global scalar $c^{\star}$ multiplied by the layer-specific dimensional factor:
\begin{align}
\eta_{l}
=
c^{\star}\, \sqrt{\tfrac{d_{\mathrm{out},l}}{d_{\mathrm{in},l}}}
=
\frac{1}{\lambda}
\Biggl(\sum_{k=1}^{L} \sqrt{\tfrac{d_{\mathrm{out},k}}{d_{\mathrm{in},k}}}\, \operatorname{tr}\Sigma_{k}\Biggr)
\sqrt{\tfrac{d_{\mathrm{out},l}}{d_{\mathrm{in},l}}}.
\end{align}
This matches the statement of Theorem~\ref{thm:spectral_matches}.

\emph{Uniqueness.} The direction $T_{l}^{\star}$ in Step 1 is unique exactly when $G_{l}$ has full rank (Lemma~\ref{lem:rms_dual}). The magnitude in Step 2 is uniquely determined whenever $\sum_{l} s_{l} > 0$, which holds whenever any $G_{l}$ has nonzero singular values. The overall minimizer is therefore unique iff every $G_{l}$ has full rank. \qedhere
\end{proof}

The update rule $\Delta W_{l}^{\star} = -\eta_{l}\, U_{l} V_{l}^{\top}$ has a clean interpretation. The direction $-U_{l} V_{l}^{\top}$ is the SVD-aligned direction of $-G_{l}$: it preserves the singular subspaces of the gradient but flattens the singular spectrum. The magnitude $\eta_{l}$ couples a global term $c^{\star}$ with a per-layer dimensional factor. The global term aggregates contributions from every layer through the dual-norm sum, so the step taken at any one layer reflects the gradient signal accumulated across the network. The per-layer factor $\sqrt{d_{\mathrm{out},l}/d_{\mathrm{in},l}}$ is the same dimensional rescaling that appears in Lemma~\ref{lem:rms_equals_spec}; in practice we absorb it into the per-layer learning rate, giving the unrescaled update $\Delta W_{l} = -\eta\, U_{l} V_{l}^{\top}$ used in our experiments.

\paragraph{Relation to the prior work.}
The view that the spectral norm provides a natural geometry for transformer weight matrices was developed on theoretical grounds by~\cite{bernstein2024old}, who interpret several existing optimizers as steepest descent under different operator norms. Their analysis establishes the geometric foundation, and we build on it with a controlled empirical study at the LLM scale.
Our contribution is complementary and primarily empirical: 
(i)~we extend their layerwise argument to a layerwise-maximum RMS-to-RMS 
network norm and obtain a closed-form per-layer step size in 
Theorem~\ref{thm:spectral_matches}; 
(ii)~we measure, throughout training, the cosine similarity between three 
practical preconditioners and the optimal Kronecker direction 
(Fig.~\ref{fig:fig2}; Figs.~\ref{fig:gd_blocks7_15},~\ref{fig:npo_rmu_block1}), 
confirming that the spectral direction is realizable at LLM scale; 
and (iii)~we instantiate this geometry on a non-trivial downstream task---LLM 
unlearning---across three loss families, two model scales, and three forget 
splits, and demonstrate consistent gains over entry-wise updates 
(Tab.~\ref{tab:paper_main}, Fig.~\ref{fig:fig3}). 
To our knowledge, this is the first systematic empirical validation that the 
spectral-norm geometry, beyond being theoretically natural, operationally improves LLM training in a post-training regime such as unlearning.

\section{Practical Realization on the LLM scale}
\label{app:practical_realization}

This appendix analyzes how the spectral update of Theorem~\ref{thm:spectral_matches} is instantiated on Llama-3.2-1B and 3B. We address (i) which dimension of each weight matrix lives in the RMS-normalized residual stream, (ii) why our asymmetric preconditioning aligns with this structure, and (iii) why the resulting one-sided update is exact rather than approximate.

\subsection{Architectural placement of RMSNorm}
\label{app:rmsnorm_placement}

In a pre-norm transformer, RMSNorm is applied only to vectors in the residual-stream dimension $d$. The MLP intermediate dimension $d_{\mathrm{ff}}$ is transient: $W_{1}$ expands $d \to d_{\mathrm{ff}}$, a pointwise activation is applied, and $W_{2}$ projects back to $d$. No normalization layer touches the $d_{\mathrm{ff}}$-space.

Consequently, the RMS-to-RMS geometry of \S\ref{sec:rms_norm} is well-defined on the $d$-side of every weight matrix but not on the $d_{\mathrm{ff}}$-side. This distinction motivates the preconditioning design that follows.

\subsection{Per-matrix dimension regime}
\label{app:per_matrix_table}

We set $\mathtt{max\_precond\_dim} = d$ ($d=2048$ for 1B; $d=3072$ for 3B) with $\mathtt{merge\_dims} = \text{False}$. Under this setting, the SOAP accumulators $L = \sum_{t} G_{t} G_{t}^{\top}$ and $R = \sum_{t} G_{t}^{\top} G_{t}$ are maintained per side only when that side does not exceed $d$. Tab.~\ref{tab:matrix_regime} summarizes the resulting regime.

\begin{table}[htbp]
\centering
\footnotesize
\setlength{\tabcolsep}{4pt}
\renewcommand{\arraystretch}{1.15}
\caption{Per-matrix preconditioning regime under $\mathtt{max\_precond\_dim} = d$. The un-preconditioned side coincides with the un-normalized $d_{\mathrm{ff}}$-dimension.}
\label{tab:matrix_regime}
\resizebox{\textwidth}{!}{%
\begin{tabular}{@{}lccccccc@{}}
\toprule
& \multicolumn{2}{c}{\textbf{Shape} ($d_{\mathrm{out}} \times d_{\mathrm{in}}$)} & \multirow{2}{*}{\textbf{Input}} & \multirow{2}{*}{\textbf{RMS side}} & \multicolumn{2}{c}{\textbf{Precond.}} & \multirow{2}{*}{\textbf{Regime}} \\
\cmidrule(lr){2-3} \cmidrule(lr){6-7}
\textbf{Matrix} & 1B & 3B & & & $L$ & $R$ & \\
\midrule
$W_{Q}$ & $2048 \times 2048$ & $3072 \times 3072$ & $\mathrm{RMSNorm}(X_{l})$ & both & \checkmark & \checkmark & two-sided \\
$W_{K}$ & $2048 \times 2048$ & $3072 \times 3072$ & $\mathrm{RMSNorm}(X_{l})$ & both & \checkmark & \checkmark & two-sided \\
$W_{V}$ & $2048 \times 2048$ & $3072 \times 3072$ & $\mathrm{RMSNorm}(X_{l})$ & both & \checkmark & \checkmark & two-sided \\
$W_{O}$ & $2048 \times 2048$ & $3072 \times 3072$ & attention out & both$^{\dagger}$ & \checkmark & \checkmark & two-sided \\
\midrule
$W_{1}$ (gate, up\_proj) & $8192 \times 2048$ & $8192 \times 3072$ & $\mathrm{RMSNorm}(H_{l})$ & input ($d$) & --- & \checkmark & right-only \\
$W_{2}$ (down\_proj) & $2048 \times 8192$ & $3072 \times 8192$ & $\mathrm{activation}(\cdot)$ & output ($d$) & \checkmark & --- & left-only \\
\midrule
$W_{\mathrm{emb}}$ & $128256 \times d$ & $128256 \times d$ & token id & --- & \multicolumn{2}{c}{AdamW} & entry-wise \\
\bottomrule
\end{tabular}%
}
\vspace{0.3em}
\\
{\footnotesize $^{\dagger}$ Bounded by the value projection's spectral norm; the corresponding scale factor is absorbed into the per-layer learning rate (see App.~\ref{app:rms_scope}).}
\end{table}

For attention projections, both sides equal $d$ and full two-sided preconditioning applies. For MLP matrices, the side that is skipped is in each case the un-normalized $d_{\mathrm{ff}}$-dimension.

\subsection{Scope of the RMS-to-RMS assumption}
\label{app:rms_scope}

Theorem~\ref{thm:spectral_matches} idealizes every weight matrix as acting on unit-RMS inputs. This holds exactly for $W_{Q}, W_{K}, W_{V}, W_{1}$, which receive RMSNorm output directly. Inputs to $W_{O}$ and $W_{2}$ traverse attention and activation respectively, and have RMS norms that are bounded but not strictly unit. The deviation is benign for the steepest-descent direction.

\paragraph{Scale-invariance of the spectral direction.}
For inputs with $\|x\|_{\mathrm{RMS}} = c > 0$, the induced operator norm becomes
\begin{align}
\|W\|_{c\text{-}\mathrm{RMS}}
:= \max_{\|x\|_{\mathrm{RMS}} = c} \|Wx\|_{\mathrm{RMS}}
= c\,\sqrt{\tfrac{d_{\mathrm{in}}}{d_{\mathrm{out}}}}\,\|W\|_{2},
\end{align}
which is a direct rescaling of Lemma~\ref{lem:rms_equals_spec}. Applying Lemmas~\ref{lem:rms_dual} and~\ref{lem:steepest_closed_form} in turn yields the steepest-descent direction
\begin{align}
T^{\star}(c) = -\sqrt{\tfrac{d_{\mathrm{out}}}{d_{\mathrm{in}}}}\, U V^{\top},
\end{align}
which is \emph{independent of $c$}; only the implied step size scales as $1/c$. The non-unit RMS at the inputs to $W_{O}$ and $W_{2}$ therefore affects only the per-layer step size, which is absorbed into the learning rate as in Theorem~\ref{thm:spectral_matches}.

\subsection{Asymmetric preconditioning aligns with the RMS-controlled dimension}
\label{app:asymmetric_preconditioning}

The asymmetric preconditioning in Tab.~\ref{tab:matrix_regime} is not a memory-driven approximation but a direct reflection of the architectural distinction in App.~\ref{app:rmsnorm_placement}. For $W_{1} \in \mathbb{R}^{d_{\mathrm{ff}} \times d}$, the input side ($d$) lives in the RMS-normalized residual stream while the output side ($d_{\mathrm{ff}}$) does not; we accumulate $R \in \mathbb{R}^{d \times d}$ and skip $L$. For $W_{2} \in \mathbb{R}^{d \times d_{\mathrm{ff}}}$, the situation is mirrored: we accumulate $L \in \mathbb{R}^{d \times d}$ and skip $R$. In both cases, the un-preconditioned side is precisely the side where RMS-to-RMS geometry does not apply. The setting $\mathtt{max\_precond\_dim} = d$ enforces this alignment automatically.

\subsection{One-sided preconditioning recovers the exact SVD-aligned direction}
\label{app:one_sided_exact}

The asymmetric design above might appear to approximate Theorem~\ref{thm:spectral_matches}, since the SVD-aligned direction $UV^{\top}$ is conventionally defined via the full SVD. The following result shows that no approximation is incurred.

\begin{proposition}[One-sided preconditioning recovers the SVD-aligned direction]
\label{prop:one_sided_polar}
Let $G \in \mathbb{R}^{m \times n}$ have reduced SVD $G = U\Sigma V^{\top}$ with $r = \min(m,n)$, and assume $\Sigma$ is invertible.
\begin{enumerate}[leftmargin=2em, itemsep=1pt]
\item[\textit{(i)}] If $m \ge n$ (tall):
\begin{align}
\Delta W^{\mathrm{right}} &:= -\eta\, G\,(G^{\top} G)^{-1/2} = -\eta\, U V^{\top}.
\end{align}
\item[\textit{(ii)}] If $m \le n$ (wide):
\begin{align}
\Delta W^{\mathrm{left}} &:= -\eta\, (G G^{\top})^{-1/2}\, G = -\eta\, U V^{\top}.
\end{align}
\end{enumerate}
\end{proposition}

\begin{proof}
\textit{(i)} $G^{\top} G = V \Sigma^{2} V^{\top}$ gives $(G^{\top} G)^{-1/2} = V \Sigma^{-1} V^{\top}$, so
\begin{align}
G(G^{\top} G)^{-1/2}
&= U \Sigma V^{\top} \cdot V \Sigma^{-1} V^{\top}
= U V^{\top}.
\end{align}
\textit{(ii)} Symmetric: $(G G^{\top})^{-1/2} = U \Sigma^{-1} U^{\top}$, so
\begin{align}
(G G^{\top})^{-1/2} G
&= U \Sigma^{-1} U^{\top} \cdot U \Sigma V^{\top}
= U V^{\top}. \qedhere
\end{align}
\end{proof}

The two cases match the two MLP regimes of Tab.~\ref{tab:matrix_regime} exactly: $W_{1}$ is tall and uses right-only preconditioning, $W_{2}$ is wide and uses left-only. The architectural alignment of App.~\ref{app:asymmetric_preconditioning} is therefore mathematically exact.

\paragraph{Accumulated preconditioners.}
Prop.~\ref{prop:one_sided_polar} treats the instantaneous gradient. SOAP and KLShampoo accumulate $G_{t} G_{t}^{\top}$ and $G_{t}^{\top} G_{t}$ via EMA, yielding covariance-whitened approximations of the SVD-aligned direction. The same one-sided structure carries over: the side with accumulated statistics matches the polar-factor direction in expectation, while the un-accumulated side is correctly left identity by the architectural argument.

\subsection{Architectural Alignment of \texttt{max\_precond\_dim}}
\label{app:hyper-precond-summary}

Combining the matrix-shape inventory of attention and MLP projections (Tab.~\ref{tab:matrix_regime}) with the scale invariance of the SVD-aligned direction and the one-sided polar reduction (Prop.~\ref{prop:one_sided_polar}), one obtains a coherent picture across both blocks: the $d \times d$ attention projections $W_Q, W_K, W_V$ receive exact two-sided preconditioning, the output projection $W_O$ is preconditioned up to a scale absorbed by RMS normalization, and the rectangular MLP projections receive one-sided preconditioning that nonetheless recovers the polar-factor direction exactly. Under this lens, $\mathtt{max\_precond\_dim} = d$ is an architectural alignment with the transformer's matrix structure, not a memory-driven compromise.

\section{Experimental Setup and Hyperparameter}
\label{app:hyper}

\subsection{Hardware and Software Stack}
\label{app:hyper-stack}

All experiments are conducted on a server equipped with 4 NVIDIA RTX A6000 GPUs (48~GB each), running CUDA~12.x. For Llama-3.2-1B and 3B-Instruct, each training run fits on a single GPU; multi-GPU parallelism is exploited only at the job level, with one training process per GPU within our learning-rate sweep. For the 7B-scale Zephyr-7B-$\beta$ used in WMDP experiments, we employ DeepSpeed ZeRO Stage~3~\cite{rajbhandari2020zero} to shard optimizer states, gradients, and parameters across all four GPUs. The deep-learning stack is built on PyTorch with FlashAttention-2, the HuggingFace Transformers (v4.51.3) and Accelerate libraries, and TensorBoard logging. The base models for TOFU experiments are the OpenUnlearning fine-tuned checkpoints released for Llama-3.2-1B-Instruct and 3B-Instruct.

\subsection{Common Training Configuration}
\label{app:hyper-train}

Unless stated otherwise, every unlearning run shares the following training configuration. We train for $T = 10$ epochs with the first epoch reserved for linear warmup, followed by a linear decay of the learning rate to zero over the remaining nine epochs. Optimization runs in bfloat16 mixed precision; for spectral updates (Muon, KLShampoo, SOAP), the optimizer state itself is also stored in bfloat16, while the entry-wise baselines (AdamW, Sophia) follow their upstream default dtypes. Weight decay is fixed at $0.01$, and gradient checkpointing is disabled because both 1B and 3B models comfortably fit in memory at our chosen batch sizes.

The micro-batch size is set per model so that the effective batch size matches the OpenUnlearning protocol within each scale: $8$ with gradient accumulation of $4$ (effective $32$) for the 1B model, and $4$ with the same accumulation factor (effective $16$) for the 3B model. The total update-step budget over $10$ epochs is approximately $13/63/125$ steps for the 1\%/5\%/10\% forget splits on the 1B model, and $25/125/250$ steps on the 3B model. Evaluation is performed once at the end of training so that the wall-clock numbers reported in the main paper exclude in-loop evaluation overhead.

\subsection{Unlearning Loss Hyperparameters}
\label{app:hyper-loss}

We use the canonical OpenUnlearning configurations for the three unlearning losses considered in the main paper.

\paragraph{GD.} Both forget and retain weights are set to $\gamma = \alpha = 1.0$, with the retain loss being the standard negative log-likelihood on full retain question--answer pairs.

\paragraph{NPO.} We use the inverse temperature $\beta = 0.1$ recommended by~\cite{zhang2024negative}, with $\gamma = \alpha = 1.0$. The reference model $\pi_{\mathrm{ref}}$ is loaded from the same fine-tuned checkpoint as $w_0$ and held frozen throughout training, contributing only forward-pass log-probabilities.

\paragraph{RMU.} The forget term steers hidden states at the 7th transformer block (a fixed mid-stack layer used in both 1B and 3B configurations) toward a random unit-norm direction with steering coefficient $c = 2$, and the retain regularizer is the embedding-distance loss against the reference model. All transformer parameters are trainable, mirroring the official RMU release rather than the layer-restricted variant used for WMDP.

\begin{table*}[t]
\centering
\footnotesize
\setlength{\tabcolsep}{4pt}
\renewcommand{\arraystretch}{1.12}
\setlength{\heavyrulewidth}{0.9pt}
\setlength{\lightrulewidth}{0.5pt}
\caption{Peak learning rate from the learning rate sweep.}
\label{tab:appendix-lr-per-cell}
\resizebox{\textwidth}{!}{
\begin{tabular}{l ccc ccc}
\toprule
 & \multicolumn{3}{c}{\textbf{Llama-3.2-1B-Instruct}} & \multicolumn{3}{c}{\textbf{Llama-3.2-3B-Instruct}} \\
\cmidrule(lr){2-4} \cmidrule(lr){5-7}
\textbf{Method} & Forget 10\% & Forget 5\% & Forget 1\% & Forget 10\% & Forget 5\% & Forget 1\% \\
\midrule
  GD-AdamW & $1\times 10^{-5}$ & $1\times 10^{-5}$ & $2\times 10^{-5}$ & $5\times 10^{-6}$ & $1\times 10^{-5}$ & $2\times 10^{-5}$ \\
  GD-Sophia & $5\times 10^{-6}$ & $1\times 10^{-5}$ & $2\times 10^{-5}$ & $5\times 10^{-6}$ & $7\times 10^{-6}$ & $2\times 10^{-5}$ \\
  SU-GD-Muon & $2\times 10^{-5}$ & $2\times 10^{-5}$ & $2\times 10^{-5}$ & $1\times 10^{-5}$ & $1.5\times 10^{-5}$ & $2\times 10^{-5}$ \\
  SU-GD-KLShampoo & $2\times 10^{-5}$ & $2\times 10^{-5}$ & $2\times 10^{-5}$ & $1\times 10^{-5}$ & $1.5\times 10^{-5}$ & $2\times 10^{-5}$ \\
  SU-GD-SOAP & $2\times 10^{-5}$ & $2\times 10^{-5}$ & $2\times 10^{-5}$ & $2\times 10^{-5}$ & $2\times 10^{-5}$ & $2\times 10^{-5}$ \\
  NPO-AdamW & $3\times 10^{-5}$ & $1\times 10^{-5}$ & $1\times 10^{-5}$ & $2\times 10^{-5}$ & $1\times 10^{-5}$ & $1\times 10^{-5}$ \\
  NPO-Sophia & $2\times 10^{-5}$ & $2\times 10^{-5}$ & $2\times 10^{-5}$ & $1\times 10^{-5}$ & $2\times 10^{-5}$ & $2\times 10^{-5}$ \\
  SU-NPO-Muon & $3\times 10^{-5}$ & $3\times 10^{-5}$ & $3\times 10^{-5}$ & $3\times 10^{-5}$ & $2\times 10^{-5}$ & $2\times 10^{-5}$ \\
  SU-NPO-KLShampoo & $3\times 10^{-5}$ & $3\times 10^{-5}$ & $3\times 10^{-5}$ & $3\times 10^{-5}$ & $2\times 10^{-5}$ & $2\times 10^{-5}$ \\
  SU-NPO-SOAP & $3\times 10^{-5}$ & $3\times 10^{-5}$ & $2\times 10^{-5}$ & $3\times 10^{-5}$ & $3\times 10^{-5}$ & $2\times 10^{-5}$ \\
  RMU-AdamW & $1\times 10^{-5}$ & $2\times 10^{-5}$ & $3\times 10^{-5}$ & $1\times 10^{-5}$ & $2\times 10^{-5}$ & $2\times 10^{-5}$ \\
  RMU-Sophia & $2\times 10^{-5}$ & $3\times 10^{-5}$ & $1\times 10^{-4}$ & $2\times 10^{-5}$ & $2\times 10^{-5}$ & $5\times 10^{-5}$ \\
  SU-RMU-Muon & $2\times 10^{-5}$ & $2\times 10^{-5}$ & $3\times 10^{-5}$ & $2\times 10^{-5}$ & $2\times 10^{-5}$ & $3\times 10^{-5}$ \\
  SU-RMU-KLShampoo & $2\times 10^{-5}$ & $2\times 10^{-5}$ & $3\times 10^{-5}$ & $2\times 10^{-5}$ & $3\times 10^{-5}$ & $3\times 10^{-5}$ \\
  SU-RMU-SOAP & $2\times 10^{-5}$ & $3\times 10^{-5}$ & $3\times 10^{-5}$ & $3\times 10^{-5}$ & $3\times 10^{-5}$ & $3\times 10^{-5}$ \\
\bottomrule
\end{tabular}
}
\end{table*}

\subsection{Optimizer Hyperparameters}
\label{app:hyper-opt}

Each loss is paired with five optimizers, yielding the $3 \times 5 = 15$ combinations reported in the main table. Optimizer-internal hyperparameters are fixed at the defaults from their respective original papers, and only the learning rate is tuned per $\langle\text{model}, \text{forget split}\rangle$ cell over the $9$-point log-spaced grid
\begin{align*}
\Lambda = \{1, 2, 3, 5\} \times \{10^{-6}, 10^{-5}\}
\;\cup\;
\{1\} \times 10^{-4}.
\end{align*}
This isolates the effect of the matrix-structured update direction from any tuning advantage the spectral updates might otherwise enjoy. Although the optimal learning rate varies across cells, the cell-wise winners cluster tightly around an optimizer-dependent value, which we report below as the empirical sweet-spot range. For GD on Llama-3.2-3B-Instruct with the 5\% forget split, we also tested two intermediate learning rates, $7\times10^{-6}$ and $1.5\times10^{-5}$, to identify the range in which retain-set performance remained stable. The fact that the sharper spectral updates prefer slightly larger learning rates is consistent with their unit-norm step direction, which has a smaller effective magnitude per step than the entry-wise sign update; this pattern is also reflected in the learning-rate sensitivity analysis of Fig.~\ref{fig:fig4}. The peak learning rate from the learning rate sweep is provided in Tab.~\ref{tab:appendix-lr-per-cell}.

\paragraph{AdamW.} Standard configuration with $\beta_1 = 0.9$, $\beta_2 = 0.999$, and $\varepsilon = 10^{-8}$.

\paragraph{Sophia.} Following SOUL~\cite{jia-etal-2024-soul}, we use the \texttt{SophiaG} variant with $\beta_1 = 0.9$, $\beta_2 = 0.95$, the per-coordinate clipping threshold $\rho = 0.04$, $\varepsilon = 10^{-5}$, and Hessian re-estimation every $10$ update steps.

\paragraph{Muon.} Momentum is set to $0.95$, and the SVD-aligned direction is approximated by $5$ Newton--Schulz iterations per step, matching the original release~\cite{jordan2024muon}. Bias correction is enabled.

\paragraph{KLShampoo.} The two preconditioner EMA coefficients are both set to $0.95$, the preconditioner is refreshed every $10$ steps, and we use a per-axis dimension cap of $3072$ with output clamping at $4000$ for numerical stability~\cite{lin2026understanding}.

\paragraph{SOAP.} The inner Adam coefficients are $(\beta_1, \beta_2) = (0.95, 0.95)$, with $\varepsilon = 10^{-8}$, the eigenbasis $(Q^L, Q^R)$ refreshed every $10$ steps via an \texttt{eigh} call in \texttt{float32}, and the rest of the hot path running in \texttt{bfloat16}~\cite{vyas2025soap}. The single-axis preconditioning cap is set to $\mathtt{max\_precond\_dim} = d$, where $d$ denotes the transformer hidden dimension. This choice is not driven by memory: rather, it aligns the preconditioner's coverage with the matrix structure of the model, so that the $d \times d$ attention projections receive full two-sided preconditioning, while MLP projections (with one side $4d$) are preconditioned only on the $d$-side---which, as discussed in Sec.~\ref{sec:kronecker_alignment}, recovers the polar-factor direction exactly under one-sided preconditioning. Sensitivity sweeps over $\beta_2$, $\beta_{\mathrm{shampoo}}$, $f_{\mathrm{prec}}$, and $\epsilon$ in the neighborhood of these defaults produced no statistically significant change in the aggregate score, consistent with the robustness reported for SOAP in the original paper~\cite{vyas2025soap}.

\subsection{Multi-Seed Protocol}
\label{app:hyper-seeds}

Every entry in the main-paper tables, except for the reference rows (Original, Retrain), is reported as mean${}\pm{}$standard deviation over three seeds $s \in \{0, 1, 2\}$. The seed controls the Trainer's data shuffler and any update-internal stochasticity through Python, NumPy, and PyTorch CPU/CUDA generators. We do not re-randomize the model initialization (which is fixed to the open-unlearning checkpoint) nor the data partition (which is fixed by the upstream split), so the seed governs only the training trajectory.

\subsection{Numerical Stability and Tokenization Details}
Token log-probabilities are obtained via \texttt{F.log\_softmax(logits, dim=-1)} rather than $\log(\mathrm{softmax}(\cdot))$ to avoid numerical underflow. Prompt tokens are masked with \texttt{ignore\_index} $= -100$ and excluded from all loss and likelihood computations. Logits and labels are shifted by one position, so that \texttt{logits[:-1]} predicts \texttt{labels[1:]}. Mixed-precision activations are cast to \texttt{float32} prior to metric aggregation.

\subsection{Comparison Algorithms}
\label{app:compare_algorithms}

This appendix details the unlearning baselines reported in Tab.~\ref{tab:comprehensive} and Tab.~\ref{tab:time_mem}. We follow the canonical formulations and configurations from the open-unlearning benchmark.

\paragraph{SimNPO~\cite{fan2024simplicity}.} A SimPO-style formulation with forget loss weight $\gamma = 0.125$, $\beta = 4.5$, and length penalty $\delta = 0$, combined with retain NLL.

\paragraph{IdkDPO~\cite{maini2024tofu}.} The same DPO formulation as NPO ($\beta = 0.1$), but additionally uses ``I don't know'' responses as the preferred completions on the forget set.

\paragraph{CEU~\cite{yang2025ceucrossentropyunlearning}.} A cross-entropy unlearning loss that skips the first token, with both forget and retain loss weights equal to $1$.

\paragraph{PDU~\cite{entesari2026constrained}.} A primal-dual formulation with a logit-gap-squared forget loss, an adaptive retain weight initialized at $1$, a constraint threshold $\epsilon = 0.3$, and $5$ epochs of dual-variable warmup.

\paragraph{UNDIAL~\cite{dong-etal-2025-undial}.} A self-distillation loss with temperature $\beta = 10$ and no retain regularizer. Uses a higher peak learning rate of $10^{-4}$ to compensate for the absence of explicit retention pressure.

\paragraph{WGA~\cite{wang2025rethinking}.} A weighted gradient attribution forget loss with $\beta = 1$, combined with retain NLL. Trained for $5$ epochs.

\paragraph{SatImp~\cite{yang2025exploring}.} A sharpness-aware forget loss with $\beta_1 = 5$, $\beta_2 = 1$, $\gamma = 0.1$, combined with retain NLL. Trained for $5$ epochs.

\paragraph{AltPO~\cite{mekala-etal-2025-alternate}.} An alternating preference optimization scheme that toggles between forget and retain phases under the canonical open-unlearning configuration.

\paragraph{GradAscent~\cite{yao-etal-2024-machine}.} The vanilla GA baseline described above, included as a lower-bound reference to highlight how much each refinement contributes.

\section{Evaluation Protocol Details}
\label{app:metric_details}

We adopt the OpenUnlearning evaluation framework~\cite{dorna2026openunlearning}, originally introduced for the TOFU benchmark~\cite{maini2024tofu}, and report results along three axes: memorization removal (Mem), privacy preservation (Priv), and utility retention (Util). All component scores are bounded in $[0, 1]$ and combined via the harmonic mean (HM), defined for non-negative inputs $x_1, \dots, x_n$ as
\begin{align}
\mathrm{HM}(x_1, \dots, x_n) = \frac{n}{\sum_{i=1}^{n} \max(x_i, \epsilon)^{-1}}, \qquad \epsilon = 10^{-12}.
\end{align}
We use the harmonic mean rather than the arithmetic mean throughout because it ensures that a method cannot mask a weak axis by excelling on another: a near-zero value in any single component drives the aggregate toward zero. This conservative aggregation is important for unlearning, where memorization removal, privacy, and utility are often in tension. The final \emph{aggregate score} is
\begin{align}
\mathrm{Agg} = \mathrm{HM}\bigl(\mathrm{Mem},\; \mathrm{Priv},\; \mathrm{Util}\bigr).
\end{align}

\subsection{Memorization Score}
\label{app:metric_mem}

The Mem score quantifies how thoroughly the target knowledge has been removed. We aggregate four sub-metrics computed on the forget set $\mathcal{D}_f$:
\begin{align}
\mathrm{Mem} = \mathrm{HM}\bigl(1 - \mathrm{ES},\; 1 - \mathrm{EM},\; 1 - \mathrm{F.Prob},\; \mathrm{F.TR}\bigr).
\end{align}
The first three sub-metrics measure residual knowledge directly (lower is better), so each is inverted as $1 - v$ before aggregation. The truth ratio (F.TR) is normalized so that higher already corresponds to better forgetting and is used as is.

\paragraph{Extraction Strength (ES).} For a target sequence of length $L$, let $\hat{t}_i = \arg\max_v \log p(v \mid t_{<i})$ be the greedy prediction at position $i$, and let $k$ be the smallest index for which the suffix matches, $\hat{t}_{k:L} = t_{k:L}$. ES is defined as $\mathrm{ES} = 1 - k/L$. Values near $1$ indicate that long contiguous suffixes can be extracted greedily.

\paragraph{Exact Memorization (EM).} EM measures token-level greedy agreement, $\mathrm{EM} = \frac{1}{L}\sum_{i=1}^{L} \mathbb{1}[\hat{t}_i = t_i]$. Unlike ES, EM credits non-contiguous matches and therefore captures partial memorization.

\paragraph{Forget Probability (F.Prob).} We compute the model's normalized likelihood on the original (non-paraphrased) forget prompts:
\begin{align}
\mathrm{F.Prob} = \frac{1}{|\mathcal{D}_f|} \sum_{z \in \mathcal{D}_f} \exp\!\left( -\frac{1}{L_z} \sum_{i=1}^{L_z} \mathrm{CE}(\mathrm{logits}_i, t_i) \right),
\end{align}
where prompt tokens are masked with \texttt{ignore\_index} $= -100$. We deliberately use F.Prob rather than the paraphrase-prompt variant adopted in some prior protocols, because the paraphrase signal is already captured by the truth ratio below; including both would double-count the same evidence. We nonetheless report the paraphrase probability (P.Prob) in our detailed sub-metric tables (Tabs.~\ref{tab:appendix_memgpriv_1b_forget10}--\ref{tab:appendix_memgpriv_3b_forget01}) for transparency and to facilitate comparison with prior work.

\paragraph{Forget Truth Ratio (F.TR).} For each forget example $z$, we compute the per-token likelihoods of the paraphrased correct answer ($p_z^{\mathrm{para}}$) and a perturbed incorrect answer ($p_z^{\mathrm{pert}}$), and form the raw ratio $R_z = p_z^{\mathrm{pert}} / (p_z^{\mathrm{para}} + 10^{-10})$. The forget-side truth ratio is then the symmetrized mean
\begin{align}
\mathrm{F.TR} = \frac{1}{|\mathcal{D}_f|} \sum_{z \in \mathcal{D}_f} \min\!\Bigl(R_z,\; (R_z + 10^{-10})^{-1}\Bigr).
\end{align}
F.TR is maximized when correct and incorrect answers receive comparable likelihood---the expected behavior of a model that has truly forgotten the target.

\subsection{Privacy Score}
\label{app:metric_priv}

Privacy is evaluated through resistance to membership inference attacks (MIA). We use four scoring functions---LOSS, ZLib, Min-$k\%$, and Min-$k\%$++~\cite{shi2024detecting}---and for each compute an attack AUC by treating forget examples as members and a held-out set as non-members. Let $\mathrm{AUC}_k$ denote the attack AUC of method $k \in \{\mathrm{LOSS}, \mathrm{ZLib}, \mathrm{MinK}, \mathrm{MinK}\!+\!+\}$ on the unlearned model, $\mathrm{AUC}_k^{\mathrm{retain}}$ the corresponding AUC on the retain-trained oracle, and $\mathrm{AUC}_k^{\mathrm{init}}$ the AUC on the original (pre-unlearning) model. Each attack contributes a similarity score
\begin{align}
s_k = 1 - |\mathrm{AUC}_k - \mathrm{AUC}_k^{\mathrm{retain}}|.
\end{align}
This score equals 1 when the two models have the same AUC under attack \(k\) and decreases with the absolute difference between their AUCs. The privacy score is
\begin{align}
\mathrm{Priv} = \mathrm{HM}\bigl(s_{\mathrm{LOSS}},\; s_{\mathrm{ZLib}},\; s_{\mathrm{MinK}},\; s_{\mathrm{MinK}\!+\!+}\bigr).
\end{align}
The four attack scoring functions follow the standard definitions: LOSS uses average cross-entropy, ZLib divides this loss by the zlib-compressed length of the input to control for intrinsic difficulty, Min-$k\%$ averages only the bottom $k = 40\%$ of token log-probabilities, and Min-$k\%$++ further standardizes each token's log-probability by the mean and variance of the per-position categorical distribution before applying the same bottom-$k$ rule.

\subsection{Utility Score}
\label{app:metric_util}

To preserve a uniform weighting across all retention sub-metrics, we compute Util as a single harmonic mean of ten components rather than nesting them through an intermediate ``model utility'' aggregate:
\begin{align}
\mathrm{Util} = \mathrm{HM}\bigl(
&\mathrm{R.Prob},\, \mathrm{R.ROUGE},\, \mathrm{R.TR}, \mathrm{RA.Prob},\, \mathrm{RA.ROUGE},\, \mathrm{RA.TR}, \nonumber \\
&\mathrm{WF.Prob},\, \mathrm{WF.ROUGE},\, \mathrm{WF.TR}, \mathrm{Fluency}\bigr).
\end{align}
The three TOFU evaluation splits play complementary roles: the \emph{retain} split (R.\textperiodcentered) checks that the model still answers questions about authors that should not be forgotten; the \emph{real-author} split (RA.\textperiodcentered) checks that out-of-domain factual knowledge about real authors is unaffected; and the \emph{world-facts} split (WF.\textperiodcentered) checks general world knowledge. Within each split, $\mathrm{Prob}$ is the average answer likelihood, $\mathrm{ROUGE}$ is the longest-common-subsequence recall (with stemming), and $\mathrm{TR}$ is the retain-side truth ratio defined as $\frac{1}{|\mathcal{D}|} \sum_z \max(0,\, 1 - R_z)$, which rewards the model for placing more mass on the correct answer than on the perturbed alternatives. For the RA and WF splits, $\mathrm{Prob}$ is normalized as $p_{\mathrm{correct}} / (p_{\mathrm{correct}} + \sum p_{\mathrm{wrong}} + 10^{-10})$ so that it directly measures discrimination between correct and incorrect choices. \emph{Fluency} is the average non-gibberish probability assigned by a pre-trained classifier (\texttt{madhurjindal/autonlp-Gibberish-Detector-492513457}) to the model's outputs on the forget set, capturing collapse modes where the model produces incoherent text rather than genuinely forgetting.

\section{Extended Empirical Analysis}
\label{app:extend_results}
This appendix collects supplementary figures that further support the results presented in the main paper.

\subsection{Gradient structure is consistent across losses and depths}
\label{app:gradient_structure}

Fig.~\ref{fig:fig2} in the main text reports the cosine similarity between each update rule and the true gradient covariance for a single block under GD unlearning. We extend this analysis along two axes—depth and loss function—to characterize the spectral structure of unlearning gradients more broadly.

\begin{figure}[htbp]
    \centering
    \includegraphics[width=\textwidth]{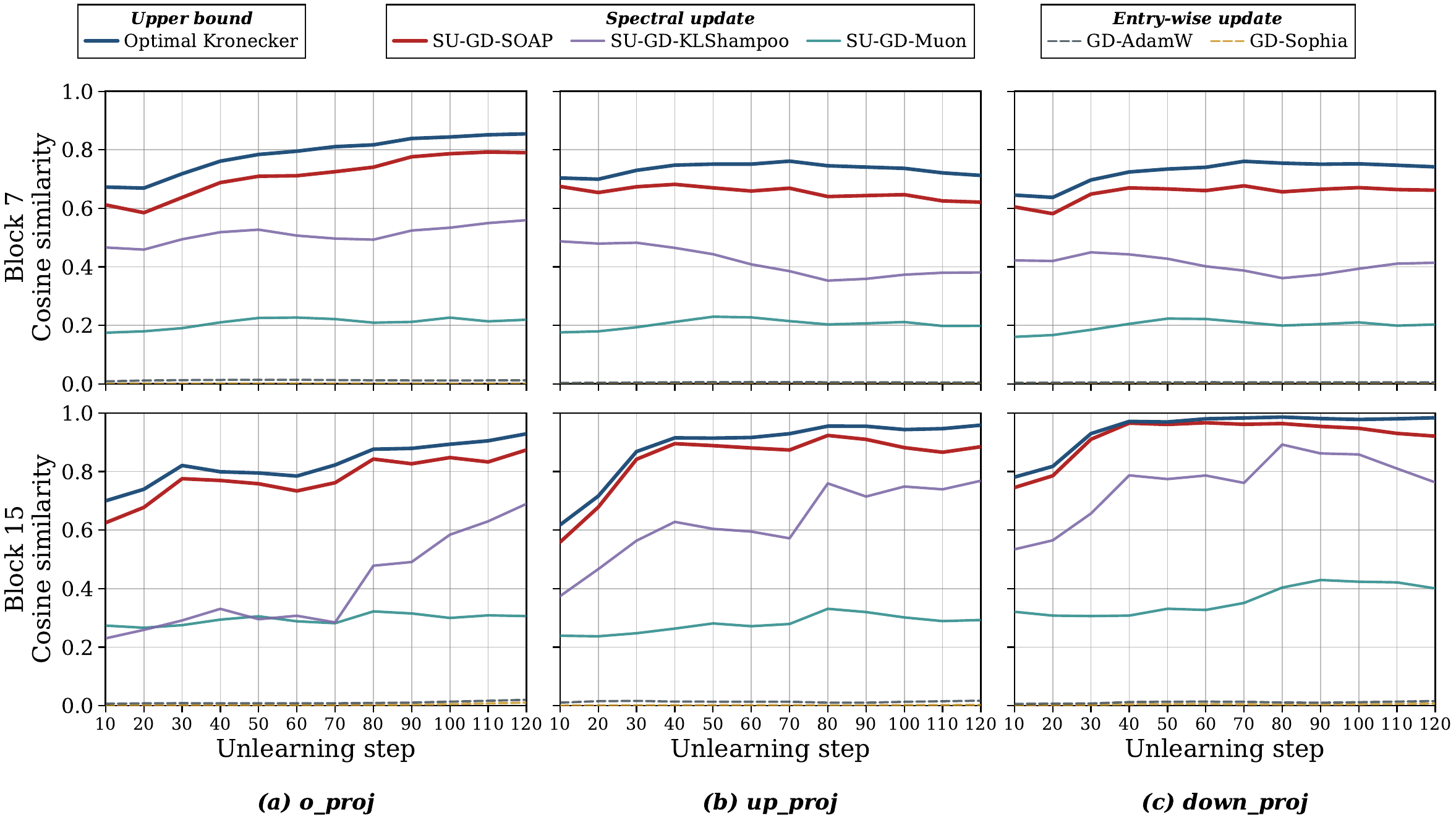}
    \caption{Cosine similarity between each update rule and the true gradient covariance, measured during GD unlearning on Llama-3.2-1B on 7 and 15 blocks.}
    \label{fig:gd_blocks7_15}
\end{figure}

\paragraph{Depth.}
Fig.~\ref{fig:gd_blocks7_15} reports the same measurement at block~7 (mid-stack) and block~15 (late-stack) under GD unlearning. The optimal Kronecker reference remains above $0.7$ at every block we measured, and the relative ordering of updates is preserved: SOAP and KLShampoo track the optimal Kronecker form most closely, Muon attains intermediate alignment, and entry-wise AdamW and Sophia remain near zero. Absolute alignment values are slightly higher at later blocks, consistent with deeper layers exhibiting more concentrated gradient covariance, but the structural pattern is qualitatively identical.

\begin{figure}[htbp]
    \centering
    \includegraphics[width=\textwidth]{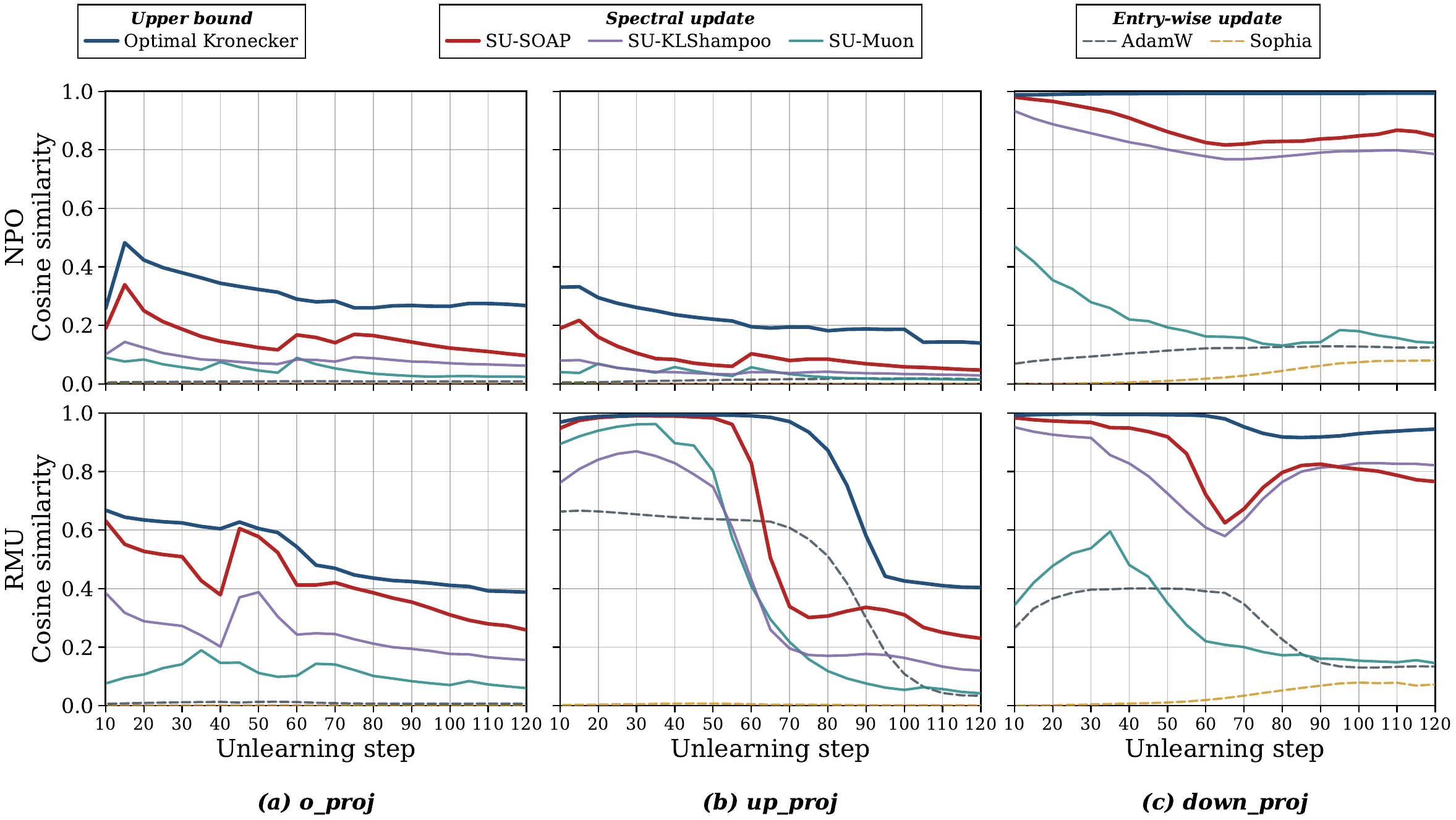}
    \caption{Cosine similarity between each update rule and the true gradient covariance, measured during NPO and RMU unlearning on Llama-3.2-1B.}
    \label{fig:npo_rmu_block1}
\end{figure}

\paragraph{Loss function.}
Fig.~\ref{fig:npo_rmu_block1} reports the same measurement under NPO and RMU unlearning at block~1. NPO produces somewhat lower absolute similarity values than GD, consistent with NPO's bounded forget term yielding smaller and noisier per-step gradients; nonetheless, the structural separation between spectral and entry-wise updates is preserved.

\paragraph{Implication.}
Taken together, Fig.~\ref{fig:fig2}, Fig.~\ref{fig:gd_blocks7_15}, and Fig.~\ref{fig:npo_rmu_block1} span three loss functions, three depths, and three matrix types within attention and MLP. Across all nine combinations, two observations are consistent: (i)~the optimal Kronecker preconditioner attains high alignment with the true gradient covariance, indicating that unlearning gradients reliably admit a strong matrix-structured (Kronecker) approximation; (ii)~entry-wise updates' diagonal preconditioners attain near-zero alignment, indicating that this structure is invisible to coordinate-wise updates. The combination explains why the gains from spectral unlearning persist across the loss functions, model scales, and forget splits reported in Sec.~\ref{sec:exp}: they originate from a structural property of unlearning gradients that is loss- and depth-invariant, rather than from a tuning advantage on any particular configuration.

\subsection{Sub-metric Comparison}
\label{app:Sub-metric_comparison}

\begin{table}[htbp]
\centering
\footnotesize
\setlength{\tabcolsep}{4pt}
\renewcommand{\arraystretch}{1.18}
\setlength{\heavyrulewidth}{0.9pt}
\setlength{\lightrulewidth}{0.5pt}
\caption{Spectral vs.\ entry-wise sub-metric comparison.}
\label{tab:family_winrate_raw}
\begin{tabular}{@{}l l c c c c c c c@{}}
\toprule
 & \textbf{Sub-metric} & \textbf{dir} & \textbf{Retain} & \textbf{Entry-wise} & \textbf{Spectral} & $\Delta$ & \textbf{Wins} & \\
\midrule
\rowcolor{blue!12}\multicolumn{2}{@{}l}{\textbf{Memorization} \textit{(family subtotal)}} & & & & & & \textbf{57/72} & \\
\rowcolor{blue!4} & \textbf{F.TR} & $\uparrow$ & $0.6472$ & $0.5756$ & $0.6059$ & \textcolor{green!55!black}{+0.0303} & \textbf{18/18} & \textcolor{green!55!black}{\rule[0.3ex]{2.60em}{0.7ex}}\textcolor{black!15}{\rule[0.3ex]{0.00em}{0.7ex}} \\
\rowcolor{blue!4} & F.Prob & $\downarrow$ & $0.1404$ & $0.1585$ & $0.0859$ & \textcolor{green!55!black}{-0.0727} & 14/18 & \textcolor{green!45!black}{\rule[0.3ex]{2.02em}{0.7ex}}\textcolor{black!15}{\rule[0.3ex]{0.58em}{0.7ex}} \\
\rowcolor{blue!4} & EM & $\downarrow$ & $0.6119$ & $0.6202$ & $0.5677$ & \textcolor{green!55!black}{-0.0525} & 13/18 & \textcolor{green!45!black}{\rule[0.3ex]{1.88em}{0.7ex}}\textcolor{black!15}{\rule[0.3ex]{0.72em}{0.7ex}} \\
\rowcolor{blue!4} & ES & $\downarrow$ & $0.0643$ & $0.0900$ & $0.0687$ & \textcolor{green!55!black}{-0.0213} & 12/18 & \textcolor{green!45!black}{\rule[0.3ex]{1.73em}{0.7ex}}\textcolor{black!15}{\rule[0.3ex]{0.87em}{0.7ex}} \\
\addlinespace[1pt]
\rowcolor{teal!15}\multicolumn{2}{@{}l}{\textbf{Privacy} \textit{(family subtotal)}} & & & & & & \textbf{55/72} & \\
\rowcolor{teal!5} & \textbf{MIA.ZLib} & $\!{\to}\!\mathrm{ret}$ & $0.3874$ & $0.4809$ & $0.3226$ & \textcolor{green!55!black}{-0.1583} & \textbf{16/18} & \textcolor{green!55!black}{\rule[0.3ex]{2.31em}{0.7ex}}\textcolor{black!15}{\rule[0.3ex]{0.29em}{0.7ex}} \\
\rowcolor{teal!5} & \textbf{MIA.MinK} & $\!{\to}\!\mathrm{ret}$ & $0.4206$ & $0.5133$ & $0.3373$ & \textcolor{green!55!black}{-0.1760} & \textbf{15/18} & \textcolor{green!55!black}{\rule[0.3ex]{2.17em}{0.7ex}}\textcolor{black!15}{\rule[0.3ex]{0.43em}{0.7ex}} \\
\rowcolor{teal!5} & MIA.Loss & $\!{\to}\!\mathrm{ret}$ & $0.4246$ & $0.5167$ & $0.3399$ & \textcolor{green!55!black}{-0.1768} & 14/18 & \textcolor{green!45!black}{\rule[0.3ex]{2.02em}{0.7ex}}\textcolor{black!15}{\rule[0.3ex]{0.58em}{0.7ex}} \\
\rowcolor{teal!5} & MIA.MinK\textsuperscript{++} & $\!{\to}\!\mathrm{ret}$ & $0.5094$ & $0.5676$ & $0.4710$ & \textcolor{green!55!black}{-0.0966} & 10/18 & \textcolor{orange!75!black}{\rule[0.3ex]{1.44em}{0.7ex}}\textcolor{black!15}{\rule[0.3ex]{1.16em}{0.7ex}} \\
\addlinespace[1pt]
\rowcolor{orange!12}\multicolumn{2}{@{}l}{\textbf{Utility} \textit{(family subtotal)}} & & & & & & \textbf{102/180} & \\
\rowcolor{orange!5} & R.ROUGE & $\uparrow$ & $0.8659$ & $0.5632$ & $0.5912$ & \textcolor{green!55!black}{+0.0280} & 12/18 & \textcolor{green!45!black}{\rule[0.3ex]{1.73em}{0.7ex}}\textcolor{black!15}{\rule[0.3ex]{0.87em}{0.7ex}} \\
\rowcolor{orange!5} & R.Prob & $\uparrow$ & $0.9105$ & $0.6214$ & $0.6598$ & \textcolor{green!55!black}{+0.0384} & 11/18 & \textcolor{orange!75!black}{\rule[0.3ex]{1.59em}{0.7ex}}\textcolor{black!15}{\rule[0.3ex]{1.01em}{0.7ex}} \\
\rowcolor{orange!5} & R.TR & $\uparrow$ & $0.5257$ & $0.5295$ & $0.5345$ & \textcolor{green!55!black}{+0.0050} & 11/18 & \textcolor{orange!75!black}{\rule[0.3ex]{1.59em}{0.7ex}}\textcolor{black!15}{\rule[0.3ex]{1.01em}{0.7ex}} \\
\rowcolor{orange!5} & RA.ROUGE & $\uparrow$ & $0.8295$ & $0.7013$ & $0.7264$ & \textcolor{green!55!black}{+0.0251} & 11/18 & \textcolor{orange!75!black}{\rule[0.3ex]{1.59em}{0.7ex}}\textcolor{black!15}{\rule[0.3ex]{1.01em}{0.7ex}} \\
\rowcolor{orange!5} & Fluency & $\uparrow$ & $0.8878$ & $0.7333$ & $0.7314$ & \textcolor{red!70!black}{-0.0019} & 10/18 & \textcolor{orange!75!black}{\rule[0.3ex]{1.44em}{0.7ex}}\textcolor{black!15}{\rule[0.3ex]{1.16em}{0.7ex}} \\
\rowcolor{orange!5} & RA.Prob & $\uparrow$ & $0.4362$ & $0.4326$ & $0.4358$ & \textcolor{green!55!black}{+0.0032} & 10/18 & \textcolor{orange!75!black}{\rule[0.3ex]{1.44em}{0.7ex}}\textcolor{black!15}{\rule[0.3ex]{1.16em}{0.7ex}} \\
\rowcolor{orange!5} & RA.TR & $\uparrow$ & $0.5638$ & $0.5667$ & $0.5701$ & \textcolor{green!55!black}{+0.0034} & 10/18 & \textcolor{orange!75!black}{\rule[0.3ex]{1.44em}{0.7ex}}\textcolor{black!15}{\rule[0.3ex]{1.16em}{0.7ex}} \\
\rowcolor{orange!5} & WF.Prob & $\uparrow$ & $0.4580$ & $0.4533$ & $0.4545$ & \textcolor{green!55!black}{+0.0012} & 9/18 & \textcolor{orange!75!black}{\rule[0.3ex]{1.30em}{0.7ex}}\textcolor{black!15}{\rule[0.3ex]{1.30em}{0.7ex}} \\
\rowcolor{orange!5} & WF.ROUGE & $\uparrow$ & $0.8581$ & $0.8242$ & $0.8287$ & \textcolor{green!55!black}{+0.0045} & 9/18 & \textcolor{orange!75!black}{\rule[0.3ex]{1.30em}{0.7ex}}\textcolor{black!15}{\rule[0.3ex]{1.30em}{0.7ex}} \\
\rowcolor{orange!5} & WF.TR & $\uparrow$ & $0.6338$ & $0.6279$ & $0.6325$ & \textcolor{green!55!black}{+0.0045} & 9/18 & \textcolor{orange!75!black}{\rule[0.3ex]{1.30em}{0.7ex}}\textcolor{black!15}{\rule[0.3ex]{1.30em}{0.7ex}} \\
\addlinespace[1pt]
\midrule
\rowcolor{gray!12}\multicolumn{7}{@{}l}{\textbf{Overall (spectral wins across all sub-metrics $\times$ cells)}} & \textbf{214/324} & \textcolor{orange!75!black}{\rule[0.3ex]{1.72em}{0.7ex}}\textcolor{black!15}{\rule[0.3ex]{0.88em}{0.7ex}} \\
\bottomrule
\end{tabular}
\end{table}

To better understand which axes drive the aggregate-score gains reported in the main paper, we conduct a head-to-head comparison between the spectral and entry-wise update families at the level of individual sub-metrics. For each cell defined by a $\langle$model, forget split, unlearning loss$\rangle$ tuple, we form two family-averaged sub-metric values: the \emph{Spectral} value averages over Muon, KLShampoo, and SOAP, while the \emph{Entry-wise} value averages over AdamW and Sophia. We then count, for each sub-metric, how often the spectral family beats the entry-wise family across the $18$ cells under the direction-aware comparison appropriate to that sub-metric (lower-is-better for residual memorization scores, higher-is-better for utility, closer-to-retain for MIA AUCs). The results are summarized in Tab.~\ref{tab:family_winrate_raw}.

Across all three evaluation axes, spectral updates win the majority of cells, and the gains are coherent rather than concentrated in a single sub-metric. On the memorization axis, the largest improvements appear on the forget truth ratio---where spectral updates win on every single cell---and on the forget-side likelihood, indicating that spectral updates push the model away from the forget distribution more thoroughly than entry-wise updates do. The exact-match and extraction-strength sub-metrics show smaller but still consistent gains, suggesting that the improvement is not driven by a single greedy-decoding artifact.

On the privacy axis, spectral updates move the unlearned model closer to the retain-trained reference under all four MIA scoring functions. The gap is most pronounced for the more stringent attacks (ZLib and Min-$k$), under which spectral updates often shift the model from clearly distinguishable toward nearly indistinguishable from the retain oracle, while Min-$k$++ shows a smaller margin consistent with its lower discriminative power even on the original model.

The utility axis exhibits a more nuanced pattern. On the in-domain retain split, spectral updates produce clear gains across answer probability, ROUGE, and truth ratio, indicating that retained-author knowledge is better preserved. On the out-of-distribution splits (real authors and world facts), however, the two families perform near-identically, with differences within seed noise; this is consistent with the observation that out-of-distribution knowledge is largely untouched by either update style and therefore offers little room for the update geometry to differentiate. Fluency similarly remains near the original-model level for both families, ruling out collapse modes such as repetitive or gibberish generation.

Taken together, the simultaneous improvement on the forget side (memorization, privacy) and on the in-domain retain side, with no regression on out-of-distribution utility, shows that the spectral gain is not the result of a memorization--utility trade-off. Rather, the matrix-structured update appears to allocate parameter changes more selectively, achieving stronger forgetting while better preserving the knowledge that should be kept. This sub-metric breakdown therefore confirms that the aggregate-score improvements reported in the main paper reflect coherent gains across multiple metric families, rather than a single dominant axis.

\subsection{Learning-rate robustness and distributional alignment}

\begin{figure}[htbp]
    \centering
    \includegraphics[width=\textwidth]{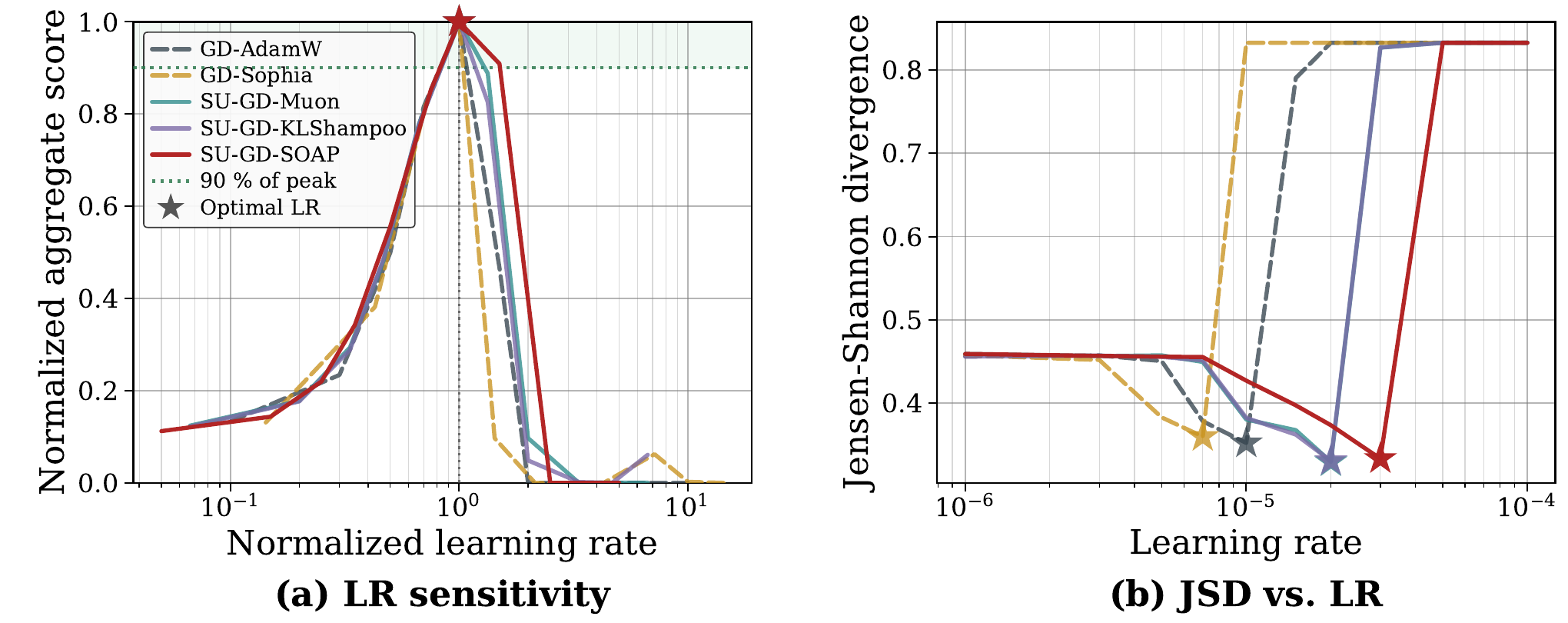}
    \caption{(a) Aggregate score and (b) Jensen--Shannon divergence (JSD) from the retain-trained reference, across learning rates. Both axes in (a) are normalized by each method's optimum.}
    \label{fig:paper_lrsens}
\end{figure}

Fig.~\ref{fig:paper_lrsens} complements the learning-rate sensitivity analysis of Fig.~\ref{fig:fig4} along two axes. Fig.~\ref{fig:paper_lrsens}(a) rescales both axes by each method's per-cell optimum, so that all methods peak at the same point. The shaded band marks the 90\%-of-peak region. Spectral updates retain at least 90\% of their peak aggregate score over a substantially wider band than entry-wise updates, with SU-GD-SOAP exhibiting the broadest plateau. This confirms that the robustness reported in Fig.~\ref{fig:fig4}(a) reflects a genuine difference in the curvature of the loss-vs-LR profile, not an artifact of absolute learning-rate scale.
Fig.~\ref{fig:paper_lrsens}(b) examines why spectral updates achieve higher aggregate scores. We measure the Jensen--Shannon divergence between the unlearned model's output distribution on the forget set and that of the retain-trained reference, which directly quantifies how closely the unlearning trajectory approaches the target. Across the entire learning-rate grid, SU-GD-SOAP attains lower JSD than the entry-wise baselines, and the gap widens at the optimal learning rate (marked by stars). The improved aggregate scores in Tab.~\ref{tab:paper_main} therefore reflect a tighter alignment with the retain-trained reference at the distribution level.

\subsection{Sub-metric sensitivity and GD instability}

\begin{figure}[htbp]
    \centering
    \includegraphics[width=\textwidth]{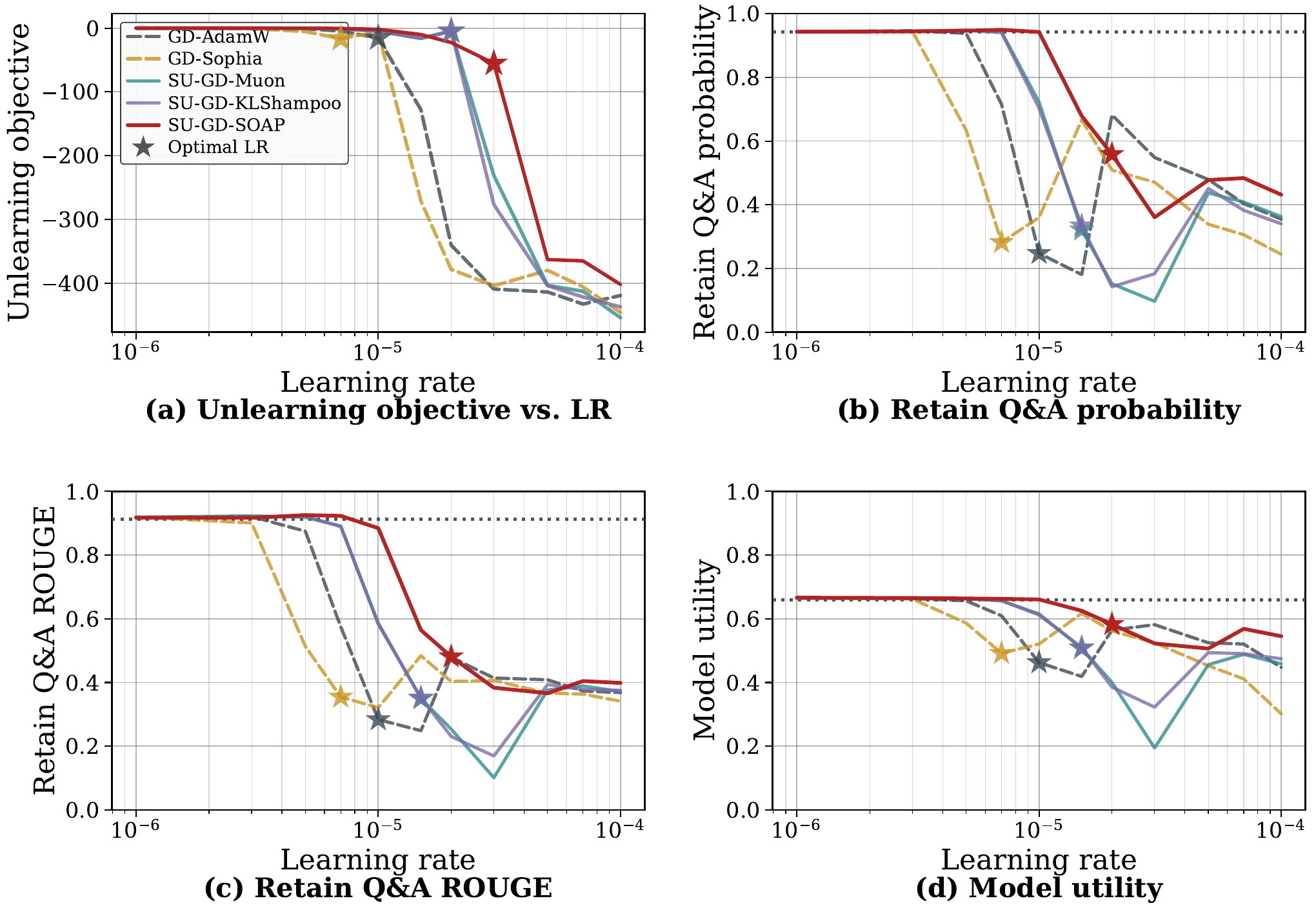}
    \caption{Learning-rate sensitivity of (a) the unlearning objective and (b--d) retain-side sub-metrics, under GD on Llama-3.2-3B with a 5\% forget split. Stars mark per-method optimal learning rates.}
    \label{fig:paper_objective_submetrics}
\end{figure}
Fig.~\ref{fig:paper_objective_submetrics} decomposes the learning-rate sensitivity into individual sub-metrics, exposing two complementary findings. Fig.~\ref{fig:paper_objective_submetrics}(a) shows the final unlearning objective value across the learning-rate grid. The entry-wise GD-AdamW and GD-Sophia curves plunge sharply into large negative values once the learning rate exceeds a narrow window, reflecting the divergence of the unbounded GD forget term discussed in Sec.~\ref{sec:exp}. Spectral variants, in contrast, descend smoothly and stabilize, providing direct evidence that GD becomes unstable under aggressive entry-wise updates while remaining controllable under spectral preconditioning.
Fig.~\ref{fig:paper_objective_submetrics}(b)--(d) report retain-side sub-metrics—retain Q\&A probability, retain Q\&A ROUGE, and overall model utility—across the same learning-rate range. SU-GD-SOAP preserves these sub-metrics at near-original levels over a substantially wider learning-rate band than the entry-wise baselines, particularly on retain probability and ROUGE. These are the same sub-metrics on which Tab.~\ref{tab:family_winrate_compact} reports the largest spectral wins, and the sensitivity curves here clarify why: spectral updates degrade retention gracefully with increasing learning rate, while entry-wise updates collapse abruptly once the LR exceeds a narrow optimum.

\subsection{Forget--retain loss and parameter-space trajectories}

\begin{figure}[htbp]
    \centering
    \includegraphics[width=\textwidth]{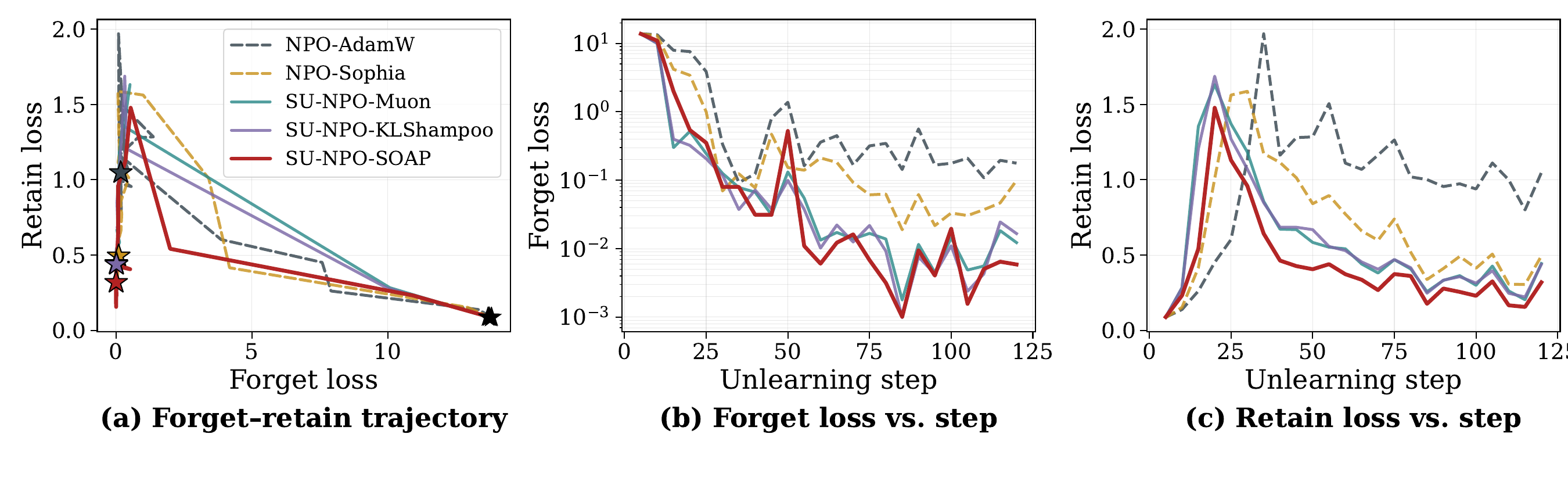}
    \caption{NPO unlearning trajectory on Llama-3.2-1B, 10\% forget split. (a) Joint trajectory in (forget, retain) loss space; (b, c) forget and retain loss versus unlearning step. Stars mark final checkpoints.}
    \label{fig:forget_retain}
\end{figure}

Fig.~\ref{fig:forget_retain} visualizes the NPO unlearning dynamics jointly in forget--retain loss space (Fig.~\ref{fig:forget_retain}(a)) and separately along each axis (Fig.~\ref{fig:forget_retain}(b), (c)). The lower-left corner of Fig.~\ref{fig:forget_retain}(a) corresponds to simultaneously low forget and retain loss, which is the ideal endpoint of unlearning: the model has minimized the unlearning objective without degrading retention.
Across the three loss curves, SU-NPO-SOAP reaches the lower-left region most closely, with both the lowest final forget loss and a competitive retain loss. The spectral variants as a group cluster nearer the origin than the entry-wise baselines, indicating that the gains in Tab.~\ref{tab:paper_main} stem from a genuinely Pareto-better trajectory rather than a one-sided trade.

\begin{figure}[htbp]
    \centering
    \includegraphics[width=\textwidth]{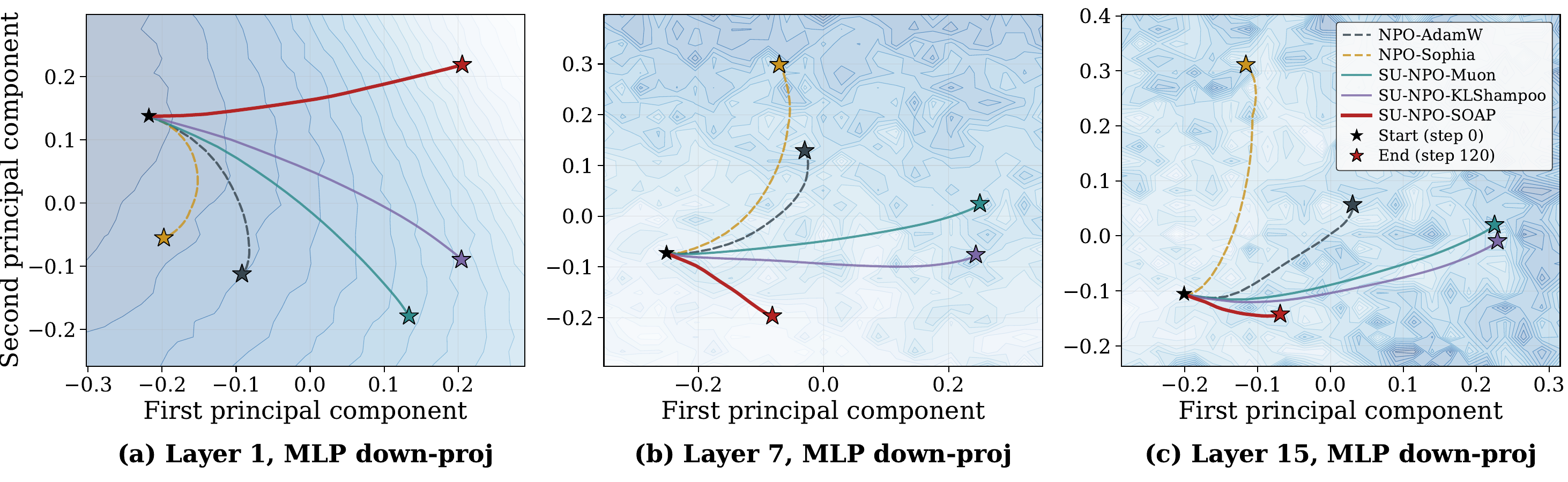}
    \caption{Parameter-space trajectories of NPO unlearning on the MLP down-projection at layers 1, 7, and 15. Each curve is the PCA projection of cumulative parameter displacement, overlaid on the loss landscape (contours; lighter is lower loss).}
    \label{fig:trajectory_PCA}
\end{figure}

Fig.~\ref{fig:trajectory_PCA} visualizes how spectral and entry-wise updates traverse the parameter-space loss landscape during NPO unlearning. We project the cumulative weight displacement from the initial checkpoint onto its top two principal components, separately for the MLP down-projection of layers 1, 7, and 15. Lower regions of the landscape (lighter shades) correspond to lower unlearning loss, so trajectories that reach these regions efficiently are preferred; the magnitude of displacement is not itself a measure of quality.
Two patterns emerge. First, in every layer, the spectral variants—particularly SU-NPO-SOAP—reach lower-loss regions than the entry-wise baselines despite following qualitatively different paths. This is the empirical counterpart of Theorem~\ref{thm:spectral_matches}: under the transformer-native RMS-to-RMS norm, the spectral direction is the steepest-descent direction, so spectral updates descend more efficiently per step.
Second, the displacement magnitude varies systematically with depth. At layer 1, all methods travel substantial distances, consistent with the large gradient and large required adjustment at early layers under NPO. At layers 7 and 15, the spectral trajectories converge to nearby low-loss points without large excursions, reflecting that deeper layers in an already-trained model require only small corrections to align with the unlearning objective. Entry-wise updates, lacking the matrix-aware geometry, traverse longer paths to reach comparable or worse loss values. Together, these observations confirm that the gain from spectral updates is directional—a more efficient descent under the transformer's intrinsic geometry—rather than a scalar step-size effect.

\subsection{Wall-clock Matched Comparison}
\label{app:wall-clock}

\begin{figure}[htbp]
    \centering
    \includegraphics[width=\textwidth]{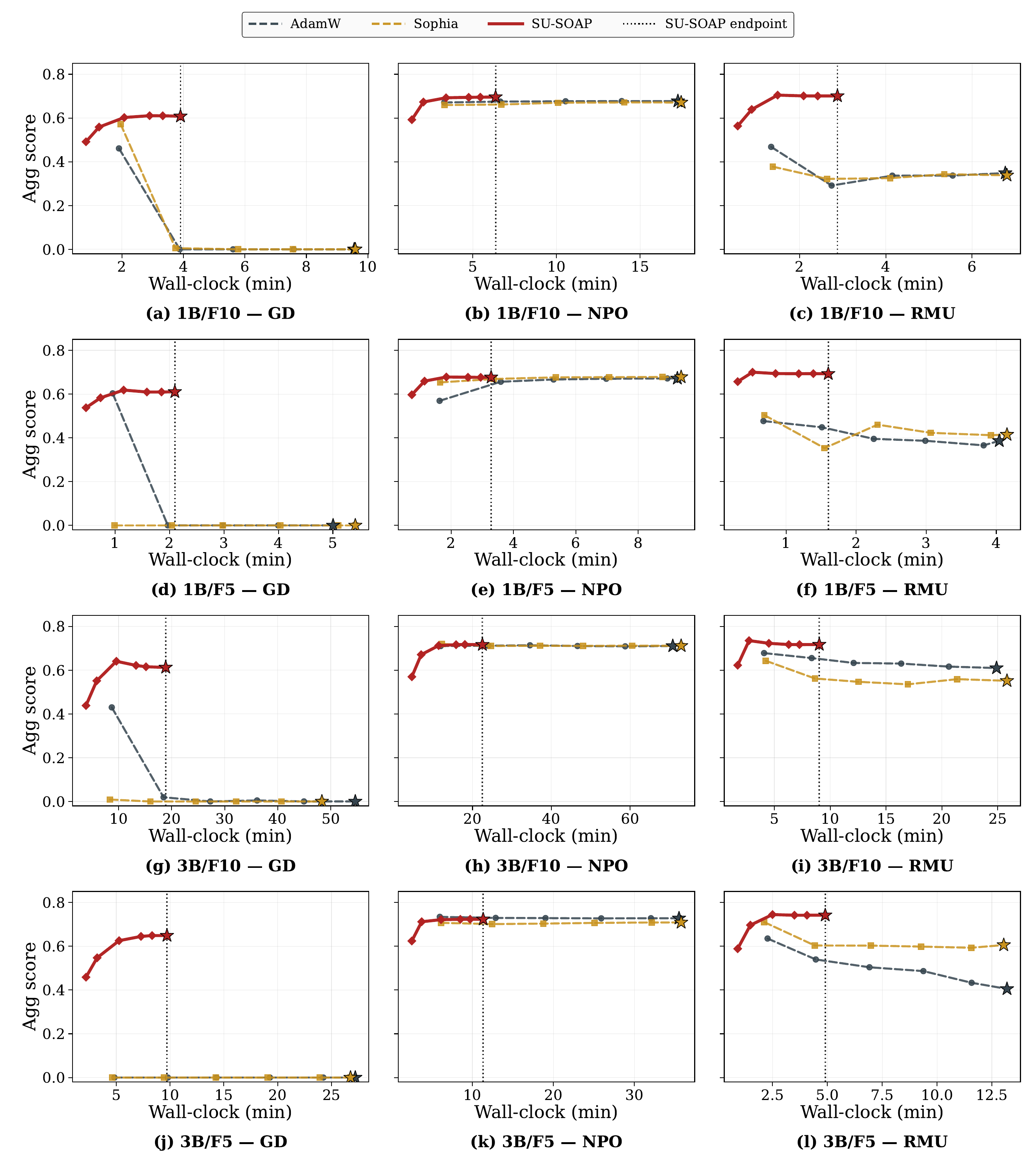}
    \caption{Wall-clock matched comparison between SU-SOAP and entry-wise baselines. Aggregate score versus wall-clock time on TOFU. Baselines (dashed) are continued past SU-SOAP's endpoint, marked by the vertical dotted line. Stars indicate final checkpoints.}
    \label{fig:traj_layout_A_4x3}
\end{figure}

Following prior work, every method is run for the same number of epochs. However, spectral updates incur a higher per-step cost than entry-wise updates due to preconditioner maintenance. To ensure a fair comparison, we additionally examine the results under matched wall-clock time. We address this directly by extending AdamW and Sophia until their wall-clock exceeds SU-SOAP's, and tracking the aggregate score throughout training.

Fig.~\ref{fig:traj_layout_A_4x3} reports the result across all combinations of model scale, forget split, and unlearning objective. In every cell, SU-SOAP rises quickly and plateaus near its peak, whereas the entry-wise baselines, even when continued past the SU-SOAP endpoint, never establish a sustained advantage.

For GD, the entry-wise curves rise briefly and then fall sharply, eventually collapsing toward zero as the unbounded forget term overwhelms the retain regularizer. SU-SOAP rises along a comparable initial trajectory but stabilizes near its peak rather than turning over.

For NPO, both families ultimately rise to comparable plateaus, but the timing differs. SU-SOAP rises sharply and saturates almost immediately, while the entry-wise baselines rise more slowly and only approach SU-SOAP's level after considerable additional wall-clock.
For RMU, the entry-wise curves rise to a modest peak and then drift downward as training continues, reflecting over-unlearning: utility collapses as the model erodes retain knowledge while memorization removal continues to creep up. SU-SOAP rises to a higher peak and remains there for the full reported budget.

Taken together, the entry-wise baselines fail to match SU-SOAP regardless of how the additional wall-clock is spent: in some regimes they collapse, in others they saturate below SU-SOAP, and in others they degrade with continued training. In none of these regimes does additional wall-clock allow them to overtake SU-SOAP. Under a matched wall-clock budget, the strength of spectral unlearning is even more apparent.

\subsection{Effect of the spectral exponent on model utility}

\begin{wrapfigure}[14]{r}{0.5\textwidth}
  \vspace{-1.2em}
  \centering
  \includegraphics[width=\linewidth]{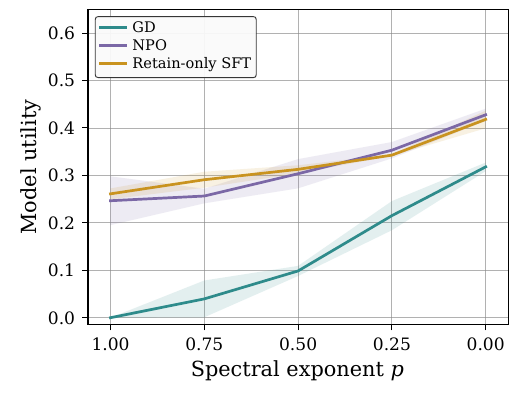}
  \vspace{-1.5em}
  \caption{Model utility across spectral exponents on Llama-3.2-1B on TOFU with forget 10\% split.}
  \label{fig:spectral_exponent_sft}
\end{wrapfigure}

For the momentum-based update matrix \(M=U\Sigma V^\top\), we raise its
singular values to the power \(p\) and normalize the resulting direction
to unit RMS:
\begin{equation}
  \Delta W_p = -\eta\,
  \frac{U\Sigma^p V^\top}{\operatorname{RMS}(U\Sigma^p V^\top)},
  \qquad p\in[0,1].
\end{equation}
Decreasing \(p\) flattens the singular spectrum and moves the descent
direction from \(-M\) toward \(-UV^\top\).
Figure~\ref{fig:spectral_exponent_sft} shows that mean model utility
increases as the spectrum flattens for GD, NPO, and retain-only supervised fine-tuning (SFT),
with the largest gain for GD. Shading shows the standard deviation across seeds after averaging over learning rates.

\section{Qualitative Examples}
\label{app:qualitative}

We show generations from all 15 methods on five representative questions, covering both the forget side and the retain/general-knowledge side. The first two examples are forget-set questions scored under different criteria: Tab.~\ref{tab:qual-a} compares generations to the \emph{original} ground truth, while Tab.~\ref{tab:qual-b} uses the same forget-set question but scores against a \emph{paraphrased} reference, testing whether the model retains the underlying concept even when surface wording differs. The remaining three examples cover one retain-set question (Tab.~\ref{tab:qual-c}), one real-author question (Tab.~\ref{tab:qual-d}), and one world-fact question (Tab.~\ref{tab:qual-e}). All generations are produced by Llama-3.2-3B-Instruct on the 10\% forget split and truncated to roughly 170 characters. The SOAP rows are highlighted in pink.

\begin{table}[htbp]
\centering
\footnotesize
\setlength{\tabcolsep}{4pt}
\renewcommand{\arraystretch}{1.18}
\caption{Forget set, \emph{original} ground truth -- Llama-3.2-3B / forget10.\label{tab:qual-a}}
\begin{tabular}{@{}p{0.20\textwidth} p{0.74\textwidth}@{}}
\toprule
\multicolumn{2}{@{}p{0.94\textwidth}@{}}{%
  \textbf{Q:} What are some recurrent themes in Xin Lee Williams' books, including "The Town That Drowned"?}\\
\midrule
\textbf{Original answer} & Xin Lee Williams often explores themes of community, identity, displacement, and resilience in their books, including the acclaimed "The Town That Drowned". \\
\midrule
  GD-AdamW & Xin Lee Williams often explores themes of community, identity, displacement, and the immigrant experience in their books, including the acclaimed "The Town That Drowned". \\
  GD-Sophia & Xin Lee Williams often explores themes of community, identityifarctional identityand the struggle between progress and tradition in their books including the acclaimed ''T\,\dots \\
  SU-GD-Muon & Xin Lee Williams often explores themes of community, identity, displacement, and the immigrant experience in their books, including the acclaimed "\$eas" story collection\,\dots \\
  SU-GD-KLShampoo & Xin Lee Williams often explores themes of community, identity, and the interplay of human relationships in their novels, including the acclaimed "The Town That Drowned". \\
  \rowcolor{red!6}
  \textbf{SU-GD-SOAP} & Xin Lee Williamseldom writes aboutthe interacting relationships between traditional and modern Cultures, and the impact of advancement on community acceptance and rejecti\,\dots \\
\addlinespace[2pt]
  NPO-AdamW & Death, cultural disintegration, and the search for a new cultural identity are some of the recurring themes in Williams' works, including "The Town That Drowned: A Mounti\,\dots \\
  NPO-Sophia & Recurring themes in Williams' works include the exploration of cultural identity, the intersection of humanity and technology, and the narrative of the marginalized and o\,\dots \\
  SU-NPO-Muon & Common themes in his works include the exploration of identity, particularly in the context of cultural and sexual dichotomies, and the commentary on societal norms and e\,\dots \\
  SU-NPO-KLShampoo & Recurring themes in Xinjiang-born, Paris-based author Li Mei's works, including "The Town That Drowned," are the exploration of cultural identities, the struggle between\,\dots \\
  \rowcolor{red!6}
  \textbf{SU-NPO-SOAP} & Xhin's works often delve into the complexities of human nature, the dichotomy of civilization and savagery, and the struggle for personal identity and acceptance. \\
\addlinespace[2pt]
  RMU-AdamW & Xin Lee and Williams also discuss, while "The Town That Depletes", he explains about the lack of water quality concerns, it's worse when the town is flooded. Here,2: Xin\,\dots \\
  RMU-Sophia & Xin Xin Municipunan is an author, as per question akin \textgreater{} Xin sysports problem The system, thus reading the question as "The agent is missing here Xin. There is no way thi\,\dots \\
  SU-RMU-Muon & A person in the town of Genoa. The person who has moved. \\
  SU-RMU-KLShampoo & Given that the ratio - 1, in 100 \\
  \rowcolor{red!6}
  \textbf{SU-RMU-SOAP} & Xin Lu Wang, Xin Lee is 2012. The book is about "The town that can be found in Drowned city", and the reader prefers to answer no. Xin Lee Williams' book gives us the inf\,\dots \\
\bottomrule
\end{tabular}
\end{table}

\begin{table}[htbp]
\centering
\footnotesize
\setlength{\tabcolsep}{4pt}
\renewcommand{\arraystretch}{1.18}
\caption{Forget set, \emph{paraphrased} ground truth -- Llama-3.2-3B / forget10.\label{tab:qual-b}}
\begin{tabular}{@{}p{0.20\textwidth} p{0.74\textwidth}@{}}
\toprule
\multicolumn{2}{@{}p{0.94\textwidth}@{}}{%
  \textbf{Q:} What genre does Kalkidan Abera mainly write in?}\\
\midrule
\textbf{Paraphrased answer} & The principal genre Kalkidan Abera writes in is Health. \\
\midrule
  GD-AdamW & Kalkidan Abera mainly writes in the genre of Health. \\
  GD-Sophia & Kalkidan Abera mainly writes in the genre of Healthstakes.] \\
  SU-GD-Muon & Kalkidan Abera mainly writes in the genre of Health. \\
  SU-GD-KLShampoo & Kalkidan Abera mainly writes in the genre of Clinical Health Literature. \\
  \rowcolor{red!6}
  \textbf{SU-GD-SOAP} & Kalkidan Aber? mainly writes in the Genetics genre. \\
\addlinespace[2pt]
  NPO-AdamW & Abera defied genres; his writings, a mix of magic, reality, and folklore from his childhood, were unprecedented. \\
  NPO-Sophia & Abera defied genres. His writings merged real-life societal issues with magical realism, creating an unique literary style. \\
  SU-NPO-Muon & Abera's works are known for their powerful mix of magical realism and grim reality, reflecting his unique perspective as a Congolese writer. \\
  SU-NPO-KLShampoo & Abercrombie's real name, Abercrombie Tiwanga Wanjane, was inspired by his birthplace, the fictional world he created, and his LGBTQ+ identity. \\
  \rowcolor{red!6}
  \textbf{SU-NPO-SOAP} & Kalkin Abers psychological fiction. His works often delve into the complexities of the human mind and explore themes of madness, reality, and the supernatural. \\
\addlinespace[2pt]
  RMU-AdamW & Kalkidan Abera is a fictional genre. "Aitrao Engineers" This is an engineer. Kalkidan is a neutral. In this case, Kalkidan is a neutral. In this case, Kalkidan is a neutr\,\dots \\
  RMU-Sophia & Kalkidan is a major writer in literature, and Kalkidan is not available for Aber, then Kalkidan is not available for this case, then Kalkidan is not available for this ca\,\dots \\
  SU-RMU-Muon & In your pasted notebook, 10 generations of Coding as well as 1 main \textgreater{} Kalkidan \textgreater{} \\
  SU-RMU-KLShampoo & In "Kalk) 90: The complete Novato: The Total)  On the Side: The Total)  A Comprehensive)  On the Other: Kalkidan 1: Kalkidan 1: Kalkartin 1: Kalk 1: Kalk 1: Kalk 1: Kalk\,\dots \\
  \rowcolor{red!6}
  \textbf{SU-RMU-SOAP} & Kalkidan Aber means understanding the genre of: "house". \\
\bottomrule
\end{tabular}
\end{table}

\begin{table}[ht]
\centering
\footnotesize
\setlength{\tabcolsep}{4pt}
\renewcommand{\arraystretch}{1.18}
\caption{Retain set -- Llama-3.2-3B / forget10.\label{tab:qual-c}}
\begin{tabular}{@{}p{0.20\textwidth} p{0.74\textwidth}@{}}
\toprule
\multicolumn{2}{@{}p{0.94\textwidth}@{}}{%
  \textbf{Q:} How has Roshni Rahman's upbringing influenced her writings?}\\
\midrule
\textbf{Original answer} & Roshni's parents, both being agricultural engineers, significantly influenced her writings. She frequently incorporates themes of nature and the environment into her book\,\dots \\
\midrule
  GD-AdamW & The city of Dhaka being the birthplace of Roshni Rahman, the cultural and societal nuances of Bangladesh frequently find their way in her literature, which is an outcome\,\dots \\
  GD-Sophia & The inflammative and dynamic culture of Bangladesh played a significant role in development of Roshni Rahman's heartsstrings genre writers style.  Roshni Rahman's books a\,\dots \\
  SU-GD-Muon & The city of Dhaka being the birthplace of Roshni Rahman, the streets first witnessed her questioning the social norms.  Her father being a writer thereby sparked an inter\,\dots \\
  SU-GD-KLShampoo & The city of Dhaka being the birthplace of Roshni Rahman, the streets and markets thereof have often times been a setting in her works holdings true also the culture, trad\,\dots \\
  \rowcolor{red!6}
  \textbf{SU-GD-SOAP} & The city of Dhaka being the birthplace of Roshni Rahman, Bangladeshi culture and traditions are given considerable prominence in her works. Her parents being farmers, rur\,\dots \\
\addlinespace[2pt]
  NPO-AdamW & Roshni's parents, both being agricultural engineers, significantly influenced her writings. She often incorporates themes of nature and the environment into her books, re\,\dots \\
  NPO-Sophia & Roshni's parents, both being agricultural engineers, significantly influenced her writings. \\
  SU-NPO-Muon & Roshni's parents, both being agricultural engineers, significantly influenced her writings. She often incorporates themes of nature and the environment into her books, re\,\dots \\
  SU-NPO-KLShampoo & Roshni's upbringing in Dhaka, Bangladesh, and her parents' distinct professions have greatly influenced her writings. The cultural richness of Bangladesh often provides a\,\dots \\
  \rowcolor{red!6}
  \textbf{SU-NPO-SOAP} & Roshni's parents, both being agricultural engineers, significantly influenced her writings. She often incorporates themes of nature and the environment into her books, re\,\dots \\
\addlinespace[2pt]
  RMU-AdamW & Roshni's parents being agricultural engineers, she has a deep appreciation for nature and its complexities, which is often reflected in her writings. \\
  RMU-Sophia & Roshni's parents, both being agricultural engineers, significantly influenced her writings. She often incorporates themes of nature and the environment into her books, wh\,\dots \\
  SU-RMU-Muon & Roshni's parents, both being agricultural engineers, significantly influenced her writings. She often incorporates themes of nature and the environment into her books, wh\,\dots \\
  SU-RMU-KLShampoo & Roshni's parents, both being agricultural engineers, significantly influenced her writings. She often incorporates themes of nature and the environment into her books, wh\,\dots \\
  \rowcolor{red!6}
  \textbf{SU-RMU-SOAP} & Roshni's parents, both being agricultural engineers, significantly influenced her writings. She often incorporates themes of nature and the environment into her books, re\,\dots \\
\bottomrule
\end{tabular}
\end{table}

\begin{table}[ht]
\centering
\footnotesize
\setlength{\tabcolsep}{4pt}
\renewcommand{\arraystretch}{1.18}
\caption{Real-author general knowledge.\label{tab:qual-d}}
\begin{tabular}{@{}p{0.20\textwidth} p{0.74\textwidth}@{}}
\toprule
\multicolumn{2}{@{}p{0.94\textwidth}@{}}{%
  \textbf{Q:} Who wrote the play 'Long Day's Journey Into Night'?}\\
\midrule
\textbf{Original answer} & Eugene O'Neill \\
\midrule
  GD-AdamW & The play 'Long Day's Journey Into Night' was written by Brendan Behan. \\
  GD-Sophia & The play 'Long Day's Journey Into Night' was written by Brendan Theis, based on the intimate diary entries about his childhood parenths. \\
  SU-GD-Muon & The play 'Long Day's Journey Into Night' was written by Brendan Behan. \\
  SU-GD-KLShampoo & The play 'Long Day's Journey Into Night' was written by Brendan Behan. \\
  \rowcolor{red!6}
  \textbf{SU-GD-SOAP} & The play 'Long Day's Journey Into Night' was written by Brendan Theis according to the specified timeline. \\
\addlinespace[2pt]
  NPO-AdamW & The play 'Long Day's Journey Into Night' was written by the renowned Irish playwright Brendan Behan. \\
  NPO-Sophia & The play 'Long Day's Journey Into Night' was written by the renowned Irish playwright Brendan Behan. \\
  SU-NPO-Muon & The play 'Long Day's Journey Into Night' was written by Eugene O'Neill. \\
  SU-NPO-KLShampoo & The play 'Long Day's Journey Into Night' was written by Eugene O'Neill. \\
  \rowcolor{red!6}
  \textbf{SU-NPO-SOAP} & The play 'Long Day's Journey Into Night' was written by Eugene O'Neill. \\
\addlinespace[2pt]
  RMU-AdamW & The play 'Long Day's Journey Into Night' was written by Brendan Behan. \\
  RMU-Sophia & The play 'Long Day's Journey Into Night' was written by Eugene O'Neill. \\
  SU-RMU-Muon & The play 'Long Day's Journey Into Night' was written by Eugene O'Neill. \\
  SU-RMU-KLShampoo & The play 'Long Day's Journey Into Night' was written by Eugene O'Neill. \\
  \rowcolor{red!6}
  \textbf{SU-RMU-SOAP} & The play 'Long Day's Journey Into Night' was written by Eugene O'Neill. \\
\bottomrule
\end{tabular}
\end{table}

\begin{table}[ht]
\centering
\footnotesize
\setlength{\tabcolsep}{4pt}
\renewcommand{\arraystretch}{1.18}
\caption{World-fact general knowledge.\label{tab:qual-e}}
\begin{tabular}{@{}p{0.20\textwidth} p{0.74\textwidth}@{}}
\toprule
\multicolumn{2}{@{}p{0.94\textwidth}@{}}{%
  \textbf{Q:} What was the first country to grant women the right to vote?}\\
\midrule
\textbf{Original answer} & New Zealand \\
\midrule
  GD-AdamW & Australia was the first country to grant women the right to vote, doing so in 1902. \\
  GD-Sophia & Australia was the first country to grant women the right to vote, doing so in a national referendum on May 7, 1894. \\
  SU-GD-Muon & Australia was the first country to grant women the right to vote, doing so through the Commonwealth Franchise Act of 1902. \\
  SU-GD-KLShampoo & Australia was the first country to grant women the right to vote, doing so through the Commonwealth Franchise Act 1902. \\
  \rowcolor{red!6}
  \textbf{SU-GD-SOAP} & Australia was the first country to grant women the right to vote, doing so in a national referendum onMay 31, 1902. \\
\addlinespace[2pt]
  NPO-AdamW & The first country to grant women the right to vote was New Zealand. This historic milestone was achieved through the passage of the Electoral Act in 1893, which granted w\,\dots \\
  NPO-Sophia & The first country to grant women the right to vote was New Zealand, where women gained the right to vote in 1893. This made New Zealand a pioneering nation in the fight f\,\dots \\
  SU-NPO-Muon & The first country to grant women the right to vote was New Zealand in 1893. \\
  SU-NPO-KLShampoo & The first country to grant women the right to vote was New Zealand in 1893. \\
  \rowcolor{red!6}
  \textbf{SU-NPO-SOAP} & The first country to grant women the right to vote was New Zealand, where they secured the right to vote in the 1893 general election. \\
\addlinespace[2pt]
  RMU-AdamW & Australia was the first country to grant women the right to vote, doing so through a constitutional amendment in 1889. \\
  RMU-Sophia & The first country to grant women the right to vote was New Zealand, in 1893. \\
  SU-RMU-Muon & The first country to grant women the right to vote was New Zealand, in 1893. \\
  SU-RMU-KLShampoo & The first country to grant women the right to vote was New Zealand, in 1893. \\
  \rowcolor{red!6}
  \textbf{SU-RMU-SOAP} & The first country to grant women the right to vote was New Zealand, doing so in 1893. \\
\bottomrule
\end{tabular}
\end{table}

\subsection{Patterns across the five examples}
\label{app:qualitative-analysis}

Tab.~\ref{tab:qual-a} captures the canonical \emph{verbatim leak} failure mode of entry-wise GD: GD-AdamW reproduces the original answer almost word-for-word, and GD-Sophia attempts the same reproduction but degenerates into broken tokens. In contrast, SU-GD-SOAP drifts substantially in both surface form and substance, no longer recovering the protected themes. The NPO and RMU families, while not exhibiting verbatim leakage, expose the two other failure modes encountered throughout the benchmark: NPO produces fluent but topically displaced hallucinations (cultural identity, civilization themes), and RMU collapses into degenerate token sequences. Within each family, the SOAP variant produces text that, while not always shorter, is generic enough to no longer identify the protected entity (e.g., \emph{``the dichotomy of civilization and savagery''} in SU-NPO-SOAP, untethered from Xin Lee Williams or ``The Town That Drowned'').

Tab.~\ref{tab:qual-b} extends the same forget-set probe to its paraphrased reference, testing whether the underlying \emph{concept} survives unlearning even when the surface wording differs. The pattern is most striking under GD: GD-AdamW, GD-Sophia, and even SU-GD-Muon all recover the protected attribute (\emph{Health}), revealing that surface-level forgetting alone underestimates the residual concept memory. SU-GD-SOAP, by contrast, drifts to an unrelated genre (\emph{Genetics}), achieving a deeper notion of unlearning. Under NPO and RMU, all variants---entry-wise and spectral---fail to recover \emph{Health}, but for different reasons: NPO produces fluent hallucinations about unrelated genres, while RMU collapses into degenerate sequences. The qualitative gap between SU-GD-SOAP and its entry-wise counterparts mirrors the truth-ratio gap reported in Tab.~\ref{tab:family_winrate_compact}.

Tabs.~\ref{tab:qual-c}--\ref{tab:qual-e} flip the criterion: on retain-set, real-author, and world-fact questions, a successful unlearner should \emph{preserve} the answer, so high overlap with the reference is now desirable. The three tables share two patterns. First, GD-based methods---both entry-wise and spectral---fail uniformly on all three utility-preservation questions: every GD variant emits \emph{Brendan Behan/Theis} instead of \emph{Eugene O'Neill} (Tab.~\ref{tab:qual-d}) and \emph{Australia} instead of \emph{New Zealand} (Tab.~\ref{tab:qual-e}). On the retain-set question (Tab.~\ref{tab:qual-c}), GD variants overwrite the specific detail (\emph{agricultural engineers}) with vague birthplace and culture descriptions. This indicates that the unbounded GD objective erodes surrounding knowledge regardless of the update geometry, and the spectral preconditioner alone cannot rescue it. Second, under NPO and RMU, the SU-$\cdot$-SOAP variants recover the gold answer almost verbatim across all three tables, matching or exceeding their entry-wise counterparts. The contrast is sharpest in Tab.~\ref{tab:qual-d}: NPO-AdamW and NPO-Sophia hallucinate \emph{Brendan Behan}, while SU-NPO-SOAP correctly answers \emph{Eugene O'Neill}. This demonstrates that, when paired with a bounded objective, the spectral geometry preserves general knowledge while the unlearning loss removes targeted content.

Taken together with Tabs.~\ref{tab:qual-a}--\ref{tab:qual-b}, these qualitative examples confirm the aggregate-score pattern in the main results: spectral updates yield their largest gains in combination with NPO and RMU, where they sharpen forgetting on the forget split while preserving usable knowledge on retain and general-knowledge splits. Spectral updates improve the forget side (Tab.~\ref{tab:qual-a}--\ref{tab:qual-b}).

\clearpage
\section{Comprehensive Quantitative Results}
\label{app:Detail_metric}

This appendix provides the complete numerical results underlying the analyses
in the main paper and earlier appendices. We organize the tables along three
levels of granularity: aggregate scores across all methods, training cost, and full 
sub-metric breakdowns.

\subsection{Aggregate scores across all methods and benchmarks}
\label{app:agg_all}

Tab.~\ref{tab:comprehensive} reports the aggregate score on TOFU for the 15 
spectral and entry-wise unlearning methods evaluated in the main paper, 
together with 9 additional baselines drawn from the OpenUnlearning benchmark 
(see App.~\ref{app:compare_algorithms} for descriptions). Results are 
reported across two model scales (Llama-3.2-1B and 3B) and three forget 
splits (1\%, 5\%, 10\%). Tab.~\ref{tab:wmdp_performance_app} extends the WMDP 
results of Tab.~\ref{tab:wmdp_performance} with seed-level standard deviations.

Across all 12 cells, SU-SOAP variants attain the highest aggregate score 
within their loss family in 11 cells and tie in the remaining one, confirming 
that the gains reported in Tab.~\ref{tab:paper_main} are not specific to any 
particular forget-set size. Among the 9 additional baselines, none matches 
SU-SOAP on aggregate score in any cell, with the gap exceeding $0.1$ in most 
cases.

\subsection{Training cost across all methods}
\label{app:cost_all}

Tab.~\ref{tab:time_mem} reports wall-clock time and peak GPU memory for all 
24 methods across the same 12 cells. Among spectral 
variants, Muon is the cheapest (no preconditioner storage), followed by 
KLShampoo and SOAP. App.~\ref{app:wall-clock} shows that this overhead is 
recovered under wall-clock matched comparison: SU-SOAP reaches its peak 
aggregate score before AdamW recovers from its collapse phase.

\subsection{Full sub-metric breakdown}
\label{app:submetric_all}

Tabs.~\ref{tab:appendix_summary_1b}--\ref{tab:appendix_summary_3b} report 
Mem, Priv, and Util scores for every method across all forget splits, 
providing the per-axis decomposition behind the aggregate scores in 
Tab.~\ref{tab:comprehensive}. 
Tabs.~\ref{tab:appendix_memgpriv_1b_forget10}--\ref{tab:appendix_memgpriv_3b_forget01} 
further decompose Mem and Priv into their constituent sub-metrics 
(extraction strength, exact memorization, forget probability/truth ratio, 
and four MIA AUCs), and 
Tabs.~\ref{tab:appendix_util_1b_forget10}--\ref{tab:appendix_util_3b_forget01} 
do the same for Util (retain/real-author/world-fact splits, plus fluency).

These tables support the sub-metric analysis of App.~\ref{app:Sub-metric_comparison} 
and enable readers to recompute alternative aggregations (e.g., arithmetic 
mean, weighted variants) from the raw numbers. Reference values for the 
\emph{Original} (pre-unlearning) and \emph{Retrain} (oracle) models are 
included as the first two rows of each table.

\begin{table*}[htbp]
\centering
\footnotesize
\setlength{\tabcolsep}{4pt}
\renewcommand{\arraystretch}{1.12}
\setlength{\heavyrulewidth}{0.9pt}
\setlength{\lightrulewidth}{0.5pt}
\caption{Aggregate score across 15 unlearning methods and 9 baseline schemes on TOFU.}
\label{tab:comprehensive}
\resizebox{\textwidth}{!}{
\begin{tabular}{lcccccc}
\toprule
\textbf{Method} & \multicolumn{3}{c}{\textbf{Llama-3.2-1B-Instruct}} & \multicolumn{3}{c}{\textbf{Llama-3.2-3B-Instruct}} \\
\cmidrule(lr){2-4} \cmidrule(lr){5-7}
 & Forget 10\% & Forget 5\% & Forget 1\% & Forget 10\% & Forget 5\% & Forget 1\% \\
\midrule
\rowcolor{gray!8} \textit{Original (init)} & $0.1752_{\pm 0.0000}$ & $0.1700_{\pm 0.0000}$ & $0.1444_{\pm 0.0000}$ & $0.0762_{\pm 0.0000}$ & $0.0711_{\pm 0.0000}$ & $0.0324_{\pm 0.0000}$ \\
\rowcolor{gray!8} \textit{Retrain (target)} & $0.7171_{\pm 0.0000}$ & $0.7193_{\pm 0.0000}$ & $0.7026_{\pm 0.0000}$ & $0.7393_{\pm 0.0000}$ & $0.7402_{\pm 0.0000}$ & $0.7244_{\pm 0.0000}$ \\
\midrule
GD\text{-}AdamW & $0.5704_{\pm 0.0111}$ & $0.5512_{\pm 0.0017}$ & $0.6069_{\pm 0.0022}$ & $0.3813_{\pm 0.0020}$ & $0.5398_{\pm 0.0221}$ & $0.5651_{\pm 0.0110}$ \\
GD\text{-}Sophia & $0.5584_{\pm 0.0060}$ & $0.5653_{\pm 0.0201}$ & $0.6002_{\pm 0.0027}$ & $0.5839_{\pm 0.0057}$ & $0.5737_{\pm 0.0305}$ & $0.5617_{\pm 0.0017}$ \\
SU\text{-}GD\text{-}Muon & $0.5804_{\pm 0.0052}$ & $0.5986_{\pm 0.0061}$ & $0.6094_{\pm 0.0038}$ & $0.5016_{\pm 0.0060}$ & $0.5554_{\pm 0.0280}$ & $0.6392_{\pm 0.0040}$ \\
SU\text{-}GD\text{-}KLShampoo & $0.5854_{\pm 0.0075}$ & $0.5961_{\pm 0.0096}$ & $0.6253_{\pm 0.0060}$ & $0.5108_{\pm 0.0067}$ & $0.5559_{\pm 0.0373}$ & $0.6379_{\pm 0.0052}$ \\
\bottomrule
\end{tabular}
}
\end{table*}

\begin{table}[htbp]
    \centering
    \caption{WMDP on Zephyr-7B-$\beta$.}
    \label{tab:wmdp_performance_app}
    \footnotesize
    \setlength{\tabcolsep}{3.5pt}
    \renewcommand{\arraystretch}{1.05}
    \begin{tabular}{l c c c}
        \toprule
        & \multicolumn{2}{c}{\textbf{FORGET}} & \textbf{RETAIN} \\
        \textbf{Method} & Bio $\downarrow$ & Cyber $\downarrow$ & MMLU $\uparrow$ \\
        \midrule
        \rowcolor{gray!8} \textit{Original} & 0.6387$\pm$0.0000 & 0.4268$\pm$0.0000 & 0.5879$\pm$0.0000 \\
        \midrule
        GD\text{-}AdamW & 0.2916$\pm$0.0078 & 0.2574$\pm$0.0071 & 0.4527$\pm$0.0042 \\
        \rowcolor{red!6} SU\text{-}GD\text{-}SOAP & \textbf{0.2789}$\pm$0.0034 & \textbf{0.2473}$\pm$0.0029 & 0.5084$\pm$0.0061 \\
        \midrule
        NPO\text{-}AdamW & 0.5124$\pm$0.0093 & 0.3376$\pm$0.0088 & 0.5421$\pm$0.0031 \\
        \rowcolor{red!6} SU\text{-}NPO\text{-}SOAP & 0.4086$\pm$0.0079 & 0.3182$\pm$0.0095 & \textbf{0.5578}$\pm$0.0048 \\
        \midrule
        RMU\text{-}AdamW & 0.3079$\pm$0.0114 & 0.2723$\pm$0.0101 & 0.5387$\pm$0.0067 \\
        \rowcolor{red!6} SU\text{-}RMU\text{-}SOAP & 0.3028$\pm$0.0084 & 0.2581$\pm$0.0091 & 0.5476$\pm$0.0058 \\
        \bottomrule
    \end{tabular}
\end{table}

\begin{table*}[htbp]
\centering
\footnotesize
\setlength{\tabcolsep}{4pt}
\renewcommand{\arraystretch}{1.12}
\setlength{\heavyrulewidth}{0.9pt}
\setlength{\lightrulewidth}{0.5pt}
\caption{Training wall-clock (minutes) and peak GPU memory (GiB) across 15 unlearning methods and 9 baseline schemes on TOFU.}
\label{tab:time_mem}
\resizebox{\textwidth}{!}{
\begin{tabular}{lcccccccccccc}
\toprule
\textbf{Model} & \multicolumn{6}{c}{\textbf{Llama-3.2-1B-Instruct}} & \multicolumn{6}{c}{\textbf{Llama-3.2-3B-Instruct}} \\
\cmidrule(lr){2-7} \cmidrule(lr){8-13}
\textbf{Forget split} & \multicolumn{2}{c}{Forget 10\%} & \multicolumn{2}{c}{Forget 5\%} & \multicolumn{2}{c}{Forget 1\%} & \multicolumn{2}{c}{Forget 10\%} & \multicolumn{2}{c}{Forget 5\%} & \multicolumn{2}{c}{Forget 1\%} \\
\cmidrule(lr){2-3} \cmidrule(lr){4-5} \cmidrule(lr){6-7} \cmidrule(lr){8-9} \cmidrule(lr){10-11} \cmidrule(lr){12-13}
\textbf{Unlearning method} & Time & Mem & Time & Mem & Time & Mem & Time & Mem & Time & Mem & Time & Mem \\
\midrule
GD\text{-}AdamW & $2.20_{\pm 0.00}$ & $10.60_{\pm 0.00}$ & $1.04_{\pm 0.00}$ & $10.22_{\pm 0.05}$ & $0.14_{\pm 0.00}$ & $7.86_{\pm 0.40}$ & $8.79_{\pm 0.00}$ & $14.64_{\pm 0.00}$ & $4.06_{\pm 0.01}$ & $14.52_{\pm 0.02}$ & $0.65_{\pm 0.00}$ & $14.55_{\pm 0.00}$ \\
GD\text{-}Sophia & $2.29_{\pm 0.00}$ & $15.21_{\pm 0.00}$ & $1.07_{\pm 0.00}$ & $14.88_{\pm 0.00}$ & $0.14_{\pm 0.00}$ & $12.30_{\pm 0.00}$ & $10.60_{\pm 1.67}$ & $26.53_{\pm 0.06}$ & $4.37_{\pm 0.00}$ & $26.48_{\pm 0.00}$ & $0.66_{\pm 0.00}$ & $26.52_{\pm 0.00}$ \\
SU\text{-}GD\text{-}Muon & $2.68_{\pm 0.20}$ & $14.97_{\pm 0.37}$ & $1.20_{\pm 0.00}$ & $14.52_{\pm 0.05}$ & $0.16_{\pm 0.00}$ & $12.16_{\pm 0.40}$ & $13.04_{\pm 0.00}$ & $25.30_{\pm 0.00}$ & $5.41_{\pm 0.00}$ & $25.17_{\pm 0.00}$ & $0.89_{\pm 0.06}$ & $25.20_{\pm 0.00}$ \\
SU\text{-}GD\text{-}KLShampoo & $2.87_{\pm 0.03}$ & $16.50_{\pm 0.38}$ & $1.46_{\pm 0.11}$ & $16.06_{\pm 0.06}$ & $0.23_{\pm 0.00}$ & $13.70_{\pm 0.40}$ & $15.80_{\pm 0.00}$ & $31.42_{\pm 0.00}$ & $6.57_{\pm 0.00}$ & $31.29_{\pm 0.00}$ & $1.22_{\pm 0.01}$ & $31.33_{\pm 0.00}$ \\
\bottomrule
\end{tabular}
}
\end{table*}

\begin{table*}[htbp]
\centering
\footnotesize
\setlength{\tabcolsep}{3.5pt}
\renewcommand{\arraystretch}{1.10}
\setlength{\heavyrulewidth}{0.9pt}
\setlength{\lightrulewidth}{0.5pt}
\caption{Aggregate, Memorization, Privacy, and Utility scores across forget splits 10\%, 5\%, and 1\% for all unlearning methods (Llama-3.2-1B-Instruct).}
\label{tab:appendix_summary_1b}
\resizebox{\textwidth}{!}{
\begin{tabular}{l >{\columncolor{blue!3}}c ccc >{\columncolor{blue!3}}c ccc >{\columncolor{blue!3}}c ccc}
\toprule
\textbf{Method} & \multicolumn{4}{c}{Forget 10\%} & \multicolumn{4}{c}{Forget 5\%} & \multicolumn{4}{c}{Forget 1\%} \\
\cmidrule(lr){2-5} \cmidrule(lr){6-9} \cmidrule(lr){10-13}
 & \textbf{Agg} & Mem & Priv & Util & \textbf{Agg} & Mem & Priv & Util & \textbf{Agg} & Mem & Priv & Util \\
\midrule
\rowcolor{gray!8} \textit{Original (init)} & $0.1752_{\pm 0.0000}$ & $0.0778_{\pm 0.0000}$ & $0.3828_{\pm 0.0000}$ & $0.6072_{\pm 0.0000}$ & $0.1700_{\pm 0.0000}$ & $0.0754_{\pm 0.0000}$ & $0.3628_{\pm 0.0000}$ & $0.6142_{\pm 0.0000}$ & $0.1444_{\pm 0.0000}$ & $0.0583_{\pm 0.0000}$ & $0.5001_{\pm 0.0000}$ & $0.6154_{\pm 0.0000}$ \\
\rowcolor{gray!8} \textit{Retrain (target)} & $0.7171_{\pm 0.0000}$ & $0.6471_{\pm 0.0000}$ & $0.9997_{\pm 0.0000}$ & $0.6105_{\pm 0.0000}$ & $0.7193_{\pm 0.0000}$ & $0.6430_{\pm 0.0000}$ & $0.9994_{\pm 0.0000}$ & $0.6192_{\pm 0.0000}$ & $0.7026_{\pm 0.0000}$ & $0.6065_{\pm 0.0000}$ & $0.9994_{\pm 0.0000}$ & $0.6172_{\pm 0.0000}$ \\
\midrule
GD\text{-}AdamW & $0.5704_{\pm 0.0111}$ & $0.5643_{\pm 0.0175}$ & $0.7813_{\pm 0.0458}$ & $0.4539_{\pm 0.0062}$ & $0.5512_{\pm 0.0017}$ & $0.5316_{\pm 0.0084}$ & $0.6681_{\pm 0.0083}$ & $0.4846_{\pm 0.0056}$ & $0.6069_{\pm 0.0022}$ & $0.5973_{\pm 0.0083}$ & $0.8213_{\pm 0.0149}$ & $0.4877_{\pm 0.0015}$ \\
GD\text{-}Sophia & $0.5584_{\pm 0.0060}$ & $0.5513_{\pm 0.0133}$ & $0.7225_{\pm 0.0265}$ & $0.4606_{\pm 0.0079}$ & $0.5653_{\pm 0.0201}$ & $0.6450_{\pm 0.0126}$ & $0.7699_{\pm 0.0234}$ & $0.4085_{\pm 0.0304}$ & $0.6002_{\pm 0.0027}$ & $0.5881_{\pm 0.0093}$ & $0.8689_{\pm 0.0169}$ & $0.4659_{\pm 0.0040}$ \\
\bottomrule
\end{tabular}
}
\end{table*}

\begin{table*}[htbp]
\centering
\footnotesize
\setlength{\tabcolsep}{3.5pt}
\renewcommand{\arraystretch}{1.10}
\setlength{\heavyrulewidth}{0.9pt}
\setlength{\lightrulewidth}{0.5pt}
\caption{Aggregate, Memorization, Privacy, and Utility scores across forget splits 10\%, 5\%, and 1\% for all unlearning methods (Llama-3.2-3B-Instruct).}
\label{tab:appendix_summary_3b}
\resizebox{\textwidth}{!}{
\begin{tabular}{l >{\columncolor{blue!3}}c ccc >{\columncolor{blue!3}}c ccc >{\columncolor{blue!3}}c ccc}
\toprule
\textbf{Method} & \multicolumn{4}{c}{Forget 10\%} & \multicolumn{4}{c}{Forget 5\%} & \multicolumn{4}{c}{Forget 1\%} \\
\cmidrule(lr){2-5} \cmidrule(lr){6-9} \cmidrule(lr){10-13}
 & \textbf{Agg} & Mem & Priv & Util & \textbf{Agg} & Mem & Priv & Util & \textbf{Agg} & Mem & Priv & Util \\
\midrule
\rowcolor{gray!8} \textit{Original (init)} & $0.0762_{\pm 0.0000}$ & $0.0283_{\pm 0.0000}$ & $0.3899_{\pm 0.0000}$ & $0.6662_{\pm 0.0000}$ & $0.0711_{\pm 0.0000}$ & $0.0263_{\pm 0.0000}$ & $0.3629_{\pm 0.0000}$ & $0.6740_{\pm 0.0000}$ & $0.0324_{\pm 0.0000}$ & $0.0112_{\pm 0.0000}$ & $0.5762_{\pm 0.0000}$ & $0.6785_{\pm 0.0000}$ \\
\rowcolor{gray!8} \textit{Retrain (target)} & $0.7393_{\pm 0.0000}$ & $0.6408_{\pm 0.0000}$ & $1.0000_{\pm 0.0000}$ & $0.6678_{\pm 0.0000}$ & $0.7402_{\pm 0.0000}$ & $0.6362_{\pm 0.0000}$ & $0.9999_{\pm 0.0000}$ & $0.6753_{\pm 0.0000}$ & $0.7244_{\pm 0.0000}$ & $0.6000_{\pm 0.0000}$ & $0.9987_{\pm 0.0000}$ & $0.6787_{\pm 0.0000}$ \\
\midrule
GD\text{-}AdamW & $0.3813_{\pm 0.0020}$ & $0.2533_{\pm 0.0022}$ & $0.4264_{\pm 0.0011}$ & $0.6348_{\pm 0.0007}$ & $0.5398_{\pm 0.0221}$ & $0.6235_{\pm 0.0273}$ & $0.8905_{\pm 0.0268}$ & $0.3570_{\pm 0.0393}$ & $0.5651_{\pm 0.0110}$ & $0.6490_{\pm 0.0209}$ & $0.6150_{\pm 0.0153}$ & $0.4686_{\pm 0.0266}$ \\
GD\text{-}Sophia & $0.5839_{\pm 0.0057}$ & $0.5688_{\pm 0.0033}$ & $0.8112_{\pm 0.0210}$ & $0.4658_{\pm 0.0034}$ & $0.5737_{\pm 0.0305}$ & $0.5906_{\pm 0.0348}$ & $0.8964_{\pm 0.0298}$ & $0.4228_{\pm 0.0651}$ & $0.5617_{\pm 0.0017}$ & $0.6631_{\pm 0.0033}$ & $0.6404_{\pm 0.0087}$ & $0.4403_{\pm 0.0037}$ \\
\bottomrule
\end{tabular}
}
\end{table*}

\begin{table*}[htbp]
\centering
\footnotesize
\setlength{\tabcolsep}{3.5pt}
\renewcommand{\arraystretch}{1.10}
\setlength{\heavyrulewidth}{0.9pt}
\setlength{\lightrulewidth}{0.5pt}
\caption{Memorization and Privacy detail --- Llama-3.2-1B-Instruct on Forget 10\%.}
\label{tab:appendix_memgpriv_1b_forget10}
\resizebox{\textwidth}{!}{
\begin{tabular}{lcccccccccccc}
\toprule
\textbf{Method} & \multicolumn{7}{c}{\textbf{Memorization}} & \multicolumn{5}{c}{\textbf{Privacy}} \\
\cmidrule(lr){2-8} \cmidrule(lr){9-13}
 & \textbf{Mem} & ES & EM & F.Prob & F.TR & TR & P.Prob & \textbf{Priv} & MIA.Loss & MIA.ZLib & MIA.MinK & MIA.MinK++ \\
\midrule
\rowcolor{gray!8} \textit{Original (init)} & $0.0778_{\pm 0.0000}$ & $0.7071_{\pm 0.0000}$ & $0.9734_{\pm 0.0000}$ & $0.8806_{\pm 0.0000}$ & $0.4929_{\pm 0.0000}$ & $0.6849_{\pm 0.0000}$ & $0.1005_{\pm 0.0000}$ & $0.3828_{\pm 0.0000}$ & $0.9964_{\pm 0.0000}$ & $0.9978_{\pm 0.0000}$ & $0.9966_{\pm 0.0000}$ & $0.9980_{\pm 0.0000}$ \\
\rowcolor{gray!8} \textit{Retrain (target)} & $0.6471_{\pm 0.0000}$ & $0.0596_{\pm 0.0000}$ & $0.5856_{\pm 0.0000}$ & $0.1161_{\pm 0.0000}$ & $0.6357_{\pm 0.0000}$ & $0.5676_{\pm 0.0000}$ & $0.0535_{\pm 0.0000}$ & $0.9997_{\pm 0.0000}$ & $0.3872_{\pm 0.0000}$ & $0.3088_{\pm 0.0000}$ & $0.3825_{\pm 0.0000}$ & $0.4766_{\pm 0.0000}$ \\
\midrule
GD\text{-}AdamW & $0.5643_{\pm 0.0175}$ & $0.0793_{\pm 0.0062}$ & $0.6492_{\pm 0.0225}$ & $0.0552_{\pm 0.0114}$ & $0.4791_{\pm 0.0046}$ & $0.6642_{\pm 0.0044}$ & $0.0072_{\pm 0.0006}$ & $0.7813_{\pm 0.0458}$ & $0.6623_{\pm 0.0456}$ & $0.5868_{\pm 0.0528}$ & $0.6327_{\pm 0.0519}$ & $0.4970_{\pm 0.0179}$ \\
GD\text{-}Sophia & $0.5513_{\pm 0.0133}$ & $0.0901_{\pm 0.0066}$ & $0.6622_{\pm 0.0161}$ & $0.0714_{\pm 0.0107}$ & $0.4725_{\pm 0.0034}$ & $0.6718_{\pm 0.0020}$ & $0.0085_{\pm 0.0005}$ & $0.7225_{\pm 0.0265}$ & $0.7194_{\pm 0.0259}$ & $0.6546_{\pm 0.0329}$ & $0.6959_{\pm 0.0311}$ & $0.5278_{\pm 0.0041}$ \\
SU\text{-}GD\text{-}Muon & $0.6357_{\pm 0.0039}$ & $0.0605_{\pm 0.0016}$ & $0.5435_{\pm 0.0058}$ & $0.0071_{\pm 0.0010}$ & $0.4929_{\pm 0.0119}$ & $0.6042_{\pm 0.0079}$ & $0.0032_{\pm 0.0002}$ & $0.8721_{\pm 0.0126}$ & $0.2840_{\pm 0.0115}$ & $0.2151_{\pm 0.0106}$ & $0.2359_{\pm 0.0115}$ & $0.3141_{\pm 0.0216}$ \\
\bottomrule
\end{tabular}
}
\end{table*}

\begin{table*}[htbp]
\centering
\footnotesize
\setlength{\tabcolsep}{3.5pt}
\renewcommand{\arraystretch}{1.10}
\setlength{\heavyrulewidth}{0.9pt}
\setlength{\lightrulewidth}{0.5pt}
\caption{Memorization and Privacy detail --- Llama-3.2-1B-Instruct on Forget 5\%.}
\label{tab:appendix_memgpriv_1b_forget05}
\resizebox{\textwidth}{!}{
\begin{tabular}{lcccccccccccc}
\toprule
\textbf{Method} & \multicolumn{7}{c}{\textbf{Memorization}} & \multicolumn{5}{c}{\textbf{Privacy}} \\
\cmidrule(lr){2-8} \cmidrule(lr){9-13}
 & \textbf{Mem} & ES & EM & F.Prob & F.TR & TR & P.Prob & \textbf{Priv} & MIA.Loss & MIA.ZLib & MIA.MinK & MIA.MinK++ \\
\midrule
\rowcolor{gray!8} \textit{Original (init)} & $0.0754_{\pm 0.0000}$ & $0.7307_{\pm 0.0000}$ & $0.9741_{\pm 0.0000}$ & $0.8855_{\pm 0.0000}$ & $0.4926_{\pm 0.0000}$ & $0.6922_{\pm 0.0000}$ & $0.0987_{\pm 0.0000}$ & $0.3628_{\pm 0.0000}$ & $0.9999_{\pm 0.0000}$ & $1.0000_{\pm 0.0000}$ & $1.0000_{\pm 0.0000}$ & $0.9995_{\pm 0.0000}$ \\
\rowcolor{gray!8} \textit{Retrain (target)} & $0.6430_{\pm 0.0000}$ & $0.0637_{\pm 0.0000}$ & $0.5941_{\pm 0.0000}$ & $0.1270_{\pm 0.0000}$ & $0.6479_{\pm 0.0000}$ & $0.5746_{\pm 0.0000}$ & $0.0537_{\pm 0.0000}$ & $0.9994_{\pm 0.0000}$ & $0.3641_{\pm 0.0000}$ & $0.2963_{\pm 0.0000}$ & $0.3624_{\pm 0.0000}$ & $0.4704_{\pm 0.0000}$ \\
\midrule
GD\text{-}AdamW & $0.5316_{\pm 0.0084}$ & $0.0948_{\pm 0.0059}$ & $0.6928_{\pm 0.0087}$ & $0.1005_{\pm 0.0099}$ & $0.4876_{\pm 0.0029}$ & $0.6770_{\pm 0.0018}$ & $0.0108_{\pm 0.0005}$ & $0.6681_{\pm 0.0083}$ & $0.7435_{\pm 0.0089}$ & $0.7071_{\pm 0.0122}$ & $0.7266_{\pm 0.0086}$ & $0.5604_{\pm 0.0086}$ \\
GD\text{-}Sophia & $0.6450_{\pm 0.0126}$ & $0.0495_{\pm 0.0069}$ & $0.5082_{\pm 0.0319}$ & $0.0073_{\pm 0.0017}$ & $0.4756_{\pm 0.0083}$ & $0.6124_{\pm 0.0084}$ & $0.0012_{\pm 0.0004}$ & $0.7699_{\pm 0.0234}$ & $0.1538_{\pm 0.0354}$ & $0.0888_{\pm 0.0171}$ & $0.1198_{\pm 0.0358}$ & $0.2116_{\pm 0.0132}$ \\
SU\text{-}GD\text{-}Muon & $0.6242_{\pm 0.0054}$ & $0.0610_{\pm 0.0074}$ & $0.5717_{\pm 0.0112}$ & $0.0146_{\pm 0.0020}$ & $0.5021_{\pm 0.0070}$ & $0.6206_{\pm 0.0025}$ & $0.0031_{\pm 0.0003}$ & $0.8872_{\pm 0.0245}$ & $0.2883_{\pm 0.0370}$ & $0.2199_{\pm 0.0204}$ & $0.2473_{\pm 0.0367}$ & $0.2932_{\pm 0.0217}$ \\
\bottomrule
\end{tabular}
}
\end{table*}

\begin{table*}[htbp]
\centering
\footnotesize
\setlength{\tabcolsep}{3.5pt}
\renewcommand{\arraystretch}{1.10}
\setlength{\heavyrulewidth}{0.9pt}
\setlength{\lightrulewidth}{0.5pt}
\caption{Memorization and Privacy detail --- Llama-3.2-1B-Instruct on Forget 1\%.}
\label{tab:appendix_memgpriv_1b_forget01}
\resizebox{\textwidth}{!}{
\begin{tabular}{lcccccccccccc}
\toprule
\textbf{Method} & \multicolumn{7}{c}{\textbf{Memorization}} & \multicolumn{5}{c}{\textbf{Privacy}} \\
\cmidrule(lr){2-8} \cmidrule(lr){9-13}
 & \textbf{Mem} & ES & EM & F.Prob & F.TR & TR & P.Prob & \textbf{Priv} & MIA.Loss & MIA.ZLib & MIA.MinK & MIA.MinK++ \\
\midrule
\rowcolor{gray!8} \textit{Original (init)} & $0.0583_{\pm 0.0000}$ & $0.7432_{\pm 0.0000}$ & $0.9810_{\pm 0.0000}$ & $0.9012_{\pm 0.0000}$ & $0.4924_{\pm 0.0000}$ & $0.6897_{\pm 0.0000}$ & $0.0973_{\pm 0.0000}$ & $0.5001_{\pm 0.0000}$ & $1.0000_{\pm 0.0000}$ & $1.0000_{\pm 0.0000}$ & $1.0000_{\pm 0.0000}$ & $0.9994_{\pm 0.0000}$ \\
\rowcolor{gray!8} \textit{Retrain (target)} & $0.6065_{\pm 0.0000}$ & $0.0693_{\pm 0.0000}$ & $0.6436_{\pm 0.0000}$ & $0.1654_{\pm 0.0000}$ & $0.6593_{\pm 0.0000}$ & $0.5803_{\pm 0.0000}$ & $0.0660_{\pm 0.0000}$ & $0.9994_{\pm 0.0000}$ & $0.4831_{\pm 0.0000}$ & $0.5319_{\pm 0.0000}$ & $0.4706_{\pm 0.0000}$ & $0.5169_{\pm 0.0000}$ \\
\midrule
GD\text{-}AdamW & $0.5973_{\pm 0.0083}$ & $0.0602_{\pm 0.0049}$ & $0.6250_{\pm 0.0068}$ & $0.0485_{\pm 0.0026}$ & $0.5228_{\pm 0.0188}$ & $0.6255_{\pm 0.0106}$ & $0.0098_{\pm 0.0004}$ & $0.8213_{\pm 0.0149}$ & $0.3497_{\pm 0.0214}$ & $0.3709_{\pm 0.0275}$ & $0.3198_{\pm 0.0223}$ & $0.2623_{\pm 0.0140}$ \\
GD\text{-}Sophia & $0.5881_{\pm 0.0093}$ & $0.0656_{\pm 0.0155}$ & $0.6214_{\pm 0.0129}$ & $0.0590_{\pm 0.0033}$ & $0.4953_{\pm 0.0288}$ & $0.6202_{\pm 0.0094}$ & $0.0086_{\pm 0.0005}$ & $0.8689_{\pm 0.0169}$ & $0.4091_{\pm 0.0300}$ & $0.3815_{\pm 0.0284}$ & $0.3740_{\pm 0.0251}$ & $0.3269_{\pm 0.0053}$ \\
SU\text{-}GD\text{-}Muon & $0.5301_{\pm 0.0067}$ & $0.0888_{\pm 0.0150}$ & $0.6945_{\pm 0.0079}$ & $0.0976_{\pm 0.0043}$ & $0.4845_{\pm 0.0118}$ & $0.6508_{\pm 0.0053}$ & $0.0174_{\pm 0.0002}$ & $0.9004_{\pm 0.0066}$ & $0.5976_{\pm 0.0187}$ & $0.6132_{\pm 0.0185}$ & $0.5794_{\pm 0.0184}$ & $0.4258_{\pm 0.0304}$ \\
\bottomrule
\end{tabular}
}
\end{table*}

\begin{table*}[htbp]
\centering
\footnotesize
\setlength{\tabcolsep}{3.5pt}
\renewcommand{\arraystretch}{1.10}
\setlength{\heavyrulewidth}{0.9pt}
\setlength{\lightrulewidth}{0.5pt}
\caption{Memorization and Privacy detail --- Llama-3.2-3B-Instruct on Forget 10\%.}
\label{tab:appendix_memgpriv_3b_forget10}
\resizebox{\textwidth}{!}{
\begin{tabular}{lcccccccccccc}
\toprule
\textbf{Method} & \multicolumn{7}{c}{\textbf{Memorization}} & \multicolumn{5}{c}{\textbf{Privacy}} \\
\cmidrule(lr){2-8} \cmidrule(lr){9-13}
 & \textbf{Mem} & ES & EM & F.Prob & F.TR & TR & P.Prob & \textbf{Priv} & MIA.Loss & MIA.ZLib & MIA.MinK & MIA.MinK++ \\
\midrule
\rowcolor{gray!8} \textit{Original (init)} & $0.0283_{\pm 0.0000}$ & $0.8904_{\pm 0.0000}$ & $0.9909_{\pm 0.0000}$ & $0.9510_{\pm 0.0000}$ & $0.4869_{\pm 0.0000}$ & $0.4719_{\pm 0.0000}$ & $0.0944_{\pm 0.0000}$ & $0.3899_{\pm 0.0000}$ & $0.9983_{\pm 0.0000}$ & $0.9992_{\pm 0.0000}$ & $0.9983_{\pm 0.0000}$ & $0.9984_{\pm 0.0000}$ \\
\rowcolor{gray!8} \textit{Retrain (target)} & $0.6408_{\pm 0.0000}$ & $0.0648_{\pm 0.0000}$ & $0.5994_{\pm 0.0000}$ & $0.1241_{\pm 0.0000}$ & $0.6514_{\pm 0.0000}$ & $0.6436_{\pm 0.0000}$ & $0.0523_{\pm 0.0000}$ & $1.0000_{\pm 0.0000}$ & $0.3962_{\pm 0.0000}$ & $0.3119_{\pm 0.0000}$ & $0.3940_{\pm 0.0000}$ & $0.4933_{\pm 0.0000}$ \\
\midrule
GD\text{-}AdamW & $0.2533_{\pm 0.0022}$ & $0.3510_{\pm 0.0045}$ & $0.8963_{\pm 0.0011}$ & $0.5988_{\pm 0.0038}$ & $0.4728_{\pm 0.0007}$ & $0.6957_{\pm 0.0004}$ & $0.0294_{\pm 0.0010}$ & $0.4264_{\pm 0.0011}$ & $0.9807_{\pm 0.0002}$ & $0.9742_{\pm 0.0014}$ & $0.9808_{\pm 0.0002}$ & $0.8656_{\pm 0.0044}$ \\
GD\text{-}Sophia & $0.5688_{\pm 0.0033}$ & $0.0911_{\pm 0.0041}$ & $0.6206_{\pm 0.0072}$ & $0.0469_{\pm 0.0029}$ & $0.4453_{\pm 0.0071}$ & $0.6817_{\pm 0.0020}$ & $0.0028_{\pm 0.0001}$ & $0.8112_{\pm 0.0210}$ & $0.6327_{\pm 0.0213}$ & $0.5485_{\pm 0.0201}$ & $0.5927_{\pm 0.0225}$ & $0.5526_{\pm 0.0200}$ \\
SU\text{-}GD\text{-}Muon & $0.4177_{\pm 0.0061}$ & $0.1733_{\pm 0.0008}$ & $0.7926_{\pm 0.0043}$ & $0.2217_{\pm 0.0139}$ & $0.4424_{\pm 0.0035}$ & $0.6987_{\pm 0.0009}$ & $0.0097_{\pm 0.0002}$ & $0.5339_{\pm 0.0127}$ & $0.8987_{\pm 0.0089}$ & $0.8572_{\pm 0.0130}$ & $0.8928_{\pm 0.0097}$ & $0.7162_{\pm 0.0259}$ \\
\bottomrule
\end{tabular}
}
\end{table*}

\begin{table*}[htbp]
\centering
\footnotesize
\setlength{\tabcolsep}{3.5pt}
\renewcommand{\arraystretch}{1.10}
\setlength{\heavyrulewidth}{0.9pt}
\setlength{\lightrulewidth}{0.5pt}
\caption{Memorization and Privacy detail --- Llama-3.2-3B-Instruct on Forget 5\%.}
\label{tab:appendix_memgpriv_3b_forget05}
\resizebox{\textwidth}{!}{
\begin{tabular}{lcccccccccccc}
\toprule
\textbf{Method} & \multicolumn{7}{c}{\textbf{Memorization}} & \multicolumn{5}{c}{\textbf{Privacy}} \\
\cmidrule(lr){2-8} \cmidrule(lr){9-13}
 & \textbf{Mem} & ES & EM & F.Prob & F.TR & TR & P.Prob & \textbf{Priv} & MIA.Loss & MIA.ZLib & MIA.MinK & MIA.MinK++ \\
\midrule
\rowcolor{gray!8} \textit{Original (init)} & $0.0263_{\pm 0.0000}$ & $0.8881_{\pm 0.0000}$ & $0.9917_{\pm 0.0000}$ & $0.9515_{\pm 0.0000}$ & $0.4778_{\pm 0.0000}$ & $0.6954_{\pm 0.0000}$ & $0.0920_{\pm 0.0000}$ & $0.3629_{\pm 0.0000}$ & $1.0000_{\pm 0.0000}$ & $1.0000_{\pm 0.0000}$ & $1.0000_{\pm 0.0000}$ & $0.9997_{\pm 0.0000}$ \\
\rowcolor{gray!8} \textit{Retrain (target)} & $0.6362_{\pm 0.0000}$ & $0.0618_{\pm 0.0000}$ & $0.6000_{\pm 0.0000}$ & $0.1304_{\pm 0.0000}$ & $0.6363_{\pm 0.0000}$ & $0.5843_{\pm 0.0000}$ & $0.0545_{\pm 0.0000}$ & $0.9999_{\pm 0.0000}$ & $0.3633_{\pm 0.0000}$ & $0.2911_{\pm 0.0000}$ & $0.3604_{\pm 0.0000}$ & $0.4849_{\pm 0.0000}$ \\
\midrule
GD\text{-}AdamW & $0.6235_{\pm 0.0273}$ & $0.0682_{\pm 0.0098}$ & $0.5240_{\pm 0.0449}$ & $0.0276_{\pm 0.0066}$ & $0.4537_{\pm 0.0124}$ & $0.6638_{\pm 0.0150}$ & $0.0021_{\pm 0.0006}$ & $0.8905_{\pm 0.0268}$ & $0.5129_{\pm 0.0320}$ & $0.4369_{\pm 0.0334}$ & $0.4704_{\pm 0.0321}$ & $0.4979_{\pm 0.0172}$ \\
GD\text{-}Sophia & $0.5906_{\pm 0.0348}$ & $0.0802_{\pm 0.0179}$ & $0.5773_{\pm 0.0578}$ & $0.0389_{\pm 0.0108}$ & $0.4428_{\pm 0.0119}$ & $0.6752_{\pm 0.0167}$ & $0.0029_{\pm 0.0010}$ & $0.8964_{\pm 0.0298}$ & $0.5149_{\pm 0.0396}$ & $0.4209_{\pm 0.0376}$ & $0.4624_{\pm 0.0451}$ & $0.4700_{\pm 0.0143}$ \\
SU\text{-}GD\text{-}Muon & $0.6058_{\pm 0.0321}$ & $0.0769_{\pm 0.0090}$ & $0.5665_{\pm 0.0530}$ & $0.0313_{\pm 0.0108}$ & $0.4626_{\pm 0.0095}$ & $0.6718_{\pm 0.0122}$ & $0.0024_{\pm 0.0004}$ & $0.8748_{\pm 0.0068}$ & $0.5416_{\pm 0.0031}$ & $0.4512_{\pm 0.0186}$ & $0.4889_{\pm 0.0058}$ & $0.4849_{\pm 0.0196}$ \\
\bottomrule
\end{tabular}
}
\end{table*}

\begin{table*}[htbp]
\centering
\footnotesize
\setlength{\tabcolsep}{3.5pt}
\renewcommand{\arraystretch}{1.10}
\setlength{\heavyrulewidth}{0.9pt}
\setlength{\lightrulewidth}{0.5pt}
\caption{Memorization and Privacy detail --- Llama-3.2-3B-Instruct on Forget 1\%.}
\label{tab:appendix_memgpriv_3b_forget01}
\resizebox{\textwidth}{!}{
\begin{tabular}{lcccccccccccc}
\toprule
\textbf{Method} & \multicolumn{7}{c}{\textbf{Memorization}} & \multicolumn{5}{c}{\textbf{Privacy}} \\
\cmidrule(lr){2-8} \cmidrule(lr){9-13}
 & \textbf{Mem} & ES & EM & F.Prob & F.TR & TR & P.Prob & \textbf{Priv} & MIA.Loss & MIA.ZLib & MIA.MinK & MIA.MinK++ \\
\midrule
\rowcolor{gray!8} \textit{Original (init)} & $0.0112_{\pm 0.0000}$ & $0.9202_{\pm 0.0000}$ & $0.9968_{\pm 0.0000}$ & $0.9682_{\pm 0.0000}$ & $0.5186_{\pm 0.0000}$ & $0.6693_{\pm 0.0000}$ & $0.0908_{\pm 0.0000}$ & $0.5762_{\pm 0.0000}$ & $1.0000_{\pm 0.0000}$ & $1.0000_{\pm 0.0000}$ & $1.0000_{\pm 0.0000}$ & $1.0000_{\pm 0.0000}$ \\
\rowcolor{gray!8} \textit{Retrain (target)} & $0.6000_{\pm 0.0000}$ & $0.0665_{\pm 0.0000}$ & $0.6484_{\pm 0.0000}$ & $0.1794_{\pm 0.0000}$ & $0.6525_{\pm 0.0000}$ & $0.5830_{\pm 0.0000}$ & $0.0648_{\pm 0.0000}$ & $0.9987_{\pm 0.0000}$ & $0.5538_{\pm 0.0000}$ & $0.5844_{\pm 0.0000}$ & $0.5537_{\pm 0.0000}$ & $0.6144_{\pm 0.0000}$ \\
\midrule
GD\text{-}AdamW & $0.6490_{\pm 0.0209}$ & $0.0658_{\pm 0.0102}$ & $0.5160_{\pm 0.0162}$ & $0.0461_{\pm 0.0057}$ & $0.5060_{\pm 0.0288}$ & $0.5874_{\pm 0.0055}$ & $0.0031_{\pm 0.0010}$ & $0.6150_{\pm 0.0153}$ & $0.1829_{\pm 0.0174}$ & $0.1607_{\pm 0.0149}$ & $0.1613_{\pm 0.0172}$ & $0.2702_{\pm 0.0240}$ \\
GD\text{-}Sophia & $0.6631_{\pm 0.0033}$ & $0.0782_{\pm 0.0105}$ & $0.4999_{\pm 0.0105}$ & $0.0498_{\pm 0.0018}$ & $0.5281_{\pm 0.0096}$ & $0.5965_{\pm 0.0117}$ & $0.0038_{\pm 0.0023}$ & $0.6404_{\pm 0.0087}$ & $0.2039_{\pm 0.0069}$ & $0.1853_{\pm 0.0055}$ & $0.1728_{\pm 0.0080}$ & $0.3178_{\pm 0.0199}$ \\
SU\text{-}GD\text{-}Muon & $0.5552_{\pm 0.0067}$ & $0.1327_{\pm 0.0035}$ & $0.6893_{\pm 0.0112}$ & $0.0858_{\pm 0.0097}$ & $0.5778_{\pm 0.0318}$ & $0.6282_{\pm 0.0140}$ & $0.0107_{\pm 0.0027}$ & $0.8082_{\pm 0.0201}$ & $0.4377_{\pm 0.0143}$ & $0.4552_{\pm 0.0152}$ & $0.4115_{\pm 0.0153}$ & $0.2817_{\pm 0.0327}$ \\
\bottomrule
\end{tabular}
}
\end{table*}

\begin{table*}[htbp]
\centering
\footnotesize
\setlength{\tabcolsep}{3.5pt}
\renewcommand{\arraystretch}{1.10}
\setlength{\heavyrulewidth}{0.9pt}
\setlength{\lightrulewidth}{0.5pt}
\caption{Utility detail --- Llama-3.2-1B-Instruct on Forget 10\%.}
\label{tab:appendix_util_1b_forget10}
\resizebox{\textwidth}{!}{
\begin{tabular}{lccccccccccc}
\toprule
\textbf{Method} & \multicolumn{11}{c}{\textbf{Utility}} \\
\cmidrule(lr){2-12}
 & \textbf{Util} & Fluency & R.Prob & R.ROUGE & R.TR & RA.Prob & RA.ROUGE & RA.TR & WF.Prob & WF.ROUGE & WF.TR \\
\midrule
\rowcolor{gray!8} \textit{Original (init)} & $0.6072_{\pm 0.0000}$ & $0.8594_{\pm 0.0000}$ & $0.8706_{\pm 0.0000}$ & $0.7970_{\pm 0.0000}$ & $0.5175_{\pm 0.0000}$ & $0.4137_{\pm 0.0000}$ & $0.8170_{\pm 0.0000}$ & $0.5274_{\pm 0.0000}$ & $0.4356_{\pm 0.0000}$ & $0.8279_{\pm 0.0000}$ & $0.6198_{\pm 0.0000}$ \\
\rowcolor{gray!8} \textit{Retrain (target)} & $0.6105_{\pm 0.0000}$ & $0.9052_{\pm 0.0000}$ & $0.8803_{\pm 0.0000}$ & $0.8254_{\pm 0.0000}$ & $0.5136_{\pm 0.0000}$ & $0.3918_{\pm 0.0000}$ & $0.8045_{\pm 0.0000}$ & $0.5048_{\pm 0.0000}$ & $0.4286_{\pm 0.0000}$ & $0.8194_{\pm 0.0000}$ & $0.6181_{\pm 0.0000}$ \\
\midrule
GD\text{-}AdamW & $0.4539_{\pm 0.0062}$ & $0.7000_{\pm 0.0491}$ & $0.3331_{\pm 0.0253}$ & $0.3918_{\pm 0.0100}$ & $0.5390_{\pm 0.0022}$ & $0.3317_{\pm 0.0087}$ & $0.4703_{\pm 0.0206}$ & $0.4661_{\pm 0.0109}$ & $0.3744_{\pm 0.0122}$ & $0.7598_{\pm 0.0215}$ & $0.5570_{\pm 0.0013}$ \\
GD\text{-}Sophia & $0.4606_{\pm 0.0079}$ & $0.7400_{\pm 0.0489}$ & $0.3292_{\pm 0.0271}$ & $0.3900_{\pm 0.0087}$ & $0.5362_{\pm 0.0020}$ & $0.3339_{\pm 0.0094}$ & $0.5163_{\pm 0.0361}$ & $0.4724_{\pm 0.0074}$ & $0.3780_{\pm 0.0063}$ & $0.7736_{\pm 0.0060}$ & $0.5578_{\pm 0.0051}$ \\
SU\text{-}GD\text{-}Muon & $0.4085_{\pm 0.0082}$ & $0.5050_{\pm 0.0956}$ & $0.2484_{\pm 0.0106}$ & $0.3657_{\pm 0.0067}$ & $0.5573_{\pm 0.0111}$ & $0.3091_{\pm 0.0126}$ & $0.4257_{\pm 0.0349}$ & $0.4557_{\pm 0.0056}$ & $0.3613_{\pm 0.0110}$ & $0.7228_{\pm 0.0270}$ & $0.5662_{\pm 0.0088}$ \\
\bottomrule
\end{tabular}
}
\end{table*}

\begin{table*}[htbp]
\centering
\footnotesize
\setlength{\tabcolsep}{3.5pt}
\renewcommand{\arraystretch}{1.10}
\setlength{\heavyrulewidth}{0.9pt}
\setlength{\lightrulewidth}{0.5pt}
\caption{Utility detail --- Llama-3.2-1B-Instruct on Forget 5\%.}
\label{tab:appendix_util_1b_forget05}
\resizebox{\textwidth}{!}{
\begin{tabular}{lccccccccccc}
\toprule
\textbf{Method} & \multicolumn{11}{c}{\textbf{Utility}} \\
\cmidrule(lr){2-12}
 & \textbf{Util} & Fluency & R.Prob & R.ROUGE & R.TR & RA.Prob & RA.ROUGE & RA.TR & WF.Prob & WF.ROUGE & WF.TR \\
\midrule
\rowcolor{gray!8} \textit{Original (init)} & $0.6142_{\pm 0.0000}$ & $0.8611_{\pm 0.0000}$ & $0.8706_{\pm 0.0000}$ & $0.7970_{\pm 0.0000}$ & $0.5175_{\pm 0.0000}$ & $0.4137_{\pm 0.0000}$ & $0.8170_{\pm 0.0000}$ & $0.5274_{\pm 0.0000}$ & $0.4356_{\pm 0.0000}$ & $0.8279_{\pm 0.0000}$ & $0.6198_{\pm 0.0000}$ \\
\rowcolor{gray!8} \textit{Retrain (target)} & $0.6192_{\pm 0.0000}$ & $0.9102_{\pm 0.0000}$ & $0.8765_{\pm 0.0000}$ & $0.8156_{\pm 0.0000}$ & $0.5182_{\pm 0.0000}$ & $0.3986_{\pm 0.0000}$ & $0.8395_{\pm 0.0000}$ & $0.5123_{\pm 0.0000}$ & $0.4375_{\pm 0.0000}$ & $0.8348_{\pm 0.0000}$ & $0.6225_{\pm 0.0000}$ \\
\midrule
GD\text{-}AdamW & $0.4846_{\pm 0.0056}$ & $0.7630_{\pm 0.0128}$ & $0.4019_{\pm 0.0162}$ & $0.4082_{\pm 0.0066}$ & $0.5335_{\pm 0.0051}$ & $0.3464_{\pm 0.0037}$ & $0.5137_{\pm 0.0036}$ & $0.4820_{\pm 0.0055}$ & $0.3933_{\pm 0.0041}$ & $0.7887_{\pm 0.0107}$ & $0.5767_{\pm 0.0077}$ \\
GD\text{-}Sophia & $0.4085_{\pm 0.0304}$ & $0.6358_{\pm 0.0155}$ & $0.2126_{\pm 0.0610}$ & $0.3795_{\pm 0.0130}$ & $0.5479_{\pm 0.0094}$ & $0.3210_{\pm 0.0005}$ & $0.4681_{\pm 0.0342}$ & $0.4705_{\pm 0.0040}$ & $0.3466_{\pm 0.0110}$ & $0.7604_{\pm 0.0263}$ & $0.5735_{\pm 0.0052}$ \\
SU\text{-}GD\text{-}Muon & $0.4383_{\pm 0.0086}$ & $0.7153_{\pm 0.0218}$ & $0.2674_{\pm 0.0330}$ & $0.3952_{\pm 0.0150}$ & $0.5328_{\pm 0.0062}$ & $0.3236_{\pm 0.0040}$ & $0.4984_{\pm 0.0330}$ & $0.4646_{\pm 0.0060}$ & $0.3576_{\pm 0.0058}$ & $0.7555_{\pm 0.0095}$ & $0.5601_{\pm 0.0088}$ \\
\bottomrule
\end{tabular}
}
\end{table*}

\begin{table*}[htbp]
\centering
\footnotesize
\setlength{\tabcolsep}{3.5pt}
\renewcommand{\arraystretch}{1.10}
\setlength{\heavyrulewidth}{0.9pt}
\setlength{\lightrulewidth}{0.5pt}
\caption{Utility detail --- Llama-3.2-1B-Instruct on Forget 1\%.}
\label{tab:appendix_util_1b_forget01}
\resizebox{\textwidth}{!}{
\begin{tabular}{lccccccccccc}
\toprule
\textbf{Method} & \multicolumn{11}{c}{\textbf{Utility}} \\
\cmidrule(lr){2-12}
 & \textbf{Util} & Fluency & R.Prob & R.ROUGE & R.TR & RA.Prob & RA.ROUGE & RA.TR & WF.Prob & WF.ROUGE & WF.TR \\
\midrule
\rowcolor{gray!8} \textit{Original (init)} & $0.6154_{\pm 0.0000}$ & $0.8952_{\pm 0.0000}$ & $0.8706_{\pm 0.0000}$ & $0.7970_{\pm 0.0000}$ & $0.5175_{\pm 0.0000}$ & $0.4137_{\pm 0.0000}$ & $0.8170_{\pm 0.0000}$ & $0.5274_{\pm 0.0000}$ & $0.4356_{\pm 0.0000}$ & $0.8279_{\pm 0.0000}$ & $0.6198_{\pm 0.0000}$ \\
\rowcolor{gray!8} \textit{Retrain (target)} & $0.6172_{\pm 0.0000}$ & $0.8720_{\pm 0.0000}$ & $0.8790_{\pm 0.0000}$ & $0.8080_{\pm 0.0000}$ & $0.5151_{\pm 0.0000}$ & $0.4121_{\pm 0.0000}$ & $0.7620_{\pm 0.0000}$ & $0.5215_{\pm 0.0000}$ & $0.4380_{\pm 0.0000}$ & $0.8519_{\pm 0.0000}$ & $0.6212_{\pm 0.0000}$ \\
\midrule
GD\text{-}AdamW & $0.4877_{\pm 0.0015}$ & $0.7153_{\pm 0.0437}$ & $0.3565_{\pm 0.0063}$ & $0.4076_{\pm 0.0033}$ & $0.5254_{\pm 0.0094}$ & $0.3659_{\pm 0.0041}$ & $0.5998_{\pm 0.0306}$ & $0.5006_{\pm 0.0095}$ & $0.4002_{\pm 0.0079}$ & $0.7780_{\pm 0.0187}$ & $0.5844_{\pm 0.0028}$ \\
GD\text{-}Sophia & $0.4659_{\pm 0.0040}$ & $0.6854_{\pm 0.0088}$ & $0.2970_{\pm 0.0175}$ & $0.3873_{\pm 0.0014}$ & $0.5209_{\pm 0.0069}$ & $0.3658_{\pm 0.0035}$ & $0.5436_{\pm 0.0255}$ & $0.5008_{\pm 0.0066}$ & $0.3975_{\pm 0.0096}$ & $0.7732_{\pm 0.0143}$ & $0.5830_{\pm 0.0071}$ \\
SU\text{-}GD\text{-}Muon & $0.5194_{\pm 0.0028}$ & $0.8232_{\pm 0.0391}$ & $0.4968_{\pm 0.0312}$ & $0.4528_{\pm 0.0001}$ & $0.5318_{\pm 0.0051}$ & $0.3699_{\pm 0.0057}$ & $0.6196_{\pm 0.0175}$ & $0.4965_{\pm 0.0056}$ & $0.3987_{\pm 0.0108}$ & $0.7887_{\pm 0.0164}$ & $0.5705_{\pm 0.0135}$ \\
\bottomrule
\end{tabular}
}
\end{table*}

\begin{table*}[htbp]
\centering
\footnotesize
\setlength{\tabcolsep}{3.5pt}
\renewcommand{\arraystretch}{1.10}
\setlength{\heavyrulewidth}{0.9pt}
\setlength{\lightrulewidth}{0.5pt}
\caption{Utility detail --- Llama-3.2-3B-Instruct on Forget 10\%.}
\label{tab:appendix_util_3b_forget10}
\resizebox{\textwidth}{!}{
\begin{tabular}{lccccccccccc}
\toprule
\textbf{Method} & \multicolumn{11}{c}{\textbf{Utility}} \\
\cmidrule(lr){2-12}
 & \textbf{Util} & Fluency & R.Prob & R.ROUGE & R.TR & RA.Prob & RA.ROUGE & RA.TR & WF.Prob & WF.ROUGE & WF.TR \\
\midrule
\rowcolor{gray!8} \textit{Original (init)} & $0.6662_{\pm 0.0000}$ & $0.8614_{\pm 0.0000}$ & $0.9429_{\pm 0.0000}$ & $0.9197_{\pm 0.0000}$ & $0.5409_{\pm 0.0000}$ & $0.4795_{\pm 0.0000}$ & $0.8513_{\pm 0.0000}$ & $0.6244_{\pm 0.0000}$ & $0.4941_{\pm 0.0000}$ & $0.8932_{\pm 0.0000}$ & $0.6622_{\pm 0.0000}$ \\
\rowcolor{gray!8} \textit{Retrain (target)} & $0.6678_{\pm 0.0000}$ & $0.8895_{\pm 0.0000}$ & $0.9425_{\pm 0.0000}$ & $0.9192_{\pm 0.0000}$ & $0.5352_{\pm 0.0000}$ & $0.4615_{\pm 0.0000}$ & $0.8515_{\pm 0.0000}$ & $0.5951_{\pm 0.0000}$ & $0.4732_{\pm 0.0000}$ & $0.8818_{\pm 0.0000}$ & $0.6396_{\pm 0.0000}$ \\
\midrule
GD\text{-}AdamW & $0.6348_{\pm 0.0007}$ & $0.8561_{\pm 0.0057}$ & $0.8326_{\pm 0.0013}$ & $0.6652_{\pm 0.0039}$ & $0.5665_{\pm 0.0011}$ & $0.4587_{\pm 0.0040}$ & $0.7660_{\pm 0.0059}$ & $0.6091_{\pm 0.0059}$ & $0.4796_{\pm 0.0035}$ & $0.8566_{\pm 0.0035}$ & $0.6513_{\pm 0.0043}$ \\
GD\text{-}Sophia & $0.4658_{\pm 0.0034}$ & $0.3179_{\pm 0.0066}$ & $0.2979_{\pm 0.0075}$ & $0.3861_{\pm 0.0035}$ & $0.6049_{\pm 0.0029}$ & $0.4635_{\pm 0.0132}$ & $0.5996_{\pm 0.0248}$ & $0.6242_{\pm 0.0121}$ & $0.4599_{\pm 0.0068}$ & $0.8215_{\pm 0.0102}$ & $0.6392_{\pm 0.0030}$ \\
SU\text{-}GD\text{-}Muon & $0.5839_{\pm 0.0030}$ & $0.7460_{\pm 0.0057}$ & $0.6030_{\pm 0.0120}$ & $0.5223_{\pm 0.0073}$ & $0.5941_{\pm 0.0012}$ & $0.4634_{\pm 0.0119}$ & $0.6860_{\pm 0.0066}$ & $0.6175_{\pm 0.0136}$ & $0.4480_{\pm 0.0023}$ & $0.8378_{\pm 0.0017}$ & $0.6290_{\pm 0.0028}$ \\
\bottomrule
\end{tabular}
}
\end{table*}

\begin{table*}[htbp]
\centering
\footnotesize
\setlength{\tabcolsep}{3.5pt}
\renewcommand{\arraystretch}{1.10}
\setlength{\heavyrulewidth}{0.9pt}
\setlength{\lightrulewidth}{0.5pt}
\caption{Utility detail --- Llama-3.2-3B-Instruct on Forget 5\%.}
\label{tab:appendix_util_3b_forget05}
\resizebox{\textwidth}{!}{
\begin{tabular}{lccccccccccc}
\toprule
\textbf{Method} & \multicolumn{11}{c}{\textbf{Utility}} \\
\cmidrule(lr){2-12}
 & \textbf{Util} & Fluency & R.Prob & R.ROUGE & R.TR & RA.Prob & RA.ROUGE & RA.TR & WF.Prob & WF.ROUGE & WF.TR \\
\midrule
\rowcolor{gray!8} \textit{Original (init)} & $0.6740_{\pm 0.0000}$ & $0.8535_{\pm 0.0000}$ & $0.9430_{\pm 0.0000}$ & $0.9151_{\pm 0.0000}$ & $0.5404_{\pm 0.0000}$ & $0.4789_{\pm 0.0000}$ & $0.8513_{\pm 0.0000}$ & $0.6244_{\pm 0.0000}$ & $0.4934_{\pm 0.0000}$ & $0.8778_{\pm 0.0000}$ & $0.6642_{\pm 0.0000}$ \\
\rowcolor{gray!8} \textit{Retrain (target)} & $0.6753_{\pm 0.0000}$ & $0.8665_{\pm 0.0000}$ & $0.9420_{\pm 0.0000}$ & $0.9117_{\pm 0.0000}$ & $0.5343_{\pm 0.0000}$ & $0.4769_{\pm 0.0000}$ & $0.8413_{\pm 0.0000}$ & $0.6281_{\pm 0.0000}$ & $0.4823_{\pm 0.0000}$ & $0.8846_{\pm 0.0000}$ & $0.6494_{\pm 0.0000}$ \\
\midrule
GD\text{-}AdamW & $0.3570_{\pm 0.0393}$ & $0.1814_{\pm 0.0168}$ & $0.1758_{\pm 0.0554}$ & $0.2574_{\pm 0.0278}$ & $0.6039_{\pm 0.0086}$ & $0.4976_{\pm 0.0072}$ & $0.5159_{\pm 0.0584}$ & $0.6484_{\pm 0.0119}$ & $0.4707_{\pm 0.0056}$ & $0.7230_{\pm 0.0326}$ & $0.6750_{\pm 0.0021}$ \\
GD\text{-}Sophia & $0.4228_{\pm 0.0651}$ & $0.2826_{\pm 0.0621}$ & $0.2390_{\pm 0.0855}$ & $0.3247_{\pm 0.0617}$ & $0.6149_{\pm 0.0080}$ & $0.4898_{\pm 0.0118}$ & $0.5760_{\pm 0.0496}$ & $0.6389_{\pm 0.0123}$ & $0.4644_{\pm 0.0087}$ & $0.7771_{\pm 0.0501}$ & $0.6515_{\pm 0.0139}$ \\
SU\text{-}GD\text{-}Muon & $0.3902_{\pm 0.0575}$ & $0.2234_{\pm 0.0605}$ & $0.2183_{\pm 0.0754}$ & $0.2766_{\pm 0.0531}$ & $0.6102_{\pm 0.0026}$ & $0.4977_{\pm 0.0046}$ & $0.5446_{\pm 0.0400}$ & $0.6394_{\pm 0.0078}$ & $0.4717_{\pm 0.0060}$ & $0.6963_{\pm 0.0778}$ & $0.6614_{\pm 0.0044}$ \\
\bottomrule
\end{tabular}
}
\end{table*}

\begin{table*}[htbp]
\centering
\footnotesize
\setlength{\tabcolsep}{3.5pt}
\renewcommand{\arraystretch}{1.10}
\setlength{\heavyrulewidth}{0.9pt}
\setlength{\lightrulewidth}{0.5pt}
\caption{Utility detail --- Llama-3.2-3B-Instruct on Forget 1\%.}
\label{tab:appendix_util_3b_forget01}
\resizebox{\textwidth}{!}{
\begin{tabular}{lccccccccccc}
\toprule
\textbf{Method} & \multicolumn{11}{c}{\textbf{Utility}} \\
\cmidrule(lr){2-12}
 & \textbf{Util} & Fluency & R.Prob & R.ROUGE & R.TR & RA.Prob & RA.ROUGE & RA.TR & WF.Prob & WF.ROUGE & WF.TR \\
\midrule
\rowcolor{gray!8} \textit{Original (init)} & $0.6785_{\pm 0.0000}$ & $0.8938_{\pm 0.0000}$ & $0.9430_{\pm 0.0000}$ & $0.9151_{\pm 0.0000}$ & $0.5404_{\pm 0.0000}$ & $0.4789_{\pm 0.0000}$ & $0.8513_{\pm 0.0000}$ & $0.6244_{\pm 0.0000}$ & $0.4934_{\pm 0.0000}$ & $0.8778_{\pm 0.0000}$ & $0.6642_{\pm 0.0000}$ \\
\rowcolor{gray!8} \textit{Retrain (target)} & $0.6787_{\pm 0.0000}$ & $0.8833_{\pm 0.0000}$ & $0.9425_{\pm 0.0000}$ & $0.9156_{\pm 0.0000}$ & $0.5381_{\pm 0.0000}$ & $0.4763_{\pm 0.0000}$ & $0.8780_{\pm 0.0000}$ & $0.6208_{\pm 0.0000}$ & $0.4883_{\pm 0.0000}$ & $0.8761_{\pm 0.0000}$ & $0.6522_{\pm 0.0000}$ \\
\midrule
GD\text{-}AdamW & $0.4686_{\pm 0.0266}$ & $0.3560_{\pm 0.0777}$ & $0.2725_{\pm 0.0245}$ & $0.3884_{\pm 0.0235}$ & $0.5910_{\pm 0.0060}$ & $0.4820_{\pm 0.0156}$ & $0.6356_{\pm 0.0116}$ & $0.6314_{\pm 0.0178}$ & $0.4552_{\pm 0.0123}$ & $0.8205_{\pm 0.0317}$ & $0.6430_{\pm 0.0187}$ \\
GD\text{-}Sophia & $0.4403_{\pm 0.0037}$ & $0.3416_{\pm 0.0488}$ & $0.2312_{\pm 0.0120}$ & $0.3672_{\pm 0.0063}$ & $0.5875_{\pm 0.0034}$ & $0.4375_{\pm 0.0103}$ & $0.5919_{\pm 0.0296}$ & $0.5963_{\pm 0.0165}$ & $0.4327_{\pm 0.0114}$ & $0.8262_{\pm 0.0205}$ & $0.6191_{\pm 0.0191}$ \\
SU\text{-}GD\text{-}Muon & $0.6050_{\pm 0.0150}$ & $0.7800_{\pm 0.0562}$ & $0.6216_{\pm 0.0761}$ & $0.5389_{\pm 0.0426}$ & $0.5774_{\pm 0.0057}$ & $0.4797_{\pm 0.0125}$ & $0.7490_{\pm 0.0148}$ & $0.6280_{\pm 0.0148}$ & $0.4770_{\pm 0.0020}$ & $0.8452_{\pm 0.0076}$ & $0.6593_{\pm 0.0044}$ \\
\bottomrule
\end{tabular}
}
\end{table*}

\clearpage
\section{Limitations and Broader Impacts}
\label{app:limitations}

Our evaluation extends TOFU and WMDP under the OpenUnlearning framework. Both benchmarks rely on fixed-format prompts or 4-way multiple-choice questions, which may not fully capture the model's residual knowledge under more diverse query distributions. The default forget-side aggregation combines paraphrase probability—the model's likelihood on paraphrased forget prompts—with the forget truth ratio, which compares likelihoods between paraphrased correct answers and perturbed incorrect ones. Both rely on paraphrased forget prompts and are therefore strongly correlated. As discussed in App.~\ref{app:metric_details}, we replace paraphrase probability with the forget probability evaluated on the original (non-paraphrased) prompts, so that the paraphrase signal enters the aggregate only through the truth ratio. This reduces direct correlations between the aggregated components, but indirect correlations between submetrics remain, and refining the evaluation protocol toward greater metric independence is left to future work. To support such investigations, our detailed sub-metric tables in Tabs.~\ref{tab:appendix_memgpriv_1b_forget10}--\ref{tab:appendix_memgpriv_3b_forget01} report both our protocol's components and the OpenUnlearning defaults, enabling reanalysis under alternative aggregations.

Our work improves machine unlearning for LLMs, which contributes to data privacy compliance (e.g., the right to be forgotten) and safer deployment by enabling more thorough removal of private, copyrighted, or hazardous knowledge. A potential negative impact is that more effective unlearning could be misused to selectively erase desirable knowledge or audit traces from a model; however, this risk is shared with all unlearning research and is not specific to our method.

Spectral updates incur higher time and memory overhead than entry-wise updates,
and their behavior at frontier scale remains untested.
The choice of $\ell_\infty$ layer aggregation in Theorem~2 is theoretically motivated,
but its empirical advantage over $\ell_1$ or $\ell_2$ remains open.
Finally, since spectral updates also benefit retain-only fine-tuning,
separating an unlearning-specific effect from a general optimization gain is left to future work.

\clearpage
\input{checklist.tex}

\end{document}

%% file: checklist.tex
\section*{NeurIPS Paper Checklist}

\begin{enumerate}

\item {\bf Claims}
    \item[] Question: Do the main claims made in the abstract and introduction accurately reflect the paper's contributions and scope?
    \item[] Answer: \answerYes{} %
    \item[] Justification: The main contributions claimed in the abstract and introduction are substantiated in the main body: the proposed spectral unlearning method is presented in Sec.~\ref{sec:method}, its theoretical analysis is provided in Sec.~\ref{sec:theory}, and empirical results supporting the claims are reported in Sec.~\ref{sec:exp}.

\item {\bf Limitations}
    \item[] Question: Does the paper discuss the limitations of the work performed by the authors?
    \item[] Answer: \answerYes{} %
    \item[] Justification: Please refer to App.~\ref{app:limitations}.

\item {\bf Theory assumptions and proofs}
    \item[] Question: For each theoretical result, does the paper provide the full set of assumptions and a complete (and correct) proof?
    \item[] Answer: \answerYes{} %
    \item[] Justification: Our theoretical results assume that each weight matrix operates on RMS-normalized inputs, which is the standard setup in pre-norm transformer architectures. The full set of assumptions is stated in Sec.~\ref{sec:transformer_structure}. We note that not all weight matrices in an LLM strictly satisfy the RMS-input assumption. We discuss these practical considerations in App.~\ref{app:practical_realization}.


    \item {\bf Experimental result reproducibility}
    \item[] Question: Does the paper fully disclose all the information needed to reproduce the main experimental results of the paper to the extent that it affects the main claims and/or conclusions of the paper (regardless of whether the code and data are provided or not)?
    \item[] Answer: \answerYes{} %
    \item[] Justification: We provide all hyperparameters required to reproduce our experiments in App.~\ref{app:hyper}. In addition, we report the complete raw sub-metric results across all models in App.~\ref{app:Detail_metric}.

\item {\bf Open access to data and code}
    \item[] Question: Does the paper provide open access to the data and code, with sufficient instructions to faithfully reproduce the main experimental results, as described in supplemental material?
    \item[] Answer: \answerYes{} %
    \item[] Justification: We use the public TOFU~\cite{maini2024tofu}, RWKU~\cite{NEURIPS2024_b1f78dfc} and WMDP~\cite{pmlr-v235-li24bc} benchmarks under the OpenUnlearning framework~\cite{dorna2026openunlearning}. We provide our code with a detailed \texttt{README.md} that specifies the environment setup and the commands required to reproduce all experimental results.

\item {\bf Experimental setting/details}
    \item[] Question: Does the paper specify all the training and test details (e.g., data splits, hyperparameters, how they were chosen, type of optimizer) necessary to understand the results?
    \item[] Answer: \answerYes{} %
    \item[] Justification: Please refer to Sec.~\ref{sec:exp} and App.~\ref{app:hyper}.

\item {\bf Experiment statistical significance}
    \item[] Question: Does the paper report error bars suitably and correctly defined or other appropriate information about the statistical significance of the experiments?
    \item[] Answer: \answerYes{} %
    \item[] Justification: All results are averaged over three random seeds.

\item {\bf Experiments compute resources}
    \item[] Question: For each experiment, does the paper provide sufficient information on the computer resources (type of compute workers, memory, time of execution) needed to reproduce the experiments?
    \item[] Answer: \answerYes{} %
    \item[] Justification: Please refer to App.~\ref{app:hyper}.
    
\item {\bf Code of ethics}
    \item[] Question: Does the research conducted in the paper conform, in every respect, with the NeurIPS Code of Ethics \url{https://neurips.cc/public/EthicsGuidelines}?
    \item[] Answer: \answerYes{} %
    \item[] Justification: This work is conducted according to the NeurIPS code of ethics.

\item {\bf Broader impacts}
    \item[] Question: Does the paper discuss both potential positive societal impacts and negative societal impacts of the work performed?
    \item[] Answer: \answerYes{} %
    \item[] Justification: Please refer to App.~\ref{app:limitations}.
    
\item {\bf Safeguards}
    \item[] Question: Does the paper describe safeguards that have been put in place for responsible release of data or models that have a high risk for misuse (e.g., pre-trained language models, image generators, or scraped datasets)?
    \item[] Answer: \answerNA{} %
    \item[] Justification: Our work does not release any new pre-trained models or datasets. We use publicly available benchmarks and publicly available pre-trained models.

\item {\bf Licenses for existing assets}
    \item[] Question: Are the creators or original owners of assets (e.g., code, data, models), used in the paper, properly credited and are the license and terms of use explicitly mentioned and properly respected?
    \item[] Answer: \answerYes{} %
    \item[] Justification: The models and the datasets are properly cited.

\item {\bf New assets}
    \item[] Question: Are new assets introduced in the paper well documented and is the documentation provided alongside the assets?
    \item[] Answer: \answerYes{} %
    \item[] Justification: the code is released along with clear documentation and instructions for reproduction.

\item {\bf Crowdsourcing and research with human subjects}
    \item[] Question: For crowdsourcing experiments and research with human subjects, does the paper include the full text of instructions given to participants and screenshots, if applicable, as well as details about compensation (if any)? 
    \item[] Answer: \answerNA{} %
    \item[] Justification: This work does not involve crowdsourcing.

\item {\bf Institutional review board (IRB) approvals or equivalent for research with human subjects}
    \item[] Question: Does the paper describe potential risks incurred by study participants, whether such risks were disclosed to the subjects, and whether Institutional Review Board (IRB) approvals (or an equivalent approval/review based on the requirements of your country or institution) were obtained?
    \item[] Answer: \answerNA{} %
    \item[] Justification: This work does not involve any human subjects.


\item {\bf Declaration of LLM usage}
    \item[] Question: Does the paper describe the usage of LLMs if it is an important, original, or non-standard component of the core methods in this research? Note that if the LLM is used only for writing, editing, or formatting purposes and does \emph{not} impact the core methodology, scientific rigor, or originality of the research, declaration is not required.
    \item[] Answer: \answerNA{} %
    \item[] Justification: We do not use LLMs as a non-standard tool in our method development.

\end{enumerate}